\documentclass[times,twocolumn,final]{elsarticle}

\usepackage{medima}
\usepackage{framed,multirow}
\usepackage[fleqn]{amsmath}
\usepackage{amssymb,amsfonts}
\usepackage{algorithmic}

\usepackage{amssymb}
\usepackage{latexsym}

\usepackage{url}
\usepackage{xcolor}
\usepackage{booktabs}
\usepackage{pifont}
\usepackage{etoolbox}
\BeforeBeginEnvironment{tabular}{\tiny}

\usepackage{booktabs}

\definecolor{SkyBlue}{RGB}{70, 196, 225}
\AtBeginDocument{
    \hypersetup{
      linkcolor=SkyBlue,   
      anchorcolor=SkyBlue, 
      citecolor=SkyBlue,    
      filecolor=SkyBlue,  
      urlcolor=SkyBlue      
    }
  }
\usepackage[colorlinks,
            linkcolor=SkyBlue,
            anchorcolor=SkyBlue,
            citecolor=SkyBlue,
            filecolor=SkyBlue,  
            urlcolor=SkyBlue]{hyperref}
\usepackage{caption}
\definecolor{newcolor}{rgb}{.8,.349,.1}

\journal{Medical Image Analysis}

\begin{document}


\verso{Le Jiang \textit{et~al.}}

\begin{frontmatter}

\title{Shadow defense against gradient inversion attack in federated learning}%
\tnotetext[tnote1]{Code is available online at \hyperref[https://github.com/tekap404/ShadowDef]{https://github.com/tekap404/ShadowDef}.}

\author[a]{Le Jiang}
\author[b]{Liyan Ma\corref{cor1}}
\author[a,c,d,e]{Guang Yang\corref{cor1}}
\cortext[cor1]{Corresponding Authors: Liyan Ma and Guang Yang E-mail addresses: liyanma@shu.edu.cn; g.yang@imperial.ac.uk;}

\address[a]{Bioengineering Department and Imperial-X, Imperial College London, London W12 7SL, UK}
\address[b]{School of Computer Engineering and Science, Shanghai university, Shanghai 200444, China}
\address[c]{National Heart and Lung Institute, Imperial College London, London SW7 2AZ, UK}
\address[d]{Cardiovascular Research Centre, Royal Brompton Hospital, London SW3 6NP, UK}
\address[e]{School of Biomedical Engineering \& Imaging Sciences, King's College London, London WC2R 2LS, UK}

\received{1 May 2013}
\finalform{10 May 2013}
\accepted{13 May 2013}
\availableonline{15 May 2013}
\communicated{S. Sarkar}
\begin{keyword}
  \KWD \\Federated learning\\Gradient inversion attack\\Medical images\\Privacy protection
\end{keyword}

\begin{abstract}
Federated learning (FL) has emerged as a transformative framework for privacy-preserving distributed training, allowing clients to collaboratively train a global model without sharing their local data. This is especially crucial in sensitive fields like healthcare, where protecting patient data is paramount. However, privacy leakage remains a critical challenge, as the communication of model updates can be exploited by potential adversaries. Gradient inversion attacks (GIAs), for instance, allow adversaries to approximate the gradients used for training and reconstruct training images, thus stealing patient privacy. Existing defense mechanisms obscure gradients, yet lack a nuanced understanding of which gradients or types of image information are most vulnerable to such attacks. These indiscriminate calibrated perturbations result in either excessive privacy protection degrading model accuracy, or insufficient one failing to safeguard sensitive information. Therefore, we introduce a framework that addresses these challenges by leveraging a shadow model with interpretability for identifying sensitive areas. This enables a more targeted and sample-specific noise injection. Specially, our defensive strategy achieves discrepancies of 3.73 in PSNR and 0.2 in SSIM compared to the circumstance without defense on the ChestXRay dataset, and 2.78 in PSNR and 0.166 in the EyePACS dataset. Moreover, it minimizes adverse effects on model performance, with less than 1\% F1 reduction compared to SOTA methods. Our extensive experiments, conducted across diverse types of medical images, validate the generalization of the proposed framework. The stable defense improvements for FedAvg are consistently over 1.5\% times in LPIPS and SSIM. It also offers a universal defense against various GIA types, especially for these sensitive areas in images.
\end{abstract}

\end{frontmatter}


\section{Introduction}
\label{intro}
As a response to the data privacy law (\cite{gdpr2016general}), federated learning (FL) (\cite{mcmahan2017communication}) has been developed to permit the joint training of deep learning models (\cite{lecun2015deep}) without necessitating exchange of original data. In FL, a global model is collaboratively trained across multiple global epochs to yield task performance similar to ones trained with centralized data. At the start of each global epoch, clients are allocated with a global model from the server, which forms the basis for local models. Subsequent to several local training epochs, trained local models or their updates are uploaded to the server for aggregation into the global model, thereby avoiding the need to transmit any original data.

Although FL is regarded as a learning framework that protects privacy, recent research (\cite{huang2021evaluating}) has revealed that transferring updated models or gradients to the server carries the risk of training data being reconstructed. The reconstructive strategy, known as gradient inversion attack (GIA) (\cite{xu2022agic, geng2023improved, liang2023egia}), aims to minimize the difference between real gradients and reconstructed ones. It also employs auxiliary information (\cite{yin2021see}), such as batch normalization (BN) statistics, to improve the fidelity of reconstructed data. Due to a large amount of patient privacy in medical images, such attack may lead to severe violation of privacy protection laws, which is against the principle of FL. Therefore, it is crucial to establish defensive measures against GIA to ensure data privacy.

\begin{figure*}[ht]
  \begin{center}
  \includegraphics[width=2\columnwidth]{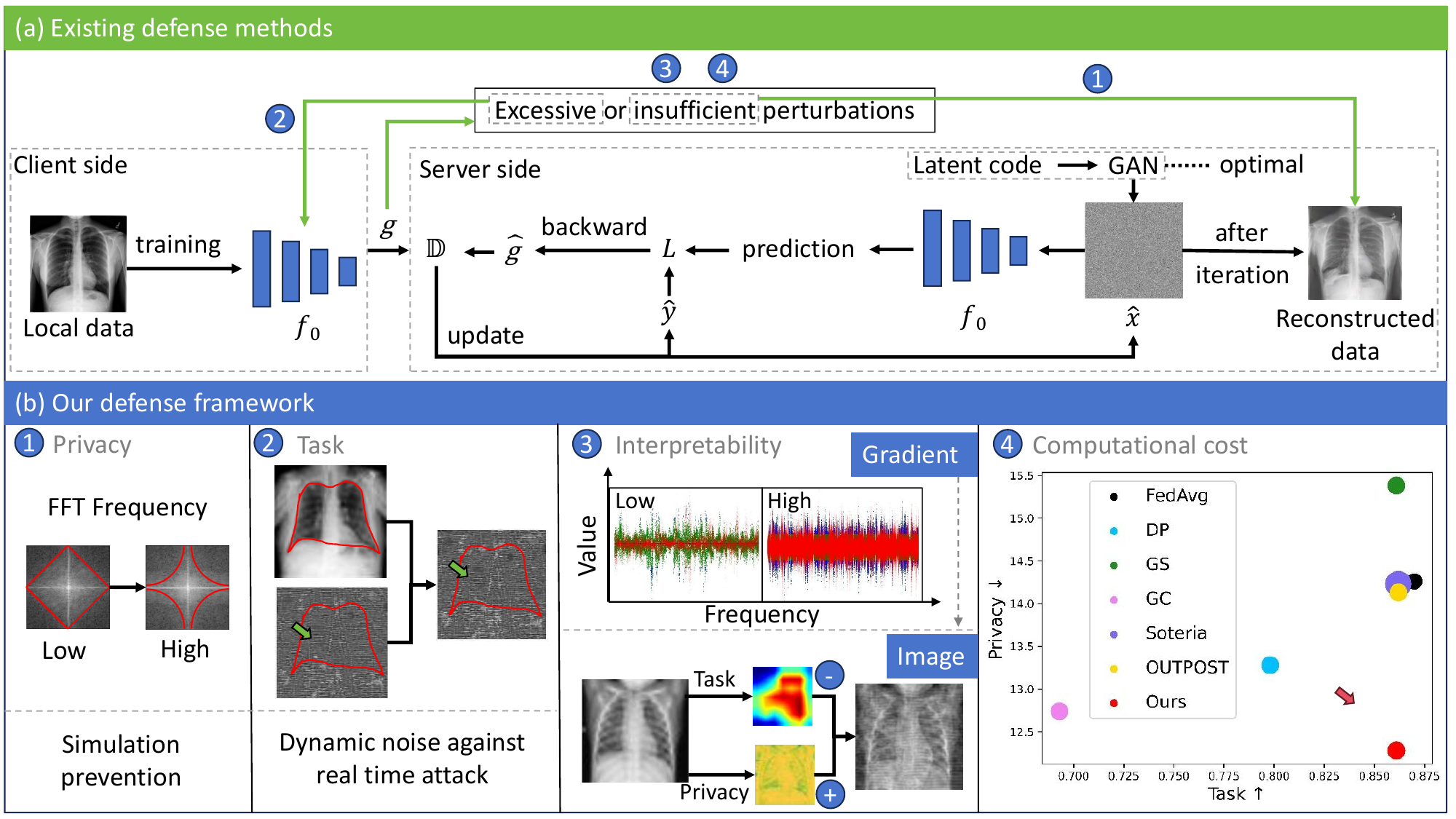}
  \end{center}
  \caption{(a) Basic pipeline of gradient inversion attack and drawbacks of defense methods. The curious server reconstructs dummy images $\hat{x}$ and labels $\hat{y}$ to mimic training data from clients based on their uploaded gradients $g$ of the global model $f_0$. The gradient inversion loss $\mathbb{D}$ is calculated between $g$ and dummy gradients $\hat{g}$. Model-based GIA additionally uses GAN to recover these images. Four challenges are in existing GIA defense works: \ding{172} If perturbations are insufficient, uploaded information leaks more data privacy to the attacker. \ding{174} If perturbations are excessive, task performance of the global model will be reduced. \ding{174} There is a lack of interpretation on location and degree of data leakage. \ding{175} Defensive strategies introduce much more training time. (b) \ding{172} FFT refers to Fast Fourier Transform(\cite{bu2023towards}). Frequency information from low to high is protected. \ding{173} Foreground noise is reduced according to the training process. \ding{174} Existing GIA defensive methods (the first row) are based on gradient distributions across model layers. Our proposed framework (the second row) separate task-dominant and privacy-sensitive regions of training data to generate noisy images for better balance between task performance and privacy protection. \ding{175} Benefitted from pretainable defense and region-coherent noise, our method achieves a better balance between task performance, privacy protection, and computational efficacy (point size).}
  \label{fig1}
\end{figure*}

To confront GIA, recent studies have proposed various defense mechanisms. These include differential privacy (\cite{abadi2016deep}), secure multi-party computation (\cite{zhang2022augmented}), and techniques to blur gradients, such as gradient sparsification (\cite{zhu2019deep, chang2024gradient}) and gradient clipping (\cite{geyer2017differentially, wei2021gradient}), alongside perturbations in data representation (\cite{sun2021soteria}), mixed methods (\cite{wang2024more}), and orthogonal subspace bayesian sampling (\cite{zhang2025censor}). However, these strategies cannot explain which gradients or image information are vulnerable to privacy leaks, resulting in either aggressive or inadequate perturbations. This further leads to a suboptimal balance between privacy safeguards and model efficacy. As privacy protection is a cornerstone of FL, the development of bespoke defensive measures that address the issue of image-level data leakage is imperative.

Drawing inspiration from the study on model inversion defense (\cite{wang2023privacy}), we are devoted to designing a defense strategy based on deep models, while avoiding considerable computational expenses incurred by direct mappings from gradients to images (\cite{wu2023learning}). We introduce a shadow model-based privacy protection framework endowed with the capability to interpret sensitive areas. This enables us to selectively introduce sample-level noises, thereby impairing the correlation between sensitive images and gradients or auxiliary information, while mitigating the reduction of task performance, as shown in \textcolor{SkyBlue}{Fig.} \ref{fig1}. (b).

In particular, we employ generative adversarial network (GAN) (\cite{karras2021alias}) to emulate behaviors of potential foes, calculating noise maps from its output. These maps are then equalized to obfuscate pathways to sensitive data, which is a proventive strategy against GIA through defense imitation (\cite{li2022auditing}). Furthermore, to optimize the trade-off between privacy protection and model performance, disturbances at task-critical areas are weakened. Considering the nature of GIA under strong assumptions, i.e., the effectiveness of this attack increases as training progresses (\cite{hatamizadeh2023gradient}), we calibrate the noise intensification accordingly. Momentum noise maps then act as guidance for regions to add noises in subsequent phases, easing the computational demands of the shadow model.

Based on two medical image datasets and models used in previous study (\cite{hatamizadeh2023gradient}), our empirical findings suggest that the proposed approach significantly improves the effectiveness of privacy defenses, simultaneously keeping the task performance similar to State-of-the-arts (SOTA) defense methods. Additionally, our method exhibits a versatile protective capacity against both model-based GIA (\cite{jeon2021gradient}) and optimization-based one (\cite{hatamizadeh2023gradient}). This shows the efficacy of the proposed framework in weaken the mapping from sensitive gradients or auxiliary information to images.

\section{Related Work}
\subsection{Gradient inversion attack}\label{Gradient inversion attack}
Gradient inversion attack (GIA) can be classified into two branches: optimization-based and model-based strategies. In the optimization-based branch (\cite{zhu2019deep}), dummy data and labels are first initialized, thereafter calculating the gradient of dummy data by backpropagating through a model. These dummy data and labels are then progressively updated by minimizing their divergence from the actual shared gradients. It has been theoretically demonstrated in iDLG (\cite{zhao2020idlg}) that, for classification tasks, the sign of gradients in fully connected (FC) layers can act as hints of labels. A novel technique for the iterative refinement of dummy inputs via closed-form solutions has been proposed in R-gap (\cite{citation-0}). The viability of GIA is probed by Huang et al. (\cite{huang2021evaluating}) under relaxed assumptions in FL. In Hatamizadeh et al. (\cite{hatamizadeh2023gradient}), it is shown that with BN statistics from a model and a template for dummy data, the strength of GIA can be considerably amplified in medical imagery scenarios. A series of regularizing terms, such as total variance regularization and group regularization, has been suggested in E2EGI (\cite{li2022e2egi}) to augment the fidelity of dummy data reconstruction when dealing with a large batch size.

Due to a lack of prior information, optimization-based methods can falter in accurate reconstruction of real images. The model-based strategy was first proposed in GIAS (\cite{jeon2021gradient}), similar to optimization-based methods that minimize dual gradients. The difference lies in the treatment of dummy data, which is the output of GAN rather than being directly optimized. This technique initiates by updating inputs of GAN, i.e., latent codes. Subsequently, there is a fine-tuning stage for pre-trained parameters of GAN. Extending from GIAS (\cite{jeon2021gradient}), in GGL (\cite{li2022auditing}), defensive methods are integrated in the process of GIA to imitate shared real gradients, which achieves great results across several defensive strategies. In GIFD (\cite{fang2023gifd}), based on a pre-trained GAN, intermediate features from the generator of GAN are updated layer-wise, and these features are bound by spherical regularization, all in pursuit of improving the fidelity of reconstructed images.

Unlike mainstream techniques, in Wu et al. (\cite{wu2023learning}) proposed a novel mapping schema constructed directly via multi-layer perceptrons between shared real gradients and dummy images. This is improved by a feature hashing algorithm to condense large gradients. Despite the straightforwardness of this method, its practicability is limited to attack within scenarios of low-resolution images and models of small parameter volume. It is due to the large volume of intermediate gradients that burgeon during backpropagation.

\subsection{Gradient inversion defense}\label{Gradient inversion defense}
To counteract GIA, existing defensive methods are manifold. Differential privacy (DP) (\cite{abadi2016deep, mcmahan2018learning}) adds noise into privacy-sensitive information, achieving a defense with theoretical assurance based on the privacy budget. However, it usually incurs substantial degradation in model performance. Secure multi-party computation (\cite{zhang2022augmented, bonawitz2017practical}) uploads and aggregates encrypted data, with decryption performed after download, which suffers from non-negligible computational costs. Gradient blur methods, such as gradient sparsification (GS) (\cite{zhu2019deep, chang2024gradient}), sets gradients of minimal amplitude to zero, while gradient clipping (GC) (\cite{geyer2017differentially, wei2021gradient}) clips gradients of maximum magnitude to a predefined threshold. Data representation perturbation, represented by Soteria (\cite{sun2021soteria}), performs disturbance in the representation from a learned FC layer to maximize reconstruction error. The fusion approach, like OUTPOST (\cite{wang2024more}), analyzes pivotal assumptions in GIA, and proposes to sparsify gradients with an empirical Fisher information lower than a threshold. For rest parts, perturbations with Gaussian noise are added to areas surpassing another threshold. The privacy risk is decided by the weight variance of each network layer, and the probability of perturbation decreases as training with weak assumptions progresses. Orthogonal subspace bayesian sampling is used in Censor (\cite{zhang2025censor}), where optimal orthogonal gradients related to training loss are searched for each batch.

Aforementioned defensive tactics fall short when it comes to explaining which gradients or image information are more prone to lead to privacy breaches. Consequently, their perturbations are either excessive or insufficient, leading to a compromise between privacy protection and model performance. Hence, our goal is to get the degree of privacy leakage at the sample level first. Upon this foundation, we can employ defensive strategies in a targeted manner. This will forestall the disclosure of data with high privacy risks, while minimizing the side effect on the task performance.

\section{Preliminary}\label{Preliminary}
\subsection{Federated learning}\label{Federated learning}
Assume $N$ clients participate in federated learning (FL) to collectively train a global model $f_0$, with parameters $\theta_0$. The objective function of FL (\cite{mcmahan2017communication}) is to minimize the loss on all local datasets based on the global model:  
\begin{flalign}
    &\text{min}_{\{\theta_0\}} \frac{1}{N} \sum_{i=1}^N L_i(D_i,\theta_0),&
\end{flalign}
where each client owns a local dataset denoted as $D_i={(x_j,y_j)}_{j=1}^{n_i}$, which comprises $n_i$ sample pairs, that is, image $x_i$ and label $y_j$. $\theta_0$ is obtained through a weighted aggregation of all local model parameters based on the number of local data. Although FL obviates the need to transmit original data, it still requires to send updated models or gradients, approximating model updates, to the server. This operation poses a risk of training data being reconstructed (\cite{huang2021evaluating}).

\begin{figure*}[ht]
  \begin{center}
  \includegraphics[width=2\columnwidth]{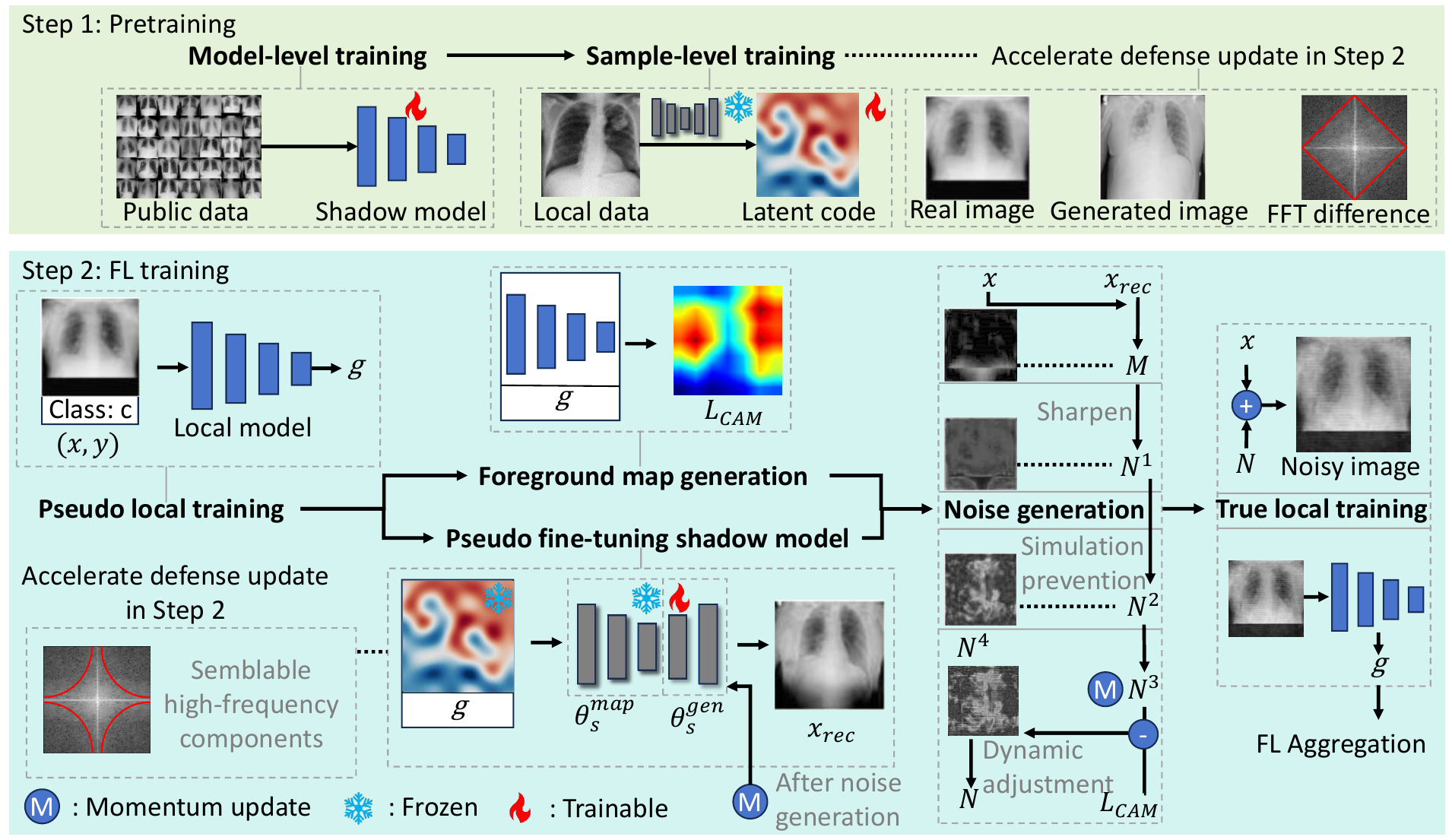}
  \end{center}
  \caption{The overall framework of our method. In the first pretraining stage, the shadow model and latent codes are updated to accelerate the FL local training by fitting low-frequency components (\cite{bu2023towards}).  In the second stage, local training is first performed to generate victim gradients $g$. The top branch produces foreground map $L_{CAM}$ to imply task-dominant areas. For the lower branch, model weights of the image generator in the shadow model $\theta^{gen}_s$ are fine-tuned to mimic potential adversaries with strong fitting ability on high-frequency components. Based on its outputs, we calculate defensive noises $N$ to protect privacy of training images, which are used for actual local training.}
  \label{fig2}
\end{figure*}
\subsection{Gradient inversion attack}\label{Gradient inversion attack pre}
In FL, an honest but curious server \footnote{The server wants to get private training data from clients, and do not harm the task performance of the global model.} can deploy GIA to reconstruct original data by leveraging uploaded model updates to approximate gradients, as illustrated in \textcolor{SkyBlue}{Fig.} \ref{fig1}. (a). For optimization-based GIA (\cite{zhu2019deep, zhao2020idlg, citation-0, huang2021evaluating}), optimization targets are dummy images and labels. This process is accomplished by minimizing the discrepancy between the shared real gradient $g_i$, a.k.a., target gradient, and the dummy gradient $\hat{g}$:
\begin{flalign}
    &\mathop{\arg\min}\limits_{\hat{x},\hat{y}}~\mathbb{D}(g_i,\hat{g}(\hat{x},\hat{y})),& \\
    &\text{s.t.}~\hat{g}(\hat{x},\hat{y}) = \frac{\partial L(f_0(\hat{x}),\hat{y})}{\partial \theta_0},&
\end{flalign}
where $\mathbb{D}$ is a distance function, typically chosen to be the L2 distance or cosine distance. $\hat{x},\hat{y}$ are dummy images and labels initialized randomly. $L$ is the loss function for the primary task. $f_0,\theta_0$ denote the global model and its parameters, respectively. 

For model-based GIA (\cite{li2022auditing, jeon2021gradient, fang2023gifd}), dummy images are generated using a pretrained GAN and optimization targets are parameters of GAN, input latent codes, and labels:
\begin{flalign}
  &\mathop{\arg\min}\limits_{\theta_{A}, z_0,\hat{y}}~\mathbb{D}(g_i,\hat{g}(\theta_{A}, z_0,\hat{y})),& \\
  &\text{s.t.}~\hat{g}(\theta_{A}, z_0,\hat{y}) = \frac{\partial L(f_0(f_A(z_0, \theta_{A})),\hat{y})}{\partial \theta_0},&
\end{flalign}
where $f_A, \theta_A$ are generative model and its parameters, $z_0$ is the input latent codes. While optimization-based GIA is applicable to wide scenarios, the computational time for convergence is high and it is more likely to be trapped in local optimums. In comparison, model-based GIA is more efficient but memory-consuming due to the usage of GAN. 

To better design a targeted defense framework, we need to know several key attributes of GIA considering its effectiveness:

\textbf{Statistics of batch normalization layers improves fidelity of reconstructed images.} In existing GIA literatures (\cite{huang2021evaluating, hatamizadeh2023gradient, li2022e2egi}), it is found that if BN statistics are uploaded along with gradients, an honest but curious server can enhance the efficacy of attack significantly by incorporating BN regularization terms during GIA. This is because momentum mean and variance in BN layers offer pixel-wise distribution and contrast which are key information related to image details. Since medical images are less diverse in a single dataset compared with natural images, their BN statistics are thus more stable, which greatly increases privacy leakage of patients. To defend against GIA, it is insufficient to only consider perturbing gradients. Therefore, in this work, we propose to break the mapping from gradients or auxiliary information to images, disabling the BN regularization term of GIA.

\textbf{The extent of privacy leakage increases with training dynamic.} Due to a decreased trend of gradient magnitude, a gradually decreased defense strength is proposed in OUTPOST according to the GIA trend (\cite{wang2024more}). However, with BN regularization, it is found that the GIA trend is on the contrary (\cite{hatamizadeh2023gradient}), since BN statistics become more accurate during training. Therefore, we design a defense technique with gradually increased strength. Such technique can also be treated as regularization for task model, reducing overfitting due to a limited amount of data in medical imaging.

\textbf{The number of samples belonging to each class affect GIA strength.} For the classification task, labels of images can be exactly reconstructed as long as label repetition rate is low (\cite{zhu2019deep}). However, for medical image diagnosis, the number of classes are usually small, thus increase the difficulty of GIA. Besides, a non-IID setting in FL also make the estimation part harder compared to an IID setting.

\textbf{Training from a pretrained model reveals more privacy.} If a task model is trained from pretrained weights, sensitivity of gradients to private data will be higher in ealier stages (\cite{hatamizadeh2023gradient}). Besides, new BN statistics will converge faster and inter-class features are more separated, both of which facilitate better GIA. Therefore, a pretained task model is trained in our experiment to mimic vulnerable target gradients and BN statistics.

\textbf{Multiple iterations in local training blur target gradients.} In implementation of GIA methods under the centralized learning scenarios (\cite{zhao2020idlg, li2022e2egi}), gradients updated from one mini-batch serve as targets to attack. However, it has been validated that if target gradients are updated from several iterations (one for a mini-batch) and local rounds(\cite{xu2022agic, zhu2023surrogate}), the strength of GIA will be weakened due to the equalization of sample gradients. In our experiments, we set different iterations for clients to represent various extents of privacy leakage, in which a client with one data sample is used to test the lower bound of defense.

\section{Methods}\label{methods}
In this section, we introduce a privacy preservation framework built upon the shadow model to defense GIA under strong assumptions.
\subsection{Framework overview}\label{Framework overview}
\textcolor{SkyBlue}{Fig.} \ref{fig2} shows our proposed privacy-protection framework based on the shadow model. It includes two main steps: pre-training and federated training. To imitate potential attack of adversaries, we utilize a GAN model, renowned for exceptional generative power, as our shadow model, denoted as $f_s$. By pre-training $\theta_s$ alongside latent codes for each image input $z$, the cost of fine-tuning the shadow model in the next step can be diminished, which further improves concurrency of the whole federated training.

During the federated training phase, client $i$ first performs a pseudo-update on its local model $f_i$ to yield information required for pseudo fine-tuning of the shadow model. Reconstructed outputs $x_s$ from the pseudo fine-tuned shadow model serves as inputs for the computation of noise maps. Throughout this procedure, we utilize foreground activation maps $L_{CAM}$, obtained from the pseudo-updated local model, to mitigate disturbance in foreground regions of noise maps. This strategy avoids inappropriately compromising performance of the primary task. Noise maps are then added onto original data, following which the real update is performed for the local model. The result gradients and statistical information are uploaded to the server for global aggregation. As for the shadow model, a momentum-based real update is done, since attack effectiveness from adversaries in real-world scenarios might be limited (\cite{wang2024more, hatamizadeh2023gradient}). Since the goal of updating the shadow model is to determine key regions to add noises, and the computational expense of fine-tuning the shadow model is high, this strategy also prevents the necessity for redundant fine-tuning in subsequent iterations.

\subsection{Pretraining}\label{Pretraining}
\textbf{Model-level training: Preparation for real-time defense in FL training.} For gradient inversion defense, a critical balance must be struck not only between task performance and privacy but also considering the computational cost. Our goal is to fine-tune the shadow model efficiently during federated training to counteract real-time attack capabilities of potential strong adversaries. To this end, prior to federated training, we pre-train the weights $\theta_s$ of the shadow model on a public dataset of a similar task type. In our experiments, we adopt the pre-training strategy of StyleGAN3 (\cite{karras2021alias}).

\textbf{Sample-level training: Further acceleration with low-frequency components of images fitted.} Once the shadow model has been pre-trained, it acquires the capability to generate style features related to the data type. However, it is not adept at capturing specific details of local data. To further accelerate the fine-tuning of the shadow model during federated training, the ability of the shadow model to fit low-frequency components of data can be enhanced. Specifically, we fix pre-trained parameters of the shadow model, while randomly initializing latent codes $z_j$ for each sample $j$. We then pre-train $z_j$ by minimizing the discrepancy between original images $x$ and reconstructed images $x_s$:  $\left \| x-x_s \right \| $.

\begin{figure}[htbp!]
    \begin{center}
    \includegraphics[width=0.9\columnwidth]{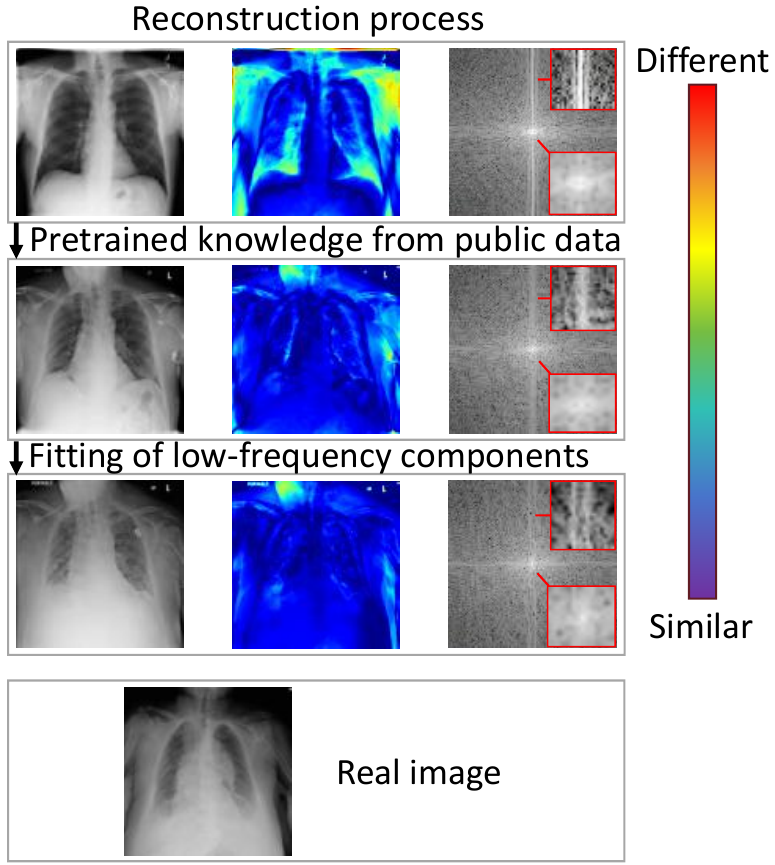}
    \end{center}
    \caption{The updating process of reconstructed images at the pre-training stage for latent codes. In the first three rows, three columns represent original images, reconstructive error maps, frequency spectrum of reconstructive error maps, in sequence.}
    \label{fig3}
\end{figure}
In practical applications, it is difficult for adversaries to estimate precise style information for each image (\cite{li2022auditing, jeon2021gradient}). Consequently, an early-stop strategy is employed when updating $z$ to minimize the additional computational time. The updating process, depicted in \textcolor{SkyBlue}{Fig.} \ref{fig2} and \textcolor{SkyBlue}{Fig.} \ref{fig3}, demonstrates that only low-frequency components of images, such as the overall structure, are accurately reconstructed when solely updating $z$. However, discrepancies remain between reconstructed and real images, particularly at edges and other fine details.

\subsection{FL training}\label{FL training}
\textbf{Pseudo local training: Generating victim gradients to be protected.} To reflect sensitive privacy in GIA scenarios, it is crucial to generate gradients of local training images. However, unlike traditional methods directly modifying gradients based on their statistics, we only treat these gradients as inputs in later steps since large absolute gradients cannot intuitively reflect sensitive privacy. In \textcolor{SkyBlue}{Fig.} \ref{fig1}. (b) and appendix, we demonstrate that even with similar distribution of overall gradients, these methods vary greatly in privacy protection.

\textbf{Pseudo fine-tuning shadow model: Imitating real-time attack strength considering sensitive privacy.} In the standard paradigm of FL, all clients are required to upload their model updates in the form of gradients after local training. This protocol, however, raises concerns for privacy breaches of training data under GIA. If we can simulate real-time attack capabilities of adversaries and apply tailored perturbations for protection at the sample level, the privacy leakage risk will be mitigated. To implement this, we first fine-tune the shadow model on a per-sample basis, using the pre-trained shadow model and latent codes. Subsequently, we use images reconstructed by the shadow model as a guide to introduce noise into original images.

During local training, all clients first pseudo-train their local models, yielding updated gradients and BN statistics to serve as simulation of potentially leaked information. Following this, we perform pseudo update for the generator of the shadow model, in order to imitate an almost optimal adversary. Specifically, we freeze the pre-trained sample-level latent codes and weights of latent mappers in the shadow model $\theta_s^{map}$ to ensure that major features of reconstructed images are more stable (\cite{karras2021alias}). Based on this, reconstructed images are fed back to the untrained local model to compute the gradient inversion loss:
\begin{flalign}
    &\mathop{\arg\min}\limits_{\theta_s^{gen}}~L_\text{shadow} = \mathbb{D}+R_\text{TV}+R_\text{BN}+R_\text{L2}+L_\text{MSE},&
\end{flalign}
where $\mathbb{D}$ represents the distance function between reconstructed gradients and actual gradients. $R_{TV}$ is the total variance regularization of reconstructed images. $R_{BN}$ is the regularization of BN statistics. $R_{L2}$ refers to the L2 regularization of reconstructed images (\cite{hatamizadeh2023gradient, fang2023gifd}). $L_{MSE}$ is the mean squared error loss between reconstructed and actual images, which accelerates convergence of the shadow model. $\theta_s^{gen}$ represents weights of the image generator in the shadow model.

After completing pseudo fine-tuning of the shadow model, we obtain parameters $\theta'_s$. Based on these, we first generate image noise, ensuring that the defense is always stronger than potential attacks at early stages. Then, we perform the momentum actual update on the shadow model. This simulates potential attacks on the current training progress and serves as a guide for crucial regions to add noises in subsequent rounds:
\begin{flalign}
    &\theta_s=\alpha_{ema}^{shadow} \cdot \theta_s+\left(1-\alpha_{ema}^{shadow}\right) \cdot \theta'_s,&
\end{flalign}
where $\alpha_{ema}^{shadow}$ denotes the hyperparameter coefficient for momentum updating.

Since the shadow model primarily functions as an indicator of sensitive areas during image defense, rather than directly determining the intensity of noises, we set a terminal round for shadow updates, denoted as $r_{shadow}$, to mitigate computational expenses. This reduction in computational overhead does not compromise the efficacy of the defense, as confirmed in our experiments.

\textbf{Foreground map generation: Acquiring task-dominant regions to be preserved in noise generation.} Traditional noise-injecting methods are often quite detrimental to the task performance (\cite{abadi2016deep}). To mitigate this side effect, it is imperative to minimize noises in task-relevant areas. We employ Grad-CAM++ (\cite{chattopadhay2018grad}) to first obtain pixel-level saliency maps $L^c$ from an intermediate layer of the network due to its balanced representation of semantic and spatial information.

Based on pixel-level saliency maps, areas with the highest activation values, denoted as $\mathcal{T}_{CAM}$, are selected to form a binary mask $M_{CAM}$. This mask represents the salient regions that the task model focuses on during task processing, and is used to generate the foreground activation map: 
\begin{flalign}
    &L_{CAM} =\sigma\left(\frac{M_{CAM} \cdot L^c}{T}\right),& \\
    &\text{s.t.}~M_{CAM} =\operatorname{topk}\left(L_{i j}^c, \mathcal{T}_{C A M}\right).&
\end{flalign}

\noindent\textbf{Noise generation: Reducing side effect for task adaptively and concentrating on privacy-sensitive regions.} After pseudo update of the shadow model, reconstructed images are obtained. Firstly, we calculate a pixel-wise MSE map between original images and reconstructed ones: $M=\left\{\left\|x^i-x_{\text {rec }}^i\right\|_2^2\right\}_{i=1}^{H \cdot W}$. Subsequently, this map undergoes sharpening and normalization to focus its key regions on where is prone to be reconstructed:
\begin{flalign}
    &N^1=\frac{1}{\sigma\left(\frac{M}{T}\right)},& \\
    &\text{s.t.}~\sigma\left(\frac{M_i}{T}\right)=\frac{e^{\frac{M_i}{T}}}{\sum_{j=1}^{H \cdot W} e^{\frac{M_j}{T}}},&
\end{flalign}
where $T$ is the temperature coefficient of the softmax function.

Since we have employed $L_{MSE}$ to accelerate convergence during fine-tuning of the shadow model, some reconstructed areas are with excessive precision. However, this does not imply that all such areas require a proportionately higher level of privacy protection compared to others. Moreover, if noises added to certain regions are excessive while being insufficient in others, the visual robustness of noisy images is compromised. This could cause a defect that allows adversaries to discern the implementation of defense strategies during model training more easily, and to simulate these defenses (\cite{li2022auditing}). Consequently, we apply histogram equalization (\cite{garg2017comparative}) and normalization to the initial relative noise $N^1$:
\begin{flalign}
    &N^2=\sigma\left(G\left(N^1\right)\right),& \\
    &\text{where}~G(z)=\frac{N_{gray}-1}{b-a} \sum_{i=a}^{z}p(r_i) ,&
\end{flalign}
where $N_{gray}$ represents the total number of gray levels, i.e., 256. $a,b$ denote the range of image grayscale values, i.e., 0 and 255, respectively. $p(r)$ is a normalized histogram probability of the grayscale level $r$. $\sigma$ denotes the softmax function. Unless the mapping function is known in advance, this process can be difficult to reverse. The visualization of $N^2$, as shown in \textcolor{SkyBlue}{Fig.} \ref{fig5}, indicates that regions with smaller reconstruction errors receive stronger noises, appearing darker in the noise map $N^2$, and that there is not an issue of excessive local noise.

During training, rapid convergence of the shadow model might result in significant discrepancies between noise maps from successive rounds. To avoid training instability caused by this phenomenon, we use a momentum update strategy for $N^2$:
\begin{flalign}
    &N^3=\alpha_{ema}^{noise} \cdot N^3+\left(1-\alpha_{ema}^{noise}\right) \cdot N^2,&
\end{flalign}
where $\alpha_{ema}^{noise}$ denotes the coefficient hyperparameter. $N^3$ is initialized with the first $N^2$.

\begin{figure}[htbp!]
  \begin{center}
  \includegraphics[width=\columnwidth]{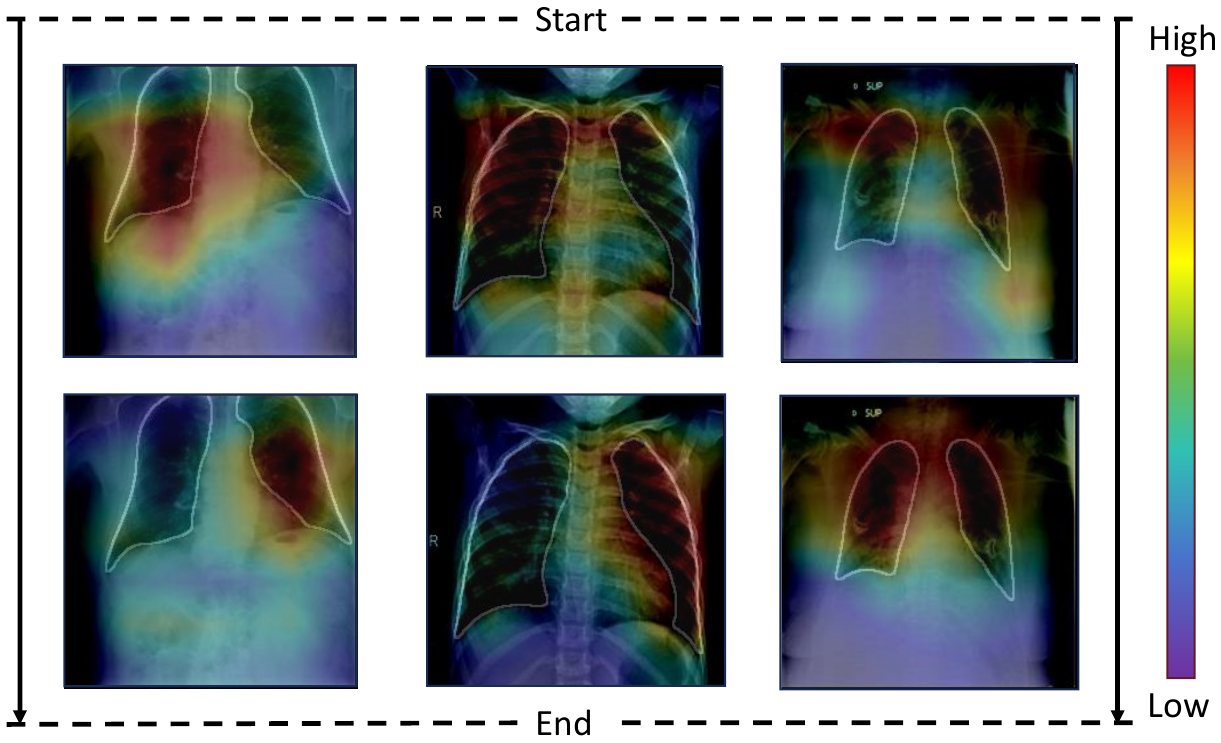}
  \end{center}
  \caption{Evolution of foreground map during training. Core concerns of task models are moved from random regions to foreground ones gradually. For the ChestXRay dataset (\cite{chowdhury2020can, rahman2021exploring}), most foreground regions locate inside or around the border of the entire lung, corresponding to lesions related to classification results (\cite{yue2023ergpnet}).}
  \label{fig4}
\end{figure}

The capability of the network to focus on key areas, related to the main task, becomes more and more accurate as training goes on, as illustrated in \textcolor{SkyBlue}{Fig.} \ref{fig4}. Consequently, the impact on foreground pixels within the noise map is reduced in accordance with the current training epoch:
\begin{flalign}
    &N^4=N^3-\alpha_{C A M} \operatorname{sign}\left(N^3\right) \cdot L_{C A M},& \\
    &\text{s.t.}~\alpha_{C A M}=\min \left(\alpha_{C A M}^{\max }, \max \left(\alpha_{C A M}^{\min }, \frac{r}{R}\right)\right),&
\end{flalign}
where $\alpha_{C A M}^{\max },\alpha_{C A M}^{\min}$ are coefficient hyperparameters that determine the maximum and minimum influence of $L_{C A M}$. $\operatorname{sign}$ is a sign function.

After determining relative noises, we need to further ascertain the absolute magnitude of noises. Considering the fact that the intensity of GIA under strong assumptions increases progressively at training (\cite{hatamizadeh2023gradient}), the overall scale of noises is similarly adjusted based on the training epoch:
\begin{flalign}
    &N=\left|\frac{\max (x)}{\max \left(N^{4}\right)} w_{N}\right| N^{4},& \\
    &\text{s.t.}~w_{N}=\alpha_{N} e^{\frac{r}{R}},&
\end{flalign}
where $\alpha_{N}$ is the hyperparameter that controls the absolute amount of noises. 

\begin{figure*}[htbp!]
    \begin{center}
    \includegraphics[width=2\columnwidth]{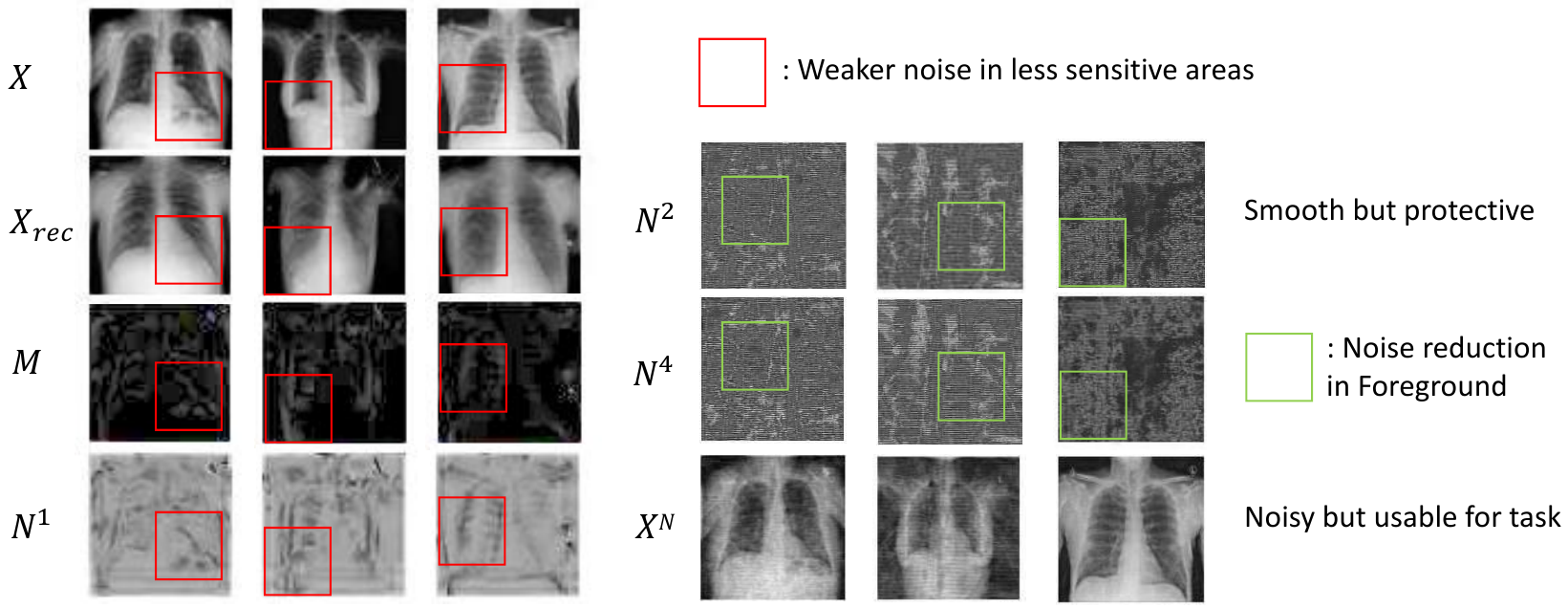}
    \end{center}
    \caption{Visualization of various types of noise and images. $X, X_{rec}, X^N$ are original images, reconstructed images, and noisy images, respectively. $M$ is the error map between $X$ and $X_{rec}$. $N^1$ is the initial relative noise. $N^2$ is generated by histogram equalization based on $N^1$. The transforming process from $N^2$ to $N^4$ includes history update and foreground noise reduction.}
    \label{fig5}
\end{figure*}

\noindent\textbf{True local training: Generating protected gradients.} Pixel-level noises $N$ are ultimately applied to original images as additive noise, and noised images $X^N$ are utilized to actually update the local model, resulting in a protected gradient for FL aggregation. This leads to a fine-grained and interpretable effect of privacy preservation, as shown in \textcolor{SkyBlue}{Fig.} \ref{fig5}.

\section{Experimental Setup and Results}\label{Experimental Setup and Results}
\subsection{Experimental setting}
We use two medical image datasets, i.e., ChestXRay\footnote{\url{https://www.kaggle.com/datasets/tawsifurrahman/covid19-radiography-database/data}} (\cite{chowdhury2020can, rahman2021exploring}) and EyePACS\footnote{\url{https://zenodo.org/records/5793241}} (\cite{devente23airogs}), for local model training in FL, separately. The former includes one-channel X-ray data, while the latter encompasses color fundus images. All images are resized to a uniform resolution of $224 \times 224$ pixels and normalized to a zero mean and a unit variance. Our dataset partitioning is same as the baseline established by the previous work (\cite{hatamizadeh2023gradient}), involving 9 clients with the last one allocated one training image to represent the upper bound of privacy leakage.

Alike the setting in Hatamizadeh et al. (\cite{hatamizadeh2023gradient}), we use ResNet18 (\cite{cardoso2022monai}) pre-trained on ImageNet as initial local models. Optimizer for task model is SGD with learning rate 1e-2 and cross-Entropy (CE) is used as loss. Batch sizes for clients are configured as follows: 4 for client 1 to client 4, 8 for client 5 to client 8, and 1 for client 9. Local round and global round for FL are set to 1 and 100, respectively. All experiments are implemented on a Nvidia GeForce RTX 3090, utilizing the PyTorch framework (\cite{imambi2021pytorch}).

For shadow models, pretraining is conducted based on the training strategy proposed in StyleGAN3 (\cite{karras2021alias}). The optimizers for shadow models and latent variables are Adam, with initial learning rates set at 1e-3 and 1e-3, respectively. These rates are further adjusted using a step scheduler. Optimization epochs for each local round are designated as 5 and 500 for shadow models and latent variables. The global round for updates of the shadow model is set at 20. An early stopping round of 5 is set for updates of latent variables. For the fine-tuning of shadow models, learning rates are set at 1e-3 for client 1 to client 4, 2e-3 for client 5 to client 8, and 1e-4 for client 9. The default epoch of fine-tuning for shadow models is 20. For Grad-CAM++, we set the maximum percentile value $\mathcal{T}_{CAM}$, to 30\%. In our experiments, the target foreground is defined by a bounding box encompassing the top 30\% of pixels based on Grad-CAM++ values. Maximum and minimum values $\alpha_{CAM}^{max}, \alpha_{CAM}^{min}$ deciding the CAM influence are 0.1 and 0.5, respectively. Hyperparameters for momentum updates of the shadow model and relative noise, $\alpha_{ema}^{shadow}, \alpha_{ema}^{noise}$, are set at 0.5 and 0.9, separately. The coefficient hyperparameter for absolute noise, $\alpha_N$, is defaulted to 0.19.

\subsection{Results}
\subsubsection{State-of-the-Art (SOTA) Comparison}
\begin{table*}[]
  \centering
  \caption{Comparison of our method with SOTA defense methods against model-based GIA. F1 represents task performance. For defense metrics, i.e., MSE, PSNR, LPIPS, SSIM, are all the mean across 9 clients and 5 FL rounds, i.e., 1, 25, 50, 75, 100. Numbers in brackets are standard deviation of these defensive metrics across these 5 FL rounds. For each reconstructed image, we calculate metrics to the most similar real image since the gradient is derived from a global round. Target regions are decided by bounding boxes based on Grad-CAM++. Bold numbers denotes the best result among all methods. For statistical meaning, $\dag$ means $p<<0.05$ in the Wilcoxon signed rank test for comparison between FedAvg and a specific method, while $\ddag$ is for comparison between a specific method and all other SOTA methods.  `DP', `GS', `GC' are abbreviation of differential privacy, gradient sparsification, and gradient clipping.}
  \resizebox{2\columnwidth}{!}{
      \begin{tabular}{clllllllllll}
          \hline
          &
          \multicolumn{1}{l}{Dataset} &
            \multicolumn{5}{l}{ChestXRay} &
            \multicolumn{5}{l}{EyePACS} \\ \cmidrule(r){3-7}\cmidrule(r){8-12}
            &
          \multicolumn{1}{l}{Method} &
            \multicolumn{1}{l}{F1↑} &
            \multicolumn{1}{l}{MSE↑} &
            \multicolumn{1}{l}{PSNR↓} &
            \multicolumn{1}{l}{LPIPS↑} &
            \multicolumn{1}{l}{SSIM↓} &
            \multicolumn{1}{l}{F1↑} &
            \multicolumn{1}{l}{MSE↑} &
            \multicolumn{1}{l}{PSNR↓} &
            \multicolumn{1}{l}{LPIPS↑} &
            SSIM↓ \\ \cline{2-12}
            &
          \multicolumn{1}{l}{FedAvg} &
            \multicolumn{1}{l}{0.978} &
            \multicolumn{1}{l}{0.035 (0.012)} &
            \multicolumn{1}{l}{15.09 (1.43)} &
            \multicolumn{1}{l}{0.385 (0.051)} &
            \multicolumn{1}{l}{0.486 (0.057)} &
            \multicolumn{1}{l}{0.870} &
            \multicolumn{1}{l}{0.068 (0.030)} &
            \multicolumn{1}{l}{12.71 (2.37)} &
            \multicolumn{1}{l}{0.520 (0.093)} &
            0.411 (0.063) \\&
          \multicolumn{1}{l}{SCAFFOLD} &
            \multicolumn{1}{l}{0.976} &
            \multicolumn{1}{l}{0.034 (0.013)} &
            \multicolumn{1}{l}{15.16 (1.57)} &
            \multicolumn{1}{l}{0.380 (0.051)} &
            \multicolumn{1}{l}{0.496 (0.060)} &
            \multicolumn{1}{l}{0.875} &
            \multicolumn{1}{l}{0.069 (0.028)} &
            \multicolumn{1}{l}{12.64 (2.05)} &
            \multicolumn{1}{l}{0.523 (0.094)} &
            0.406 (0.063) \\&
          \multicolumn{1}{l}{DP} &
            \multicolumn{1}{l}{0.860} &
            \multicolumn{1}{l}{0.037 (0.010)} &
            \multicolumn{1}{l}{14.95 (1.38)} &
            \multicolumn{1}{l}{0.438 (0.054)$\dag$} &
            \multicolumn{1}{l}{0.464 (0.059)$\dag$} &
            \multicolumn{1}{l}{0.798} &
            \multicolumn{1}{l}{0.070 (0.032)$\dag$} &
            \multicolumn{1}{l}{12.63 (2.33)} &
            \multicolumn{1}{l}{0.505 (0.095)} &
            0.409 (0.055) \\&
          \multicolumn{1}{l}{GS} &
            \multicolumn{1}{l}{0.968} &
            \multicolumn{1}{l}{0.043 (0.016)} &
            \multicolumn{1}{l}{14.43 (1.64)} &
            \multicolumn{1}{l}{0.390 (0.049)} &
            \multicolumn{1}{l}{0.437 (0.070)} &
            \multicolumn{1}{l}{0.861} &
            \multicolumn{1}{l}{0.067 (0.030)} &
            \multicolumn{1}{l}{12.83 (2.39)} &
            \multicolumn{1}{l}{0.520 (0.090)} &
            0.410 (0.055) \\&
          \multicolumn{1}{l}{GC} &
            \multicolumn{1}{l}{0.848} &
            \multicolumn{1}{l}{0.043 (0.012)} &
            \multicolumn{1}{l}{14.44 (1.38)} &
            \multicolumn{1}{l}{0.384 (0.053)} &
            \multicolumn{1}{l}{0.440 (0.066)} &
            \multicolumn{1}{l}{0.693} &
            \multicolumn{1}{l}{0.050 (0.024)} &
            \multicolumn{1}{l}{13.83 (2.43)} &
            \multicolumn{1}{l}{0.525 (0.104)} &
            0.391 (0.084) \\&
          \multicolumn{1}{l}{Soteria} &
            \multicolumn{1}{l}{0.973} &
            \multicolumn{1}{l}{0.034 (0.012)} &
            \multicolumn{1}{l}{15.05 (1.40)} &
            \multicolumn{1}{l}{0.386 (0.052)} &
            \multicolumn{1}{l}{0.490 (0.057)} &
            \multicolumn{1}{l}{0.862} &
            \multicolumn{1}{l}{0.071 (0.032)} &
            \multicolumn{1}{l}{12.54 (2.35)} &
            \multicolumn{1}{l}{0.535 (0.089)} &
            0.405 (0.060) \\&
          \multicolumn{1}{l}{OUTPOST} &
            \multicolumn{1}{l}{0.972 } &
            \multicolumn{1}{l}{0.034 (0.012)} &
            \multicolumn{1}{l}{15.16 (1.53)} &
            \multicolumn{1}{l}{0.383 (0.049)} &
            \multicolumn{1}{l}{0.491 (0.059)} &
            \multicolumn{1}{l}{0.862} &
            \multicolumn{1}{l}{0.069 (0.031)} &
            \multicolumn{1}{l}{12.67 (2.27)} &
            \multicolumn{1}{l}{0.530 (0.098)} &
            0.403 (0.069) \\&
          \multicolumn{1}{l}{Censor} &
            \multicolumn{1}{l}{0.810} &
            \multicolumn{1}{l}{0.035 (0.010)} &
            \multicolumn{1}{l}{15.08 (1.32)} &
            \multicolumn{1}{l}{0.381 (0.049)} &
            \multicolumn{1}{l}{0.496 (0.059)} &
            \multicolumn{1}{l}{0.728} &
            \multicolumn{1}{l}{0.069 (0.020)} &
            \multicolumn{1}{l}{12.70 (2.25)} &
            \multicolumn{1}{l}{0.526 (0.106)} &
            0.403 (0.076) \\ \cline{2-12}
            \multirow{-10}{*}{\rotatebox{90}{Whole Image}} &
          \multicolumn{1}{l}{\textbf{Ours}} &
            \multicolumn{1}{l}{0.967} &
            \multicolumn{1}{l}{\textbf{0.102 (0.018)}$\ddag$} &
            \multicolumn{1}{l}{\textbf{11.36 (0.93)}$\ddag$} &
            \multicolumn{1}{l}{\textbf{0.642 (0.040)}$\ddag$} &
            \multicolumn{1}{l}{\textbf{0.286 (0.072)}$\ddag$} &
            \multicolumn{1}{l}{0.861} &
            \multicolumn{1}{l}{\textbf{0.125 (0.031)}$\ddag$} &
            \multicolumn{1}{l}{\textbf{9.93 (1.47)}$\ddag$} &
            \multicolumn{1}{l}{\textbf{0.714 (0.057)}$\ddag$} &
            {\textbf{0.245 (0.071)}$\ddag$} \\ \hline
          &
          \multicolumn{1}{l}{Dataset} &
            \multicolumn{5}{l}{ChestXRay} &
            \multicolumn{5}{l}{EyePACS} \\ \cmidrule(r){3-7}\cmidrule(r){8-12}
            &
          \multicolumn{1}{l}{Method} &
            \multicolumn{1}{l}{F1↑} &
            \multicolumn{1}{l}{MSE↑} &
            \multicolumn{1}{l}{PSNR↓} &
            \multicolumn{1}{l}{LPIPS↑} &
            \multicolumn{1}{l}{SSIM↓} &
            \multicolumn{1}{l}{F1↑} &
            \multicolumn{1}{l}{MSE↑} &
            \multicolumn{1}{l}{PSNR↓} &
            \multicolumn{1}{l}{LPIPS↑} &
            SSIM↓ \\ \cline{2-12}
            &
          \multicolumn{1}{l}{FedAvg} &
            \multicolumn{1}{l}{0.978} &
            \multicolumn{1}{l}{0.015 (0.007)} &
            \multicolumn{1}{l}{18.76 (1.51)} &
            \multicolumn{1}{l}{0.125 (0.027)} &
            \multicolumn{1}{l}{0.788 (0.034)} &
            \multicolumn{1}{l}{0.870 } &
            \multicolumn{1}{l}{0.038 (0.020)} &
            \multicolumn{1}{l}{15.47 (2.76)} &
            \multicolumn{1}{l}{0.335 (0.060)} &
            0.587 (0.047) \\&
          \multicolumn{1}{l}{SCAFFOLD} &
            \multicolumn{1}{l}{0.976} &
            \multicolumn{1}{l}{0.023 (0.009)} &
            \multicolumn{1}{l}{16.99 (1.61)} &
            \multicolumn{1}{l}{0.190 (0.035)} &
            \multicolumn{1}{l}{0.672 (0.046)} &
            \multicolumn{1}{l}{0.875 } &
            \multicolumn{1}{l}{0.039 (0.019)} &
            \multicolumn{1}{l}{15.31 (2.39)} &
            \multicolumn{1}{l}{0.338 (0.066)} &
            0.585 (0.048) \\&
          \multicolumn{1}{l}{DP} &
            \multicolumn{1}{l}{0.860} &
            \multicolumn{1}{l}{0.017 (0.005)} &
            \multicolumn{1}{l}{18.81 (1.63)} &
            \multicolumn{1}{l}{0.154 (0.032)$\dag$} &
            \multicolumn{1}{l}{0.778 (0.035)$\dag$} &
            \multicolumn{1}{l}{0.798} &
            \multicolumn{1}{l}{0.053 (0.023)$\dag$} &
            \multicolumn{1}{l}{13.87 (2.41)$\dag$} &
            \multicolumn{1}{l}{0.392 (0.048)$\dag$} &
            {0.571 (0.046)$\dag$} \\&
          \multicolumn{1}{l}{GS} &
            \multicolumn{1}{l}{0.968} &
            \multicolumn{1}{l}{0.020 (0.007)$\dag$} &
            \multicolumn{1}{l}{17.83 (1.57)$\dag$} &
            \multicolumn{1}{l}{0.142 (0.028)} &
            \multicolumn{1}{l}{0.751 (0.039)$\dag$} &
            \multicolumn{1}{l}{0.861} &
            \multicolumn{1}{l}{0.039 (0.020)} &
            \multicolumn{1}{l}{15.53 (2.73)} &
            \multicolumn{1}{l}{0.343 (0.057)$\dag$} &
            0.585 (0.038) \\&
          \multicolumn{1}{l}{GC} &
            \multicolumn{1}{l}{0.848} &
            \multicolumn{1}{l}{0.020 (0.006)} &
            \multicolumn{1}{l}{17.96 (1.55)$\dag$} &
            \multicolumn{1}{l}{0.153 (0.030)$\dag$} &
            \multicolumn{1}{l}{0.758 (0.041)$\dag$} &
            \multicolumn{1}{l}{0.693 } &
            \multicolumn{1}{l}{0.032 (0.016)} &
            \multicolumn{1}{l}{16.04 (2.75)} &
            \multicolumn{1}{l}{0.385 (0.082)$\dag$} &
            {0.570 (0.063)$\dag$} \\&
          \multicolumn{1}{l}{Soteria} &
            \multicolumn{1}{l}{0.973} &
            \multicolumn{1}{l}{0.015 (0.006)} &
            \multicolumn{1}{l}{18.86 (1.50)} &
            \multicolumn{1}{l}{0.120 (0.026)} &
            \multicolumn{1}{l}{0.781 (0.032)$\dag$} &
            \multicolumn{1}{l}{0.862} &
            \multicolumn{1}{l}{0.042 (0.021)$\dag$} &
            \multicolumn{1}{l}{14.99 (2.49)$\dag$} &
            \multicolumn{1}{l}{0.359 (0.059)$\dag$} &
            0.581 (0.044) \\&
          \multicolumn{1}{l}{OUTPOST} &
            \multicolumn{1}{l}{0.972} &
            \multicolumn{1}{l}{0.014 (0.004)} &
            \multicolumn{1}{l}{18.93 (1.41)} &
            \multicolumn{1}{l}{0.134 (0.025)$\dag$} &
            \multicolumn{1}{l}{0.788 (0.036)} &
            \multicolumn{1}{l}{0.862} &
            \multicolumn{1}{l}{0.043 (0.021)$\dag$} &
            \multicolumn{1}{l}{14.81 (2.44)$\dag$} &
            \multicolumn{1}{l}{0.371 (0.061)$\dag$} &
            {0.572 (0.051)$\dag$} \\& 
          \multicolumn{1}{l}{Censor} &
            \multicolumn{1}{l}{0.810} &
            \multicolumn{1}{l}{0.020 (0.006)$\dag$} &
            \multicolumn{1}{l}{17.81 (1.25)$\dag$} &
            \multicolumn{1}{l}{0.160 (0.030)$\dag$} &
            \multicolumn{1}{l}{0.664 (0.045)$\dag$} &
            \multicolumn{1}{l}{0.728} &
            \multicolumn{1}{l}{0.040 (0.014)} &
            \multicolumn{1}{l}{15.23 (2.63)} &
            \multicolumn{1}{l}{0.340 (0.080)} &
            0.581 (0.062) \\ \cline{2-12}
            \multirow{-10}{*}{\rotatebox{90}{Target Region}} &
          \multicolumn{1}{l}{\textbf{Ours}} &
            \multicolumn{1}{l}{0.967} &
            \multicolumn{1}{l}{\textbf{0.071 (0.012)}$\ddag$} &
            \multicolumn{1}{l}{\textbf{12.91 (0.94)}$\ddag$} &
            \multicolumn{1}{l}{\textbf{0.418 (0.044)}$\ddag$} &
            \multicolumn{1}{l}{\textbf{0.507 (0.054)}$\ddag$} &
            \multicolumn{1}{l}{0.861} &
            \multicolumn{1}{l}{\textbf{0.086 (0.022)}$\ddag$} &
            \multicolumn{1}{l}{\textbf{11.68 (1.55)}$\ddag$} &
            \multicolumn{1}{l}{\textbf{0.507 (0.046)}$\ddag$} &
            \textbf{0.465 (0.051)$\ddag$} \\ \hline
          \end{tabular}
  }
  \label{table 1}
\end{table*}

We employ both model-based GIA (\cite{jeon2021gradient}) and optimization-based GIA (\cite{hatamizadeh2023gradient}) to assess the defensive capability of our proposed framework, with specific GIA details in section \ref{Gradient inversion attack pre}. Comparisons between results of SOTA defensive methods and our proposed framework are shown in Table \ref{table 1} and Table \ref{table 2}. 'FedAvg' is the standard FL training framework without privacy protection (\cite{mcmahan2017communication}). `SCAFFOLD' introduces control variates for the client-drift problem in FL (\cite{karimireddy2020scaffold}). `DP', `GS', `GC' are abbreviation of differential privacy, gradient sparsification, and gradient clipping. We use `F1' as an indicator for task performance. Meanwhile, Mean Square Error (MSE), Peak Signal to Noise Ratio (PSNR), Learned Perceptual Image Patch Similarity (LPIPS), Structural Similarity Index Measure (SSIM) are used to test defensive capabilities of SOTA methods. 

Table \ref{table 1} presents the model performance on two medical datasets and defensive results of whole images and foreground regions under model-based GIA. For methods without defense, i.e., `FedAvg' and `SCAFFOLD', both of them suffers close privacy leakage in the whole image level and in the EyePACS dataset. The only exception is that SCAFFOLD achieves high defense effect in the target region level of the ChestXRay dataset. This reduction of attack strength could be due to facts that contrast between foreground and background is higher in the ChestXRay dataset, and that SCAFFOLD reduces client drifts, generating more uniform gradients considering foreground regions. Compared with FedAvg, DP, GC, and Censor reduce task performance by more than 10\% in the ChestXRay dataset, and by more than 7\% in the EyePACS dataset. The first two of them share the common point of affecting more on relative gradients in foreground regions, defensing better in target regions. However, GC is worse in MSE (-0.008) and PSNR (-0.65) in the EyePACS dataset which is dominated by background regions, because it hardly defends low-frequency information. The reason why the defensive efficacy falls below than FedAvg can be attributed to the effect of random seeds during GIA implementation, which influences the initialization of images or latent variables, consequently amplifying disparities during the reconstruction process (\cite{li2022e2egi}). For Censor (\cite{zhang2025censor}), its defensive strength is great on target regions of the ChestXRay dataset, but not workable in the EyePACS dataset. It could be due to a larger search space in the EyePACS dataset, which reduces the probability of searching orthogonal gradients with optimial training loss.

Other defensive methods, i.e., GC, Soteria, OUTPOST, and Ours, cause minimal effects on task performance. For GS, it fails to defend the EyePACS dataset in whole image regions. Since gradients of the EyePACS dataset is already small and BN statistics are unchanged, the effect of GS is limited. Both of Soteria and OUTPOST are good at defending the RGB EyePACS and worse in most defensive metrics of the grayscale ChestXRay dataset. This could be caused by several reasons: For multi-channel images, attack is more difficult due to a larger search space; adversarial noise from Soteria or combined tricks from OUTPOST can disrupt relationship between channels; the higher contrast of ChestXRay dataset is more likely to generate large gradients facilitating GIA.

Compared to all defensive methods, our method achieves optimal performance across all defensive metrics. Compared with the second best method on 4 metrics in target regions, our improvements are 0.048, 4.08, 0.228, 0.165 in the ChestXRay dataset, and 0.033, 2.19, 0.115, 0.105 in the EyePACS dataset. Taking FedAvg as reference of task performance, our framework only degrades F1 marginally of 0.011 and 0.009 on two datasets, respectively. When conducting defense strategies, other SOTA methods have not explained which gradient or image information is more susceptible to privacy breaches. Thus, even with specialized treatment of gradients at their extremities (\cite{zhu2019deep, geyer2017differentially, wang2024more}), success is only confined to some scenarios. This limitation stems from a relative insensitivity to the varying degrees of privacy risks across different image areas. In comparison, there is a significant enhancement of privacy protection within foreground regions when our framework is employed. It indicates that the vulnerability of privacy breaches cannot be adequately inferred through the magnitude of gradient values alone. Despite the integration of Grad-CAM++ to weaken noise in foreground regions, our approach still shields these zones through the relative noise derived from shadow updates.

\begin{table*}[]
  \centering
  \caption{Comparison of our method with SOTA defense methods against model-based GIA. F1 represents task performance. For defense metrics, i.e., MSE, PSNR, LPIPS, SSIM, are all the mean across 9 clients and 5 FL rounds, i.e., 1, 25, 50, 75, 100. Numbers in brackets are standard deviation of these defensive metrics across these 5 FL rounds. For each reconstructed image, we calculate metrics to the most similar real image since the gradient is derived from a global round. Target regions are decided by bounding boxes based on Grad-CAM++. Bold numbers denotes the best result among all methods. For statistical meaning, $\dag$ means $p<<0.05$ in the Wilcoxon signed rank test for comparison between FedAvg and a specific method, while $\ddag$ is for comparison between a specific method and all other SOTA methods. `DP', `GS', `GC' are abbreviation of differential privacy, gradient sparsification, and gradient clipping.}
  \resizebox{2\columnwidth}{!}{
      \begin{tabular}{clllllllllll}
          \hline
           &
          \multicolumn{1}{l}{Dataset} &
            \multicolumn{5}{l}{ChestXRay} &
            \multicolumn{5}{l}{EyePACS} \\ \cmidrule(r){3-7}\cmidrule(r){8-12}
          & \multicolumn{1}{l}{Method} &
            \multicolumn{1}{l}{F1↑} &
            \multicolumn{1}{l}{MSE↑} &
            \multicolumn{1}{l}{PSNR↓} &
            \multicolumn{1}{l}{LPIPS↑} &
            \multicolumn{1}{l}{SSIM↓} &
            \multicolumn{1}{l}{F1↑} &
            \multicolumn{1}{l}{MSE↑} &
            \multicolumn{1}{l}{PSNR↓} &
            \multicolumn{1}{l}{LPIPS↑} &
            SSIM↓ \\ \cline{2-12}
            &
          \multicolumn{1}{l}{FedAvg} &
            \multicolumn{1}{l}{0.978} &
            \multicolumn{1}{l}{0.021 (0.004)} &
            \multicolumn{1}{l}{17.05 (0.97)} &
            \multicolumn{1}{l}{0.448 (0.034)} &
            \multicolumn{1}{l}{0.531 (0.065)} &
            \multicolumn{1}{l}{0.870} &
            \multicolumn{1}{l}{0.064 (0.012)} &
            \multicolumn{1}{l}{12.26 (0.95)} &
            \multicolumn{1}{l}{0.557 (0.041)} &
            0.411 (0.042) \\&
          \multicolumn{1}{l}{SCAFFOLD} &
            \multicolumn{1}{l}{0.976} &
            \multicolumn{1}{l}{0.021 (0.005)} &
            \multicolumn{1}{l}{17.14 (1.08)} &
            \multicolumn{1}{l}{0.450 (0.030)} &
            \multicolumn{1}{l}{0.524 (0.064)} &
            \multicolumn{1}{l}{0.875} &
            \multicolumn{1}{l}{0.063 (0.015)} &
            \multicolumn{1}{l}{12.36 (1.12)} &
            \multicolumn{1}{l}{0.558 (0.048)} &
            0.414 (0.043) \\&
          \multicolumn{1}{l}{DP} &
            \multicolumn{1}{l}{0.860} &
            \multicolumn{1}{l}{0.026 (0.004)} &
            \multicolumn{1}{l}{17.20 (0.94)} &
            \multicolumn{1}{l}{\textbf{0.710 (0.052)}$\dag$} &
            \multicolumn{1}{l}{0.531 (0.075)} &
            \multicolumn{1}{l}{0.798} &
            \multicolumn{1}{l}{\textbf{0.090 (0.020)}$\dag$} &
            \multicolumn{1}{l}{10.89 (1.06)$\dag$} &
            \multicolumn{1}{l}{0.746 (0.037)$\dag$} &
            {0.359 (0.098)$\dag$} \\&
          \multicolumn{1}{l}{GS} &
            \multicolumn{1}{l}{0.968} &
            \multicolumn{1}{l}{0.025 (0.006)} &
            \multicolumn{1}{l}{16.96 (1.28)} &
            \multicolumn{1}{l}{0.445 (0.031)} &
            \multicolumn{1}{l}{0.539 (0.064)} &
            \multicolumn{1}{l}{0.861} &
            \multicolumn{1}{l}{0.067 (0.018)} &
            \multicolumn{1}{l}{12.16 (1.27)$\dag$} &
            \multicolumn{1}{l}{0.583 (0.051)} &
            0.416 (0.041) \\&
          \multicolumn{1}{l}{GC} &
            \multicolumn{1}{l}{0.848} &
            \multicolumn{1}{l}{\textbf{0.028 (0.007)}} &
            \multicolumn{1}{l}{\textbf{16.66 (1.53)}} &
            \multicolumn{1}{l}{0.449 (0.037)} &
            \multicolumn{1}{l}{0.544 (0.084)} &
            \multicolumn{1}{l}{0.693} &
            \multicolumn{1}{l}{0.057 (0.011)} &
            \multicolumn{1}{l}{12.74 (0.90)} &
            \multicolumn{1}{l}{0.582 (0.049)} &
            0.410 (0.043) \\&
          \multicolumn{1}{l}{Soteria} &
            \multicolumn{1}{l}{0.973} &
            \multicolumn{1}{l}{0.021 (0.005)} &
            \multicolumn{1}{l}{16.98 (1.06)} &
            \multicolumn{1}{l}{0.450 (0.033)} &
            \multicolumn{1}{l}{0.529 (0.066)} &
            \multicolumn{1}{l}{0.862} &
            \multicolumn{1}{l}{0.064 (0.014)} &
            \multicolumn{1}{l}{12.26 (1.04)} &
            \multicolumn{1}{l}{0.568 (0.048)$\dag$} &
            0.406 (0.043) \\&
          \multicolumn{1}{l}{OUTPOST} &
            \multicolumn{1}{l}{0.972} &
            \multicolumn{1}{l}{0.020 (0.004)} &
            \multicolumn{1}{l}{17.35 (1.04)} &
            \multicolumn{1}{l}{0.463 (0.036)$\dag$} &
            \multicolumn{1}{l}{0.520 (0.069)} &
            \multicolumn{1}{l}{0.862} &
            \multicolumn{1}{l}{0.063 (0.013)} &
            \multicolumn{1}{l}{12.33 (1.05)} &
            \multicolumn{1}{l}{0.561 (0.044)} &
            0.410 (0.042) \\&
            \multicolumn{1}{l}{Censor} &
              \multicolumn{1}{l}{0.810} &
              \multicolumn{1}{l}{0.025 (0.006)} &
              \multicolumn{1}{l}{16.85 (1.41)} &
              \multicolumn{1}{l}{0.462 (0.033)$\dag$} &
              \multicolumn{1}{l}{0.521 (0.074)} &
              \multicolumn{1}{l}{0.728} &
              \multicolumn{1}{l}{0.064 (0.009)} &
              \multicolumn{1}{l}{12.26 (0.82)} &
              \multicolumn{1}{l}{0.558 (0.057)} &
              0.411 (0.029) \\ \cline{2-12}
            \multirow{-10}{*}{\rotatebox{90}{Whole Image}} &
            \multicolumn{1}{l}{\textbf{Ours}} &
            \multicolumn{1}{l}{0.967} &
            \multicolumn{1}{l}{0.021 (0.004)} &
            \multicolumn{1}{l}{17.10 (0.81)} &
            \multicolumn{1}{l}{\textbf{0.710 (0.039)}$\dag$} &
            \multicolumn{1}{l}{\textbf{0.384 (0.061)}$\ddag$} &
            \multicolumn{1}{l}{0.861} &
            \multicolumn{1}{l}{0.088 (0.014)$\dag$} &
            \multicolumn{1}{l}{\textbf{10.86 (0.67)}$\dag$} &
            \multicolumn{1}{l}{\textbf{0.785 (0.016)}$\ddag$} &
            \textbf{0.277 (0.029)}$\ddag$ \\ \hline
             &
          \multicolumn{1}{l}{Dataset} &
            \multicolumn{5}{l}{ChestXRay} &
            \multicolumn{5}{l}{EyePACS} \\ \cmidrule(r){3-7}\cmidrule(r){8-12}
            &
          \multicolumn{1}{l}{Method} &
            \multicolumn{1}{l}{F1↑} &
            \multicolumn{1}{l}{MSE↑} &
            \multicolumn{1}{l}{PSNR↓} &
            \multicolumn{1}{l}{LPIPS↑} &
            \multicolumn{1}{l}{SSIM↓} &
            \multicolumn{1}{l}{F1↑} &
            \multicolumn{1}{l}{MSE↑} &
            \multicolumn{1}{l}{PSNR↓} &
            \multicolumn{1}{l}{LPIPS↑} &
            SSIM↓ \\ \cline{2-12}
            &
          \multicolumn{1}{l}{FedAvg} &
            \multicolumn{1}{l}{0.978} &
            \multicolumn{1}{l}{0.010 (0.002)} &
            \multicolumn{1}{l}{20.52 (1.03)} &
            \multicolumn{1}{l}{0.105 (0.015)} &
            \multicolumn{1}{l}{0.798 (0.030)} &
            \multicolumn{1}{l}{0.870} &
            \multicolumn{1}{l}{0.042 (0.010)} &
            \multicolumn{1}{l}{14.26 (1.15)} &
            \multicolumn{1}{l}{0.372 (0.021)} &
            0.569 (0.026) \\&
          \multicolumn{1}{l}{SCAFFOLD} &
            \multicolumn{1}{l}{0.976} &
            \multicolumn{1}{l}{0.014 (0.003)} &
            \multicolumn{1}{l}{18.98 (1.06)} &
            \multicolumn{1}{l}{0.181 (0.016)} &
            \multicolumn{1}{l}{0.692 (0.042)} &
            \multicolumn{1}{l}{0.870} &
            \multicolumn{1}{l}{0.040 (0.011)} &
            \multicolumn{1}{l}{14.48 (1.29)} &
            \multicolumn{1}{l}{0.362 (0.024)} &
            0.571 (0.028) \\&
          \multicolumn{1}{l}{DP} &
            \multicolumn{1}{l}{0.860} &
            \multicolumn{1}{l}{0.012 (0.002)} &
            \multicolumn{1}{l}{20.66 (1.06)} &
            \multicolumn{1}{l}{0.143 (0.016)$\dag$} &
            \multicolumn{1}{l}{0.799 (0.036)} &
            \multicolumn{1}{l}{0.798} &
            \multicolumn{1}{l}{0.053 (0.011)$\dag$} &
            \multicolumn{1}{l}{13.28 (1.03)$\dag$} &
            \multicolumn{1}{l}{0.411 (0.022)} &
            0.575 (0.078)\\&
          \multicolumn{1}{l}{GS} &
            \multicolumn{1}{l}{0.968} &
            \multicolumn{1}{l}{0.012 (0.002)$\dag$} &
            \multicolumn{1}{l}{20.09 (1.12)} &
            \multicolumn{1}{l}{0.123 (0.019)$\dag$} &
            \multicolumn{1}{l}{0.792 (0.030)} &
            \multicolumn{1}{l}{0.861} &
            \multicolumn{1}{l}{0.033 (0.011)} &
            \multicolumn{1}{l}{15.38 (1.48)} &
            \multicolumn{1}{l}{0.327 (0.021)} &
            0.672 (0.022) \\&
          \multicolumn{1}{l}{GC} &
            \multicolumn{1}{l}{0.848} &
            \multicolumn{1}{l}{0.011 (0.003)} &
            \multicolumn{1}{l}{20.47 (1.55)} &
            \multicolumn{1}{l}{0.144 (0.020)$\dag$} &
            \multicolumn{1}{l}{0.800 (0.042)} &
            \multicolumn{1}{l}{0.693} &
            \multicolumn{1}{l}{0.057 (0.009)} &
            \multicolumn{1}{l}{12.74 (1.08)} &
            \multicolumn{1}{l}{0.582 (0.024)$\dag$} &
            0.410 (0.032) \\&
          \multicolumn{1}{l}{Soteria} &
            \multicolumn{1}{l}{0.973} &
            \multicolumn{1}{l}{0.009 (0.002)} &
            \multicolumn{1}{l}{20.70 (1.13)} &
            \multicolumn{1}{l}{0.105 (0.014)} &
            \multicolumn{1}{l}{0.794 (0.030)} &
            \multicolumn{1}{l}{0.862} &
            \multicolumn{1}{l}{0.042 (0.010)} &
            \multicolumn{1}{l}{14.23 (1.17)} &
            \multicolumn{1}{l}{0.378 (0.023)$\dag$} &
            0.569 (0.028) \\&
          \multicolumn{1}{l}{OUTPOST} &
            \multicolumn{1}{l}{0.972} &
            \multicolumn{1}{l}{0.009 (0.002)} &
            \multicolumn{1}{l}{21.17 (1.17)} &
            \multicolumn{1}{l}{0.123 (0.017)$\dag$} &
            \multicolumn{1}{l}{0.796 (0.035)} &
            \multicolumn{1}{l}{0.862} &
            \multicolumn{1}{l}{0.043 (0.010)} &
            \multicolumn{1}{l}{14.13 (1.13)} &
            \multicolumn{1}{l}{0.384 (0.018)$\dag$} &
            0.572 (0.030) \\&
            \multicolumn{1}{l}{Censor} &
              \multicolumn{1}{l}{0.810} &
              \multicolumn{1}{l}{\textbf{0.016 (0.002)}$\dag$} &
              \multicolumn{1}{l}{18.59 (1.17)$\dag$} &
              \multicolumn{1}{l}{0.155 (0.017)$\dag$} &
              \multicolumn{1}{l}{0.721 (0.035)$\dag$} &
              \multicolumn{1}{l}{0.728} &
              \multicolumn{1}{l}{0.044 (0.010)} &
              \multicolumn{1}{l}{14.06 (1.13)} &
              \multicolumn{1}{l}{0.373 (0.018)} &
              0.603 (0.030) \\ \cline{2-12}
            \multirow{-10}{*}{\rotatebox{90}{Target Region}} &
          \multicolumn{1}{l}{\textbf{Ours}} &
            \multicolumn{1}{l}{0.967} &
            \multicolumn{1}{l}{\textbf{0.016 (0.003)}$\ddag$} &
            \multicolumn{1}{l}{\textbf{18.42 (0.79)}$\ddag$} &
            \multicolumn{1}{l}{\textbf{0.335 (0.034)}$\ddag$} &
            \multicolumn{1}{l}{\textbf{0.589 (0.038)}$\ddag$} &
            \multicolumn{1}{l}{0.861} &
            \multicolumn{1}{l}{\textbf{0.063 (0.009)}$\ddag$} &
            \multicolumn{1}{l}{\textbf{12.28 (0.69)}$\ddag$} &
            \multicolumn{1}{l}{\textbf{0.476 (0.014)}$\ddag$} &
            \textbf{0.496 (0.021)}$\ddag$ \\ \hline
          \end{tabular}
  }
  \label{table 2}
\end{table*}

Table \ref{table 2} provides defensive results against optimization-based GIA. Within whole-image metrics, our framework surpasses SOTA methods across most measures. It is noteworthy that our proposed shadow defense strategy, while similar in approach to the model-based GIA, also exhibits generalizable protection against other forms of GIA. This efficacy validates the motivation of our method: to attenuate the mapping relationship between gradients or auxiliary information and sensitive images. Furthermore, in the context of optimization-based GIAs, foreground-area protection of our method remains unmatched by other SOTA methods. Beyond performance and defensive metrics, we have also listed computational costs associated with each approach in the appendix.

\begin{figure}[htbp!]
    \begin{center}
    \includegraphics[width=\columnwidth]{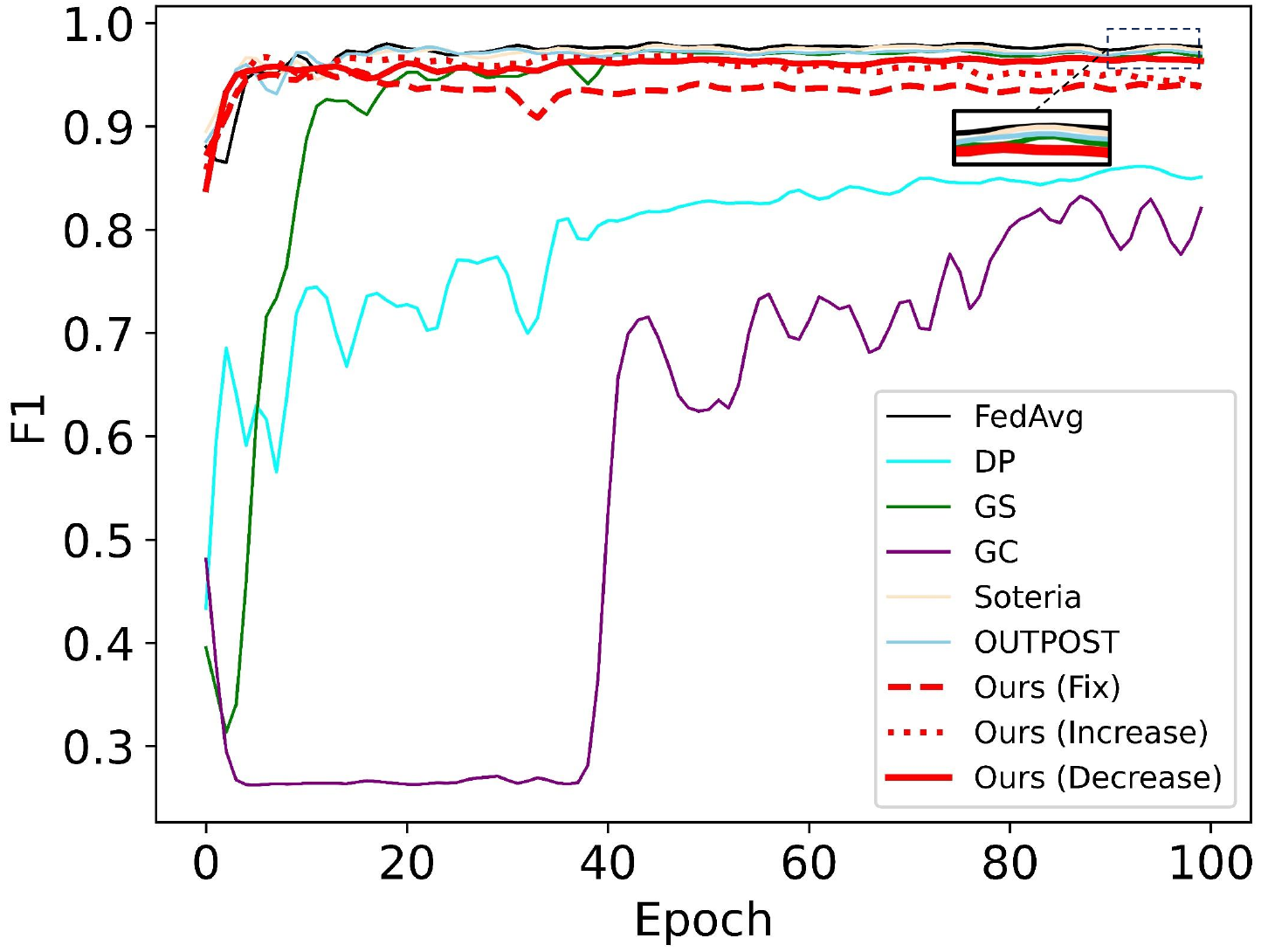}
    \end{center}
    \caption{Comparison of F1 curve during FL training for difference methods. F1 values of the global model are calculated based on an independent test set. `Fix', `Increase', `Decrease' corresponds to fixing, decreasing and increasing the amplitude of final noise $N$ during FL training, respectively.}
    \label{fig6}
\end{figure}

To investigate the impact of various defensive measures on the convergence of the task, we illustrate the F1 score curve of local training in \textcolor{SkyBlue}{Fig.} \ref{fig6}. Before the 20th global round, significant oscillations are observed in performance of DP, gradient sparsification, and gradient clipping, whereas other methods appear to approach a state of convergence around this juncture. Among these three, gradient clipping fails to stabilize even at the 100th round. DP converges around the 80th round, while gradient sparsification reaches a similar state only by the 40th round, which is twice as slow as that exhibited by other SOTA methods. Our method can be equipped with different adjustment strategies of noise amplitude, thus achieving desired balance between task performance and privacy protection, as shown in \textcolor{SkyBlue}{Fig.} \ref{fig6} and \textcolor{SkyBlue}{Fig.} \ref{fig8}. The gap between fixing and increasing the noise is less than 1\% until the 75th epoch, showing the minimum side effect of method to the main task, due to the noise substraction operation for foreground regions. After the 75th epoch, the task performance with increased noise suffers a 3\% degradation, which calls for a better solution in the long-time training scenarios.

\begin{figure*}[htbp!]
    \begin{center}
    \includegraphics[width=2\columnwidth]{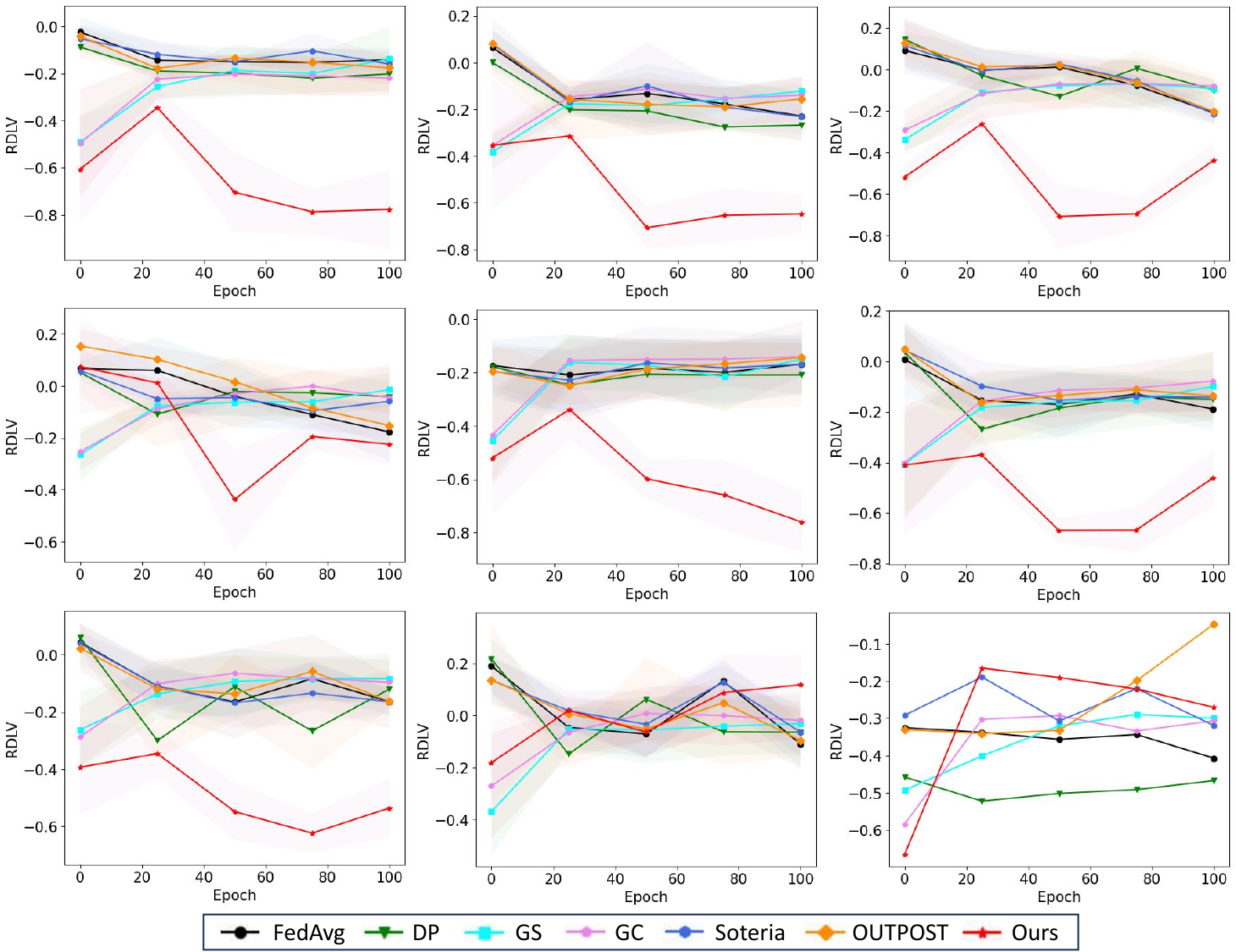}
    \end{center}
    \caption{Comparison of RDLV curve for difference methods on all clients under model-based GIA. RDLV represents privacy reveal degree compared with a template image (\cite{hatamizadeh2023gradient}). If RDLV is less than 0, privacy leakage is negligible.}
    \label{fig7}
\end{figure*}

To reflect the effectiveness of our method in protecting privacy throughout the entire FL process more accurately, we employ the Relative Data Leakage Value (RDLV), a metric proposed in Hatamizadeh et al. (\cite{hatamizadeh2023gradient}), to quantify the extent of privacy leakage. It is defined as:

\begin{flalign}
  &\operatorname{RDLV}=\frac{\operatorname{SSIM}\left(x, x_s\right)-\operatorname{SSIM}\left(x_s, P\right)}{\operatorname{SSIM}\left(x_s, P\right)} ,&
\end{flalign}
where $x, x_s, P$ are training image, reconstructed image, and prior image (the mean of an image dataset), respectively. An RDLV value below 0 indicates that the degree of privacy leakage is less than that associated with a template image, which suggests that the risk of privacy breach is negligible. As illustrated in \textcolor{SkyBlue}{Fig.} \ref{fig7}, our method, along with gradient sparsification and gradient clipping, demonstrates superior protective capabilities in early training. At these stages, our approach achieves the strongest defense on all datasets except for clients 2, 4, and 8. Overall, the trend of all curves indicate a robust privacy protection effect inherent in our method. The relatively weaker protection observed on client 9 can be attributed to the fact that this client possesses only one image. Consequently, the shadow training is more prone to overfitting on that single training image, leading to a less accurate estimate of the relative noise. However, it should be noted that RDLV values for all defensive strategies on client 9 are almost all below -0.1, signifying a level of defense that is sufficient to protect this vulnerable image. Furthermore, we show RDLV curves for optimization-based GIA throughout the training process in the appendix.

\subsubsection{Ablation Study}

\begin{table}[]
  \centering
  \caption{Module ablation of the shadow defense framework. `PT', `FT', `z', `S', `Equ', `CAM' are abbreviations of pretraining, fine-tuning, latent code, shadow model, histogram equalization, and Grad-CAM++. }
  \resizebox{\columnwidth}{!}{
    \begin{tabular}{llllll}
      \hline
      Method   & F1↑   & MSE↑  & PSNR↓ & LPIPS↑ & SSIM↓ \\ \hline
      FedAvg   & 0.978 & 0.035 & 15.09 & 0.385  & 0.486 \\ 
      w/o PT z & 0.965 & 0.069 & 12.88 & 0.583  & 0.317 \\ 
      w/o FT S & 0.959 & 0.073 & 12.87 & 0.573  & 0.378 \\ 
      w/o Equ  & 0.967 & 0.065 & 13.21 & 0.510  & 0.329 \\ 
      w/o CAM  & 0.946 & 0.098 & 11.14 & 0.658  & 0.240 \\ 
      Ours     & 0.967 & 0.102 & 11.36 & 0.642  & 0.286 \\ \hline
      \end{tabular}
  }
  \label{table 3}
\end{table}
\textbf{Module ablation.} In Table \ref{table 3}, we show an ablation study on main components of our framework. Among these components, pre-training of latent variables and fine-tuning of the shadow model occupy primary computational costs. It can be found that the removal of either has a negligible effect on model performance. However, the decreasement of most defensive metrics is significant when either is omitted, particularly the degradation in PSNR, which exceeds 1.5. Moreover, the impact on the SSIM varies as much as 0.061 between the two, indicating that the fine-tuned shadow model effectively reduces the leakage of statistical information in image features, thereby weakening the key BN regularization in the GIA loss. Therefore, within our framework, both operations are essential from a privacy preservation standpoint.

After fine-tuning the shadow model, the post-processing step of computing noise is also crucial. Without histogram equalization on relative noise, although model performance remains unchanged, there is a marked degradation across four defensive metrics by 0.011, 0.77, 0.08, and 0.038, respectively. This decline is attributed to the focus of the original noise map, which is on regions most susceptible to reconstruction, leaving other potential areas of privacy information unprotected and providing adversaries with a shortcut for attack. In contrast, neglecting noise reduction in foreground regions, identified by Grad-CAM++, results in a 0.021 decrease in task performance, but it enhances defensive capabilities. To strike a more favourable balance between task performance and privacy protection, we have retained this module. It is tailored to accommodate varying intensities of GIA in real-world scenarios (\cite{wang2024more}).

\begin{table}[]
  \centering
  \caption{Effect of different kind of image noises on GIA. `Img', `w', `s' are abbreviations of image-level, weak noise, strong noise.}
  \resizebox{\columnwidth}{!}{
    \begin{tabular}{llllll}
      \hline
      Method     & F1↑   & MSE↑  & PSNR↓ & LPIPS↑ & SSIM↓ \\ \hline
      FedAvg     & 0.978 & 0.035 & 15.09 & 0.385  & 0.486 \\ 
      Img DP (w) & 0.953 & 0.035 & 14.92 & 0.529  & 0.377 \\ 
      Img DP (s) & 0.867 & 0.060 & 13.32 & 0.778  & 0.204 \\ 
      Ours       & 0.967 & 0.102 & 11.36 & 0.642  & 0.286 \\ \hline
      \end{tabular}
  }
  \label{table 4}
\end{table}
\textbf{Image-level noise.} Our defensive strategy can be regarded as a variant of image-level DP. To this end, we present the results of directly applying DP to images instead of gradients, as shown in Table \ref{table 4}. It is observed that if the privacy budget is abundant (weaker noise) when we incorporate noise directly into images, the performance across all four defensive metrics is significantly inferior to our approach. Moreover, the task performance, as indicated by the F1 score, suffers an additional reduction of 0.014 compared to our method. It demonstrates that the sensitive area simulation strategy of the shadow model facilitates a better balance between privacy protection and task performance. Furthermore, even when intense noise is added, resulting in a tenfold performance reduction (row 3), our method still achieves stronger privacy protection capabilities on PSNR. This is because information content of different patches in an image often varies greatly. Applying noise indiscriminately across all regions does not precisely protect image information that is most vulnerable to GIA.

\begin{figure}[htbp!]
  \begin{center}
  \includegraphics[width=\columnwidth]{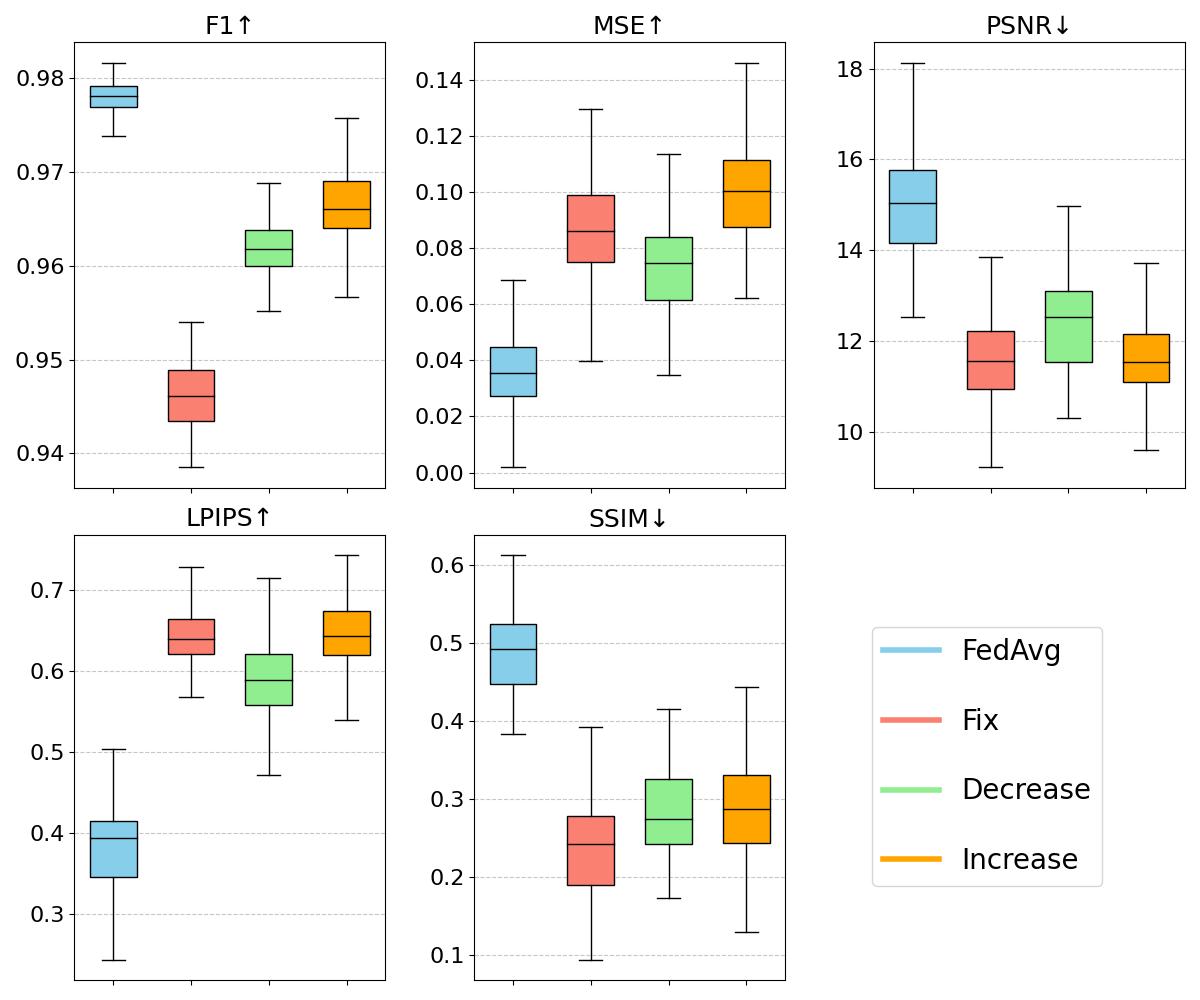}
  \end{center}
  \caption{Effect of noise adjustment strategy on GIA. `FedAvg' represents the method without defense. `Fix', `Decrease', `Increase' corresponds to fixing, decreasing and increasing the absolute noise amplitude our method during FL training, respectively.}
  \label{fig8}
\end{figure}

\textbf{Dynamic adjustment of noise.} To determine the most effective noise adjustment strategy, we present effects of different strategies on model performance and defensive capability in \textcolor{SkyBlue}{Fig.} \ref{fig8}. Compared to the unfortified FedAvg, progressively increasing the noise level as training advances results in superior F1 scores relative to either diminishing or constant noise levels. It is due to the minimal impact this strategy has on early stages of convergence. There is a fact that BN statistics are increasingly accurate during training, consequently enhancing the potency of GIA (\cite{hatamizadeh2023gradient}). Thus, escalating the noise amplitude aligns with this trend. Furthermore, this graph demonstrates defensive advantages of this method over a reduction in noise levels. Nonetheless, we observe that progressively increasing and maintaining the noise amplitude separately achieve optimal results on MSE and SSIM. The former indicates the absolute precision in pixel reconstruction, while the latter represents overall structural information. Considering significant disparities in task performance, we choose the increasing scheme of the noise level, which strikes a better balance between privacy preservation and task performance.

\begin{table}[]
  \centering
  \caption{Comparison of different fine-tuning rounds of the shadow model.}
  \setlength{\tabcolsep}{3pt}
  \resizebox{\columnwidth}{!}{
    \begin{tabular}{lllllll}
      \hline
      Round & F1↑   & MSE↑  & PSNR↓ & LPIPS↑ & SSIM↓ & Time \\ \hline
      10    & 0.969 & 0.097 & 11.92 & 0.423  & 0.277 & 1.000$\times$  \\ 
      20    & 0.967 & 0.096 & 12.02 & 0.433  & 0.271 & 1.394$\times$  \\ 
      40    & 0.966 & 0.105 & 11.30 & 0.491  & 0.224 & 1.963$\times$  \\ 
      60    & 0.973 & 0.146 & 9.78  & 0.628  & 0.170 & 2.651$\times$  \\ 
      80    & 0.967 & 0.115 & 10.62 & 0.578  & 0.191 & 3.460$\times$  \\ 
      100   & 0.974 & 0.151 & 10.04 & 0.518  & 0.211 & 4.272$\times$  \\ \hline
      \end{tabular}
  }
  \label{table 5}
\end{table}

\textbf{Fine-tuning rounds of the shadow model.} In the fine-tuning process of the shadow model, to balance computational costs, we showcase in Table \ref{table 5} the effects of different termination epochs on task performance and privacy protection for the most vulnerable client, i.e., client 8. First, the impact of various termination epochs on task performance is marginal. Subsequently, considering four defensive indices, selecting 60 as the termination epoch for fine-tuning the shadow model is shown to be optimal except for MSE. However, since the computational expense of this setting is 2.651 times greater than that required at 10 epochs, we set 20 as the final termination epoch after weighing up performance, defense, and computational costs.

\subsubsection{Visualization analysis}
\begin{figure*}[htbp!]
  \begin{center}
  \includegraphics[width=1.9\columnwidth]{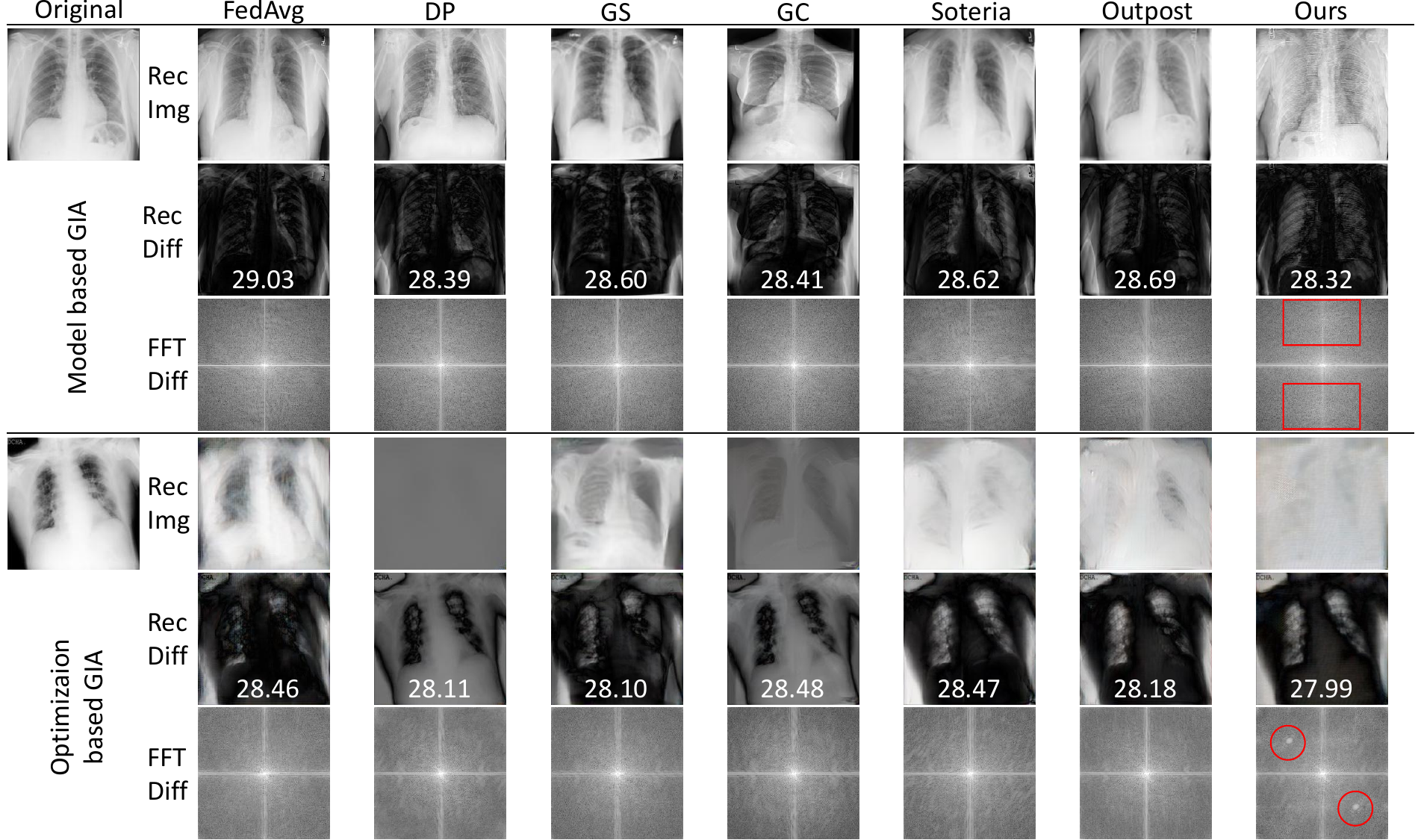}
  \end{center}
  \caption{Reconstructed images of the ChestXRay dataset from optimization-based GIA. `Rec Img', `Rec Diff', `FFT diff' represent reconstructed images, reconstructive error maps, frequency spectrum of reconstructive error maps, respectively. The numbers indicate PSNR between the original reference image and the reconstructed image.}
  \label{fig9}
\end{figure*}
\begin{figure*}[htbp!]
  \begin{center}
  \includegraphics[width=1.9\columnwidth]{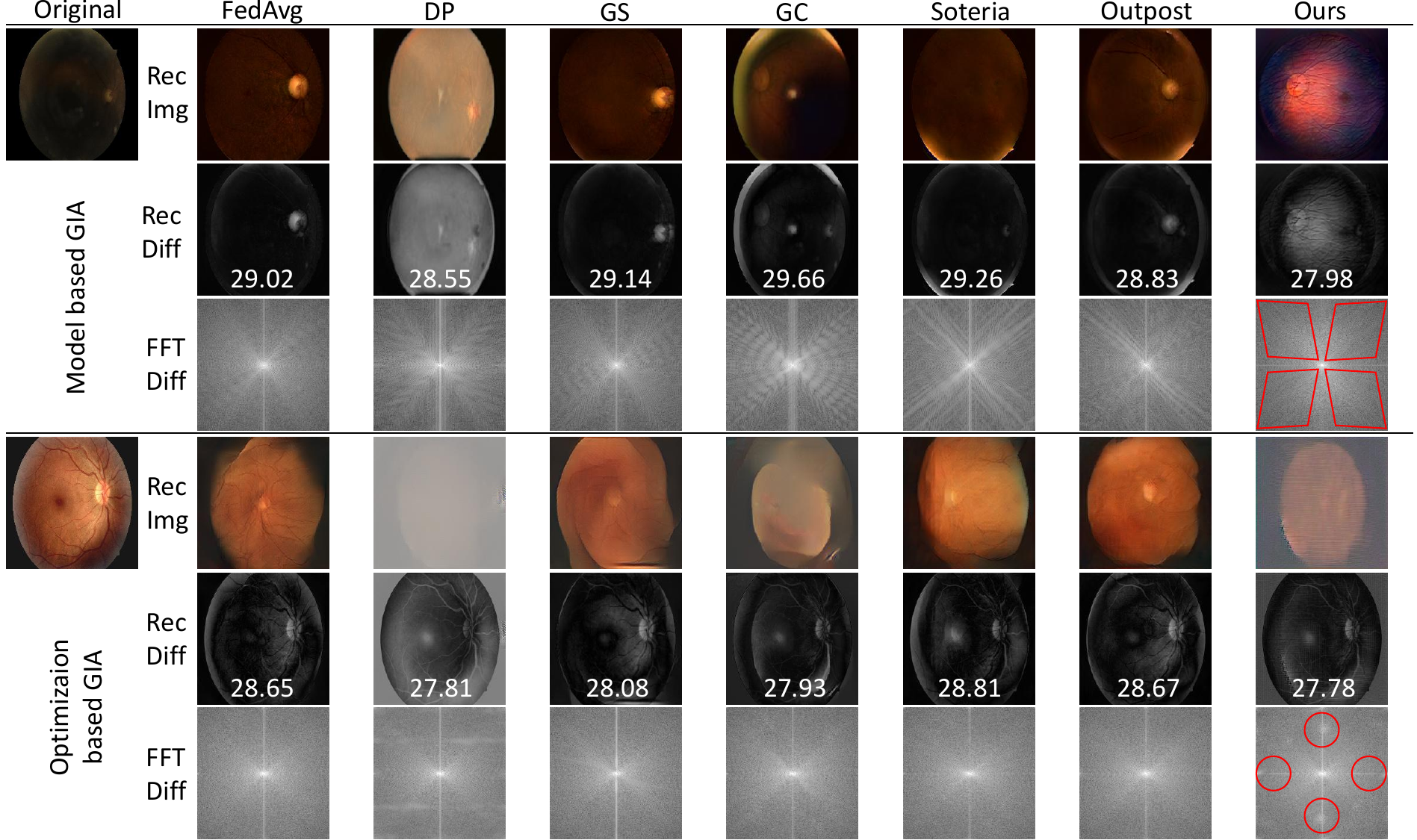}
  \end{center}
  \caption{Reconstructed images of the EyePACS dataset from optimization-based GIA `Rec Img', `Rec Diff', `FFT diff' represent reconstructed images, reconstructive error maps, frequency spectrum of reconstructive error maps, respectively. The numbers indicate PSNR between the original reference image and the reconstructed image.}
  \label{fig10}
\end{figure*}

\textbf{Reconstructed images.} We illustrate reconstruction images of the optimized-based GIA on two datasets in \textcolor{SkyBlue}{Fig.} \ref{fig9} and \textcolor{SkyBlue}{Fig.} \ref{fig10} at the middle of training. Images for FFT phase of 'FFT Diff' and conclusions are shown in the Supplementary Materials. Vulnerable training images, as well as visual results of the model-based GIA and attacks during the FL process, can be found in the appendix. For both datasets, DP and our method significantly change the overall appearance of reconstructed images under arbitrary GIA types, proving that sensitive patient privacy in foreground regions is greatly protected. Although GC has also changed the appearance, some key parts are not free from attack, e.g., the one under optimization based GIA. Similarly, GS suffers the same problem, but with weaker protection ability. Both methods also only extremum values in gradients, ignoring significance of sensitive privacy with moderate gradients. In comparison, Soteria, aiming at maximizing reconstruction error through a single layer, is fragile for all medical scenarios. For the ChestXRay dataset, Outpost is capable of providing safe safety guarantee. However, when the feature space becomes more complex, its delicate combination protection may lose its effcacy due to its unexplainable gradient perturbation nature. 

From the frequency spectrum of reconstructive error map in the thrid row of each setting (FFT Diff), we observe two specific patterns of our method. Firstly, for both datasets, when it comes to model-based GIA, shiny lines are greatly blurred, which represent high-frequency components (noise) have replaced low-frequency ones along horizontal cross lines (the single-channel ChestXRay dataset) or oblique cross ones (the multi-channel EyePACS dataset). Secondly, for optimization-based GIA, orthogonal lines of dominant ones are brighter, with two lightspot quite obvious. They usually indicate medium scale textures or repetitive patterns, such as or rib structure in the ChestXRay dataset or vascular reticular structures in the EyePACS dataset. In summary, these unique and unified patterns are probably related to those features tend to be reconstructed through specific type of attack. As a counteract, our method is generalizable enough to defend various kinds of GIA, thus serving as a reliable strategy for privacy protection.

\section{Discussion}\label{Discussion}
In this study, we propose a novel defense framework against gradient inversion attack to meet some requirements in privacy protection laws. For example, data minimization, anonymization, de-identification, lawfulness, transparency are required in GDPR (\cite{gdpr2016general}). This framework is under the scenario of FL to meet data minimization, namely collecting only local update information from clients. Since our framework prevents potential attackers to reconstruct recognizable training data, anonymization and de-identification are also met. Equipped with explainable tools for both task and privacy protection, our method is able to comply with lawfulness and transparency.

We validate several attributes of our proposed framework through experiments. Here are some crucial conclusions:

\begin{itemize}
  \item Compared with SOTA defense methods against GIA, our method achieves state-of-the-art privacy protection with minimal task performance degradation.
  \item Our method achieves significant improvements in foreground-region protection.
  \item There is strong defense throughout training lifecycle of our method.
  \item Our method is generalizability across medical image datasets, GIA types, image processing tasks.
  \item Our method is superior to non-targeted image-level differential privacy.
  \item Each component in our design is necessary.
  \item Fine-tuning of the shadow model during FL training is computationally efficient.
\end{itemize}

In addition to X-ray classification, our model is also generalizale to other medical image modalities and tasks, like magnetic resonance imaging (MRI) segmentation. Results are shown in Section 3 in Supplementary Materials. To our best knowledge, there is no existing label restoration algorithm in GIA for image segmentation, so we use ground truth masks and fix them during GIA. We find that our proposed method can also achieves a good balance between task performance and privacy protection in this setting, which validates that our method is applicable to various medical imaging scenarios. 

We also test compatibility of GIA with modern architectures, such as Vision Transformer (ViT) in Section 4 in Supplementary Materials. Due to a lack of batch normalization layer, both types of GIA fail to work at such a large model and high image resolution. Therefore, defense is not neccessary in this case and it remains a problem in GIA to work on larger models.

We have validate the effectiveness of our framework for both model-based and optimization-based GIA. While the former is based on GAN and similar to our defensive process, the later alone may not show the full potential of our method against various types of GIA. For this reason, we explore two more GIA types, i.e., CI-Net (\cite{zhang2023generative}) and MKOR (\cite{wang2024maximum}), in Section 4 and 5 of Supplementary Materials, respectively. Our proposed framework is also generalizable to CI-Net regarding privacy defense. For the analytics-based GIA, i.e., MKOR, even with the assumption of a malicious server, it can hardly achieve better reconstruction results compared with other types of GIA on the medical image dataset. 

To further test our proposed method in other privacy-sensitive scenarios, we utilize our method on the VGGFace2 dataset (\cite{cao2018vggface2}). Although our method still works on non-medical images regarding privacy defense, the task performance drop is huge, which will be a further direction of our research work.

\section{Conclusion}\label{Conclusion}
In this work, we introduce a shadow model-based defense framework to counter gradient inversion attacks within the context of Federated Learning. We develop pre-training and fine-tuning methods for shadow models to rapidly adapt to real attacks from potential adversaries in FL. Leveraging images reconstructed from shadow models, we propose an image noise generation technique to disrupt the mapping relationship between gradients or auxiliary information and training images. Our method has been experimented on two public datasets, ChestXRay and EyePACS, which outperforms SOTA defense strategies in both complete images and foreground regions. Besides, it can reduce the effectiveness of various types of gradient inversion attacks. In the future, it warrants further in-depth exploration into the efficiency of our method to accommodate large datasets and foundational models, as well as its applicability in scenarios, like few-shot learning.

\section*{Acknowledgment}
This work was supported by the National Key R\&D Program of China (No. 2021YFA1003004).

The work of Le Jiang was supported by a CSC scholarship.

Guang Yang was supported in part by the ERC IMI (101005122), the H2020 (952172), the MRC (MC/PC/21013), the Royal Society (IEC\verb|\|NSFC\verb|\|211235), the NVIDIA Academic Hardware Grant Program, the SABER project supported by Boehringer Ingelheim Ltd, NIHR Imperial Biomedical Research Centre (RDA01), The Wellcome Leap Dynamic resilience program (co-funded by Temasek Trust)., UKRI guarantee funding for Horizon Europe MSCA Postdoctoral Fellowships (EP/Z002206/1), UKRI MRC Research Grant, TFS Research Grants (MR/U506710/1), and the UKRI Future Leaders Fellowship (MR/V023799/1).

\bibliographystyle{model2-names.bst}\biboptions{authoryear}
\bibliography{example}

\begin{thebibliography}{43}
\expandafter\ifx\csname natexlab\endcsname\relax\def\natexlab#1{#1}\fi
\providecommand{\url}[1]{\texttt{#1}}
\providecommand{\href}[2]{#2}
\providecommand{\path}[1]{#1}
\providecommand{\DOIprefix}{doi:}
\providecommand{\ArXivprefix}{arXiv:}
\providecommand{\URLprefix}{URL: }
\providecommand{\Pubmedprefix}{pmid:}
\providecommand{\doi}[1]{\href{http://dx.doi.org/#1}{\path{#1}}}
\providecommand{\Pubmed}[1]{\href{pmid:#1}{\path{#1}}}
\providecommand{\bibinfo}[2]{#2}
\ifx\xfnm\relax \def\xfnm[#1]{\unskip,\space#1}\fi
\bibitem[{Abadi et~al.(2016)Abadi, Chu, Goodfellow, McMahan, Mironov, Talwar and Zhang}]{abadi2016deep}
\bibinfo{author}{Abadi, M.}, \bibinfo{author}{Chu, A.}, \bibinfo{author}{Goodfellow, I.}, \bibinfo{author}{McMahan, H.B.}, \bibinfo{author}{Mironov, I.}, \bibinfo{author}{Talwar, K.}, \bibinfo{author}{Zhang, L.}, \bibinfo{year}{2016}.
\newblock \bibinfo{title}{Deep learning with differential privacy}, in: \bibinfo{booktitle}{Proceedings of the 2016 ACM SIGSAC conference on computer and communications security}, pp. \bibinfo{pages}{308--318}.
\bibitem[{Bonawitz et~al.(2017)Bonawitz, Ivanov, Kreuter, Marcedone, McMahan, Patel, Ramage, Segal and Seth}]{bonawitz2017practical}
\bibinfo{author}{Bonawitz, K.}, \bibinfo{author}{Ivanov, V.}, \bibinfo{author}{Kreuter, B.}, \bibinfo{author}{Marcedone, A.}, \bibinfo{author}{McMahan, H.B.}, \bibinfo{author}{Patel, S.}, \bibinfo{author}{Ramage, D.}, \bibinfo{author}{Segal, A.}, \bibinfo{author}{Seth, K.}, \bibinfo{year}{2017}.
\newblock \bibinfo{title}{Practical secure aggregation for privacy-preserving machine learning}, in: \bibinfo{booktitle}{proceedings of the 2017 ACM SIGSAC Conference on Computer and Communications Security}, pp. \bibinfo{pages}{1175--1191}.
\bibitem[{Bu et~al.(2023)Bu, Huang and Cui}]{bu2023towards}
\bibinfo{author}{Bu, Q.}, \bibinfo{author}{Huang, D.}, \bibinfo{author}{Cui, H.}, \bibinfo{year}{2023}.
\newblock \bibinfo{title}{Towards building more robust models with frequency bias}, in: \bibinfo{booktitle}{Proceedings of the IEEE/CVF International Conference on Computer Vision}, pp. \bibinfo{pages}{4402--4411}.
\bibitem[{Cao et~al.(2018)Cao, Shen, Xie, Parkhi and Zisserman}]{cao2018vggface2}
\bibinfo{author}{Cao, Q.}, \bibinfo{author}{Shen, L.}, \bibinfo{author}{Xie, W.}, \bibinfo{author}{Parkhi, O.M.}, \bibinfo{author}{Zisserman, A.}, \bibinfo{year}{2018}.
\newblock \bibinfo{title}{Vggface2: A dataset for recognising faces across pose and age}, in: \bibinfo{booktitle}{2018 13th IEEE international conference on automatic face \& gesture recognition (FG 2018)}, \bibinfo{organization}{IEEE}. pp. \bibinfo{pages}{67--74}.
\bibitem[{Cardoso et~al.(2022)Cardoso, Li, Brown, Ma, Kerfoot, Wang, Murrey, Myronenko, Zhao, Yang et~al.}]{cardoso2022monai}
\bibinfo{author}{Cardoso, M.J.}, \bibinfo{author}{Li, W.}, \bibinfo{author}{Brown, R.}, \bibinfo{author}{Ma, N.}, \bibinfo{author}{Kerfoot, E.}, \bibinfo{author}{Wang, Y.}, \bibinfo{author}{Murrey, B.}, \bibinfo{author}{Myronenko, A.}, \bibinfo{author}{Zhao, C.}, \bibinfo{author}{Yang, D.}, et~al., \bibinfo{year}{2022}.
\newblock \bibinfo{title}{Monai: An open-source framework for deep learning in healthcare}.
\newblock \bibinfo{journal}{arXiv preprint arXiv:2211.02701} .
\bibitem[{Chang and Zhu(2024)}]{chang2024gradient}
\bibinfo{author}{Chang, W.}, \bibinfo{author}{Zhu, T.}, \bibinfo{year}{2024}.
\newblock \bibinfo{title}{Gradient-based defense methods for data leakage in vertical federated learning}.
\newblock \bibinfo{journal}{Computers \& Security} \bibinfo{volume}{139}, \bibinfo{pages}{103744}.
\bibitem[{Chattopadhay et~al.(2018)Chattopadhay, Sarkar, Howlader and Balasubramanian}]{chattopadhay2018grad}
\bibinfo{author}{Chattopadhay, A.}, \bibinfo{author}{Sarkar, A.}, \bibinfo{author}{Howlader, P.}, \bibinfo{author}{Balasubramanian, V.N.}, \bibinfo{year}{2018}.
\newblock \bibinfo{title}{Grad-cam++: Generalized gradient-based visual explanations for deep convolutional networks}, in: \bibinfo{booktitle}{2018 IEEE winter conference on applications of computer vision (WACV)}, \bibinfo{organization}{IEEE}. pp. \bibinfo{pages}{839--847}.
\bibitem[{Chowdhury et~al.(2020)Chowdhury, Rahman, Khandakar, Mazhar, Kadir, Mahbub, Islam, Khan, Iqbal, Al~Emadi et~al.}]{chowdhury2020can}
\bibinfo{author}{Chowdhury, M.E.}, \bibinfo{author}{Rahman, T.}, \bibinfo{author}{Khandakar, A.}, \bibinfo{author}{Mazhar, R.}, \bibinfo{author}{Kadir, M.A.}, \bibinfo{author}{Mahbub, Z.B.}, \bibinfo{author}{Islam, K.R.}, \bibinfo{author}{Khan, M.S.}, \bibinfo{author}{Iqbal, A.}, \bibinfo{author}{Al~Emadi, N.}, et~al., \bibinfo{year}{2020}.
\newblock \bibinfo{title}{Can ai help in screening viral and covid-19 pneumonia?}
\newblock \bibinfo{journal}{Ieee Access} \bibinfo{volume}{8}, \bibinfo{pages}{132665--132676}.
\bibitem[{Fang et~al.(2023)Fang, Chen, Wang, Wang and Xia}]{fang2023gifd}
\bibinfo{author}{Fang, H.}, \bibinfo{author}{Chen, B.}, \bibinfo{author}{Wang, X.}, \bibinfo{author}{Wang, Z.}, \bibinfo{author}{Xia, S.T.}, \bibinfo{year}{2023}.
\newblock \bibinfo{title}{Gifd: A generative gradient inversion method with feature domain optimization}, in: \bibinfo{booktitle}{Proceedings of the IEEE/CVF International Conference on Computer Vision}, pp. \bibinfo{pages}{4967--4976}.
\bibitem[{Garg and Jain(2017)}]{garg2017comparative}
\bibinfo{author}{Garg, P.}, \bibinfo{author}{Jain, T.}, \bibinfo{year}{2017}.
\newblock \bibinfo{title}{A comparative study on histogram equalization and cumulative histogram equalization}.
\newblock \bibinfo{journal}{International Journal of New Technology and Research} \bibinfo{volume}{3}, \bibinfo{pages}{263242}.
\bibitem[{GDPR(2016)}]{gdpr2016general}
\bibinfo{author}{GDPR, G.D.P.R.}, \bibinfo{year}{2016}.
\newblock \bibinfo{title}{General data protection regulation}.
\newblock \bibinfo{journal}{Regulation (EU) 2016/679 of the European Parliament and of the Council of 27 April 2016 on the protection of natural persons with regard to the processing of personal data and on the free movement of such data, and repealing Directive 95/46/EC} .
\bibitem[{Geng et~al.(2023)Geng, Mou, Li, Li, Beyan, Decker and Rong}]{geng2023improved}
\bibinfo{author}{Geng, J.}, \bibinfo{author}{Mou, Y.}, \bibinfo{author}{Li, Q.}, \bibinfo{author}{Li, F.}, \bibinfo{author}{Beyan, O.}, \bibinfo{author}{Decker, S.}, \bibinfo{author}{Rong, C.}, \bibinfo{year}{2023}.
\newblock \bibinfo{title}{Improved gradient inversion attacks and defenses in federated learning}.
\newblock \bibinfo{journal}{IEEE Transactions on Big Data} .
\bibitem[{Geyer et~al.(2017)Geyer, Klein and Nabi}]{geyer2017differentially}
\bibinfo{author}{Geyer, R.C.}, \bibinfo{author}{Klein, T.}, \bibinfo{author}{Nabi, M.}, \bibinfo{year}{2017}.
\newblock \bibinfo{title}{Differentially private federated learning: A client level perspective}.
\newblock \bibinfo{journal}{arXiv preprint arXiv:1712.07557} .
\bibitem[{Hatamizadeh et~al.(2023)Hatamizadeh, Yin, Molchanov, Myronenko, Li, Dogra, Feng, Flores, Kautz, Xu et~al.}]{hatamizadeh2023gradient}
\bibinfo{author}{Hatamizadeh, A.}, \bibinfo{author}{Yin, H.}, \bibinfo{author}{Molchanov, P.}, \bibinfo{author}{Myronenko, A.}, \bibinfo{author}{Li, W.}, \bibinfo{author}{Dogra, P.}, \bibinfo{author}{Feng, A.}, \bibinfo{author}{Flores, M.G.}, \bibinfo{author}{Kautz, J.}, \bibinfo{author}{Xu, D.}, et~al., \bibinfo{year}{2023}.
\newblock \bibinfo{title}{Do gradient inversion attacks make federated learning unsafe?}
\newblock \bibinfo{journal}{IEEE Transactions on Medical Imaging} \bibinfo{volume}{42}, \bibinfo{pages}{2044--2056}.
\bibitem[{Huang et~al.(2021)Huang, Gupta, Song, Li and Arora}]{huang2021evaluating}
\bibinfo{author}{Huang, Y.}, \bibinfo{author}{Gupta, S.}, \bibinfo{author}{Song, Z.}, \bibinfo{author}{Li, K.}, \bibinfo{author}{Arora, S.}, \bibinfo{year}{2021}.
\newblock \bibinfo{title}{Evaluating gradient inversion attacks and defenses in federated learning}.
\newblock \bibinfo{journal}{Advances in neural information processing systems} \bibinfo{volume}{34}, \bibinfo{pages}{7232--7241}.
\bibitem[{Imambi et~al.(2021)Imambi, Prakash and Kanagachidambaresan}]{imambi2021pytorch}
\bibinfo{author}{Imambi, S.}, \bibinfo{author}{Prakash, K.B.}, \bibinfo{author}{Kanagachidambaresan, G.}, \bibinfo{year}{2021}.
\newblock \bibinfo{title}{Pytorch}.
\newblock \bibinfo{journal}{Programming with TensorFlow: solution for edge computing applications} , \bibinfo{pages}{87--104}.
\bibitem[{Jeon et~al.(2021)Jeon, Lee, Oh, Ok et~al.}]{jeon2021gradient}
\bibinfo{author}{Jeon, J.}, \bibinfo{author}{Lee, K.}, \bibinfo{author}{Oh, S.}, \bibinfo{author}{Ok, J.}, et~al., \bibinfo{year}{2021}.
\newblock \bibinfo{title}{Gradient inversion with generative image prior}.
\newblock \bibinfo{journal}{Advances in neural information processing systems} \bibinfo{volume}{34}, \bibinfo{pages}{29898--29908}.
\bibitem[{Karimireddy et~al.(2020)Karimireddy, Kale, Mohri, Reddi, Stich and Suresh}]{karimireddy2020scaffold}
\bibinfo{author}{Karimireddy, S.P.}, \bibinfo{author}{Kale, S.}, \bibinfo{author}{Mohri, M.}, \bibinfo{author}{Reddi, S.}, \bibinfo{author}{Stich, S.}, \bibinfo{author}{Suresh, A.T.}, \bibinfo{year}{2020}.
\newblock \bibinfo{title}{Scaffold: Stochastic controlled averaging for federated learning}, in: \bibinfo{booktitle}{International conference on machine learning}, \bibinfo{organization}{PMLR}. pp. \bibinfo{pages}{5132--5143}.
\bibitem[{Karras et~al.(2021)Karras, Aittala, Laine, H{\"a}rk{\"o}nen, Hellsten, Lehtinen and Aila}]{karras2021alias}
\bibinfo{author}{Karras, T.}, \bibinfo{author}{Aittala, M.}, \bibinfo{author}{Laine, S.}, \bibinfo{author}{H{\"a}rk{\"o}nen, E.}, \bibinfo{author}{Hellsten, J.}, \bibinfo{author}{Lehtinen, J.}, \bibinfo{author}{Aila, T.}, \bibinfo{year}{2021}.
\newblock \bibinfo{title}{Alias-free generative adversarial networks}.
\newblock \bibinfo{journal}{Advances in neural information processing systems} \bibinfo{volume}{34}, \bibinfo{pages}{852--863}.
\bibitem[{LeCun et~al.(2015)LeCun, Bengio and Hinton}]{lecun2015deep}
\bibinfo{author}{LeCun, Y.}, \bibinfo{author}{Bengio, Y.}, \bibinfo{author}{Hinton, G.}, \bibinfo{year}{2015}.
\newblock \bibinfo{title}{Deep learning}.
\newblock \bibinfo{journal}{nature} \bibinfo{volume}{521}, \bibinfo{pages}{436--444}.
\bibitem[{Li et~al.(2022a)Li, Wang, Chen, Zhang, Shafiq and Gu}]{li2022e2egi}
\bibinfo{author}{Li, Z.}, \bibinfo{author}{Wang, L.}, \bibinfo{author}{Chen, G.}, \bibinfo{author}{Zhang, Z.}, \bibinfo{author}{Shafiq, M.}, \bibinfo{author}{Gu, Z.}, \bibinfo{year}{2022}a.
\newblock \bibinfo{title}{E2egi: End-to-end gradient inversion in federated learning}.
\newblock \bibinfo{journal}{IEEE Journal of Biomedical and Health Informatics} \bibinfo{volume}{27}, \bibinfo{pages}{756--767}.
\bibitem[{Li et~al.(2022b)Li, Zhang, Liu and Liu}]{li2022auditing}
\bibinfo{author}{Li, Z.}, \bibinfo{author}{Zhang, J.}, \bibinfo{author}{Liu, L.}, \bibinfo{author}{Liu, J.}, \bibinfo{year}{2022}b.
\newblock \bibinfo{title}{Auditing privacy defenses in federated learning via generative gradient leakage}, in: \bibinfo{booktitle}{Proceedings of the IEEE/CVF Conference on Computer Vision and Pattern Recognition}, pp. \bibinfo{pages}{10132--10142}.
\bibitem[{Liang et~al.(2023)Liang, Li, Zhang, Liu and Zhu}]{liang2023egia}
\bibinfo{author}{Liang, H.}, \bibinfo{author}{Li, Y.}, \bibinfo{author}{Zhang, C.}, \bibinfo{author}{Liu, X.}, \bibinfo{author}{Zhu, L.}, \bibinfo{year}{2023}.
\newblock \bibinfo{title}{Egia: An external gradient inversion attack in federated learning}.
\newblock \bibinfo{journal}{IEEE Transactions on Information Forensics and Security} .
\bibitem[{McMahan et~al.(2017)McMahan, Moore, Ramage, Hampson and y~Arcas}]{mcmahan2017communication}
\bibinfo{author}{McMahan, B.}, \bibinfo{author}{Moore, E.}, \bibinfo{author}{Ramage, D.}, \bibinfo{author}{Hampson, S.}, \bibinfo{author}{y~Arcas, B.A.}, \bibinfo{year}{2017}.
\newblock \bibinfo{title}{Communication-efficient learning of deep networks from decentralized data}, in: \bibinfo{booktitle}{Artificial intelligence and statistics}, \bibinfo{organization}{PMLR}. pp. \bibinfo{pages}{1273--1282}.
\bibitem[{McMahan et~al.(2018)McMahan, Ramage, Talwar and Zhang}]{mcmahan2018learning}
\bibinfo{author}{McMahan, H.B.}, \bibinfo{author}{Ramage, D.}, \bibinfo{author}{Talwar, K.}, \bibinfo{author}{Zhang, L.}, \bibinfo{year}{2018}.
\newblock \bibinfo{title}{Learning differentially private recurrent language models}, in: \bibinfo{booktitle}{International Conference on Learning Representations}.
\bibitem[{Rahman et~al.(2021)Rahman, Khandakar, Qiblawey, Tahir, Kiranyaz, Kashem, Islam, Al~Maadeed, Zughaier, Khan et~al.}]{rahman2021exploring}
\bibinfo{author}{Rahman, T.}, \bibinfo{author}{Khandakar, A.}, \bibinfo{author}{Qiblawey, Y.}, \bibinfo{author}{Tahir, A.}, \bibinfo{author}{Kiranyaz, S.}, \bibinfo{author}{Kashem, S.B.A.}, \bibinfo{author}{Islam, M.T.}, \bibinfo{author}{Al~Maadeed, S.}, \bibinfo{author}{Zughaier, S.M.}, \bibinfo{author}{Khan, M.S.}, et~al., \bibinfo{year}{2021}.
\newblock \bibinfo{title}{Exploring the effect of image enhancement techniques on covid-19 detection using chest x-ray images}.
\newblock \bibinfo{journal}{Computers in biology and medicine} \bibinfo{volume}{132}, \bibinfo{pages}{104319}.
\bibitem[{Sun et~al.(2021)Sun, Li, Wang, Yang, Li and Chen}]{sun2021soteria}
\bibinfo{author}{Sun, J.}, \bibinfo{author}{Li, A.}, \bibinfo{author}{Wang, B.}, \bibinfo{author}{Yang, H.}, \bibinfo{author}{Li, H.}, \bibinfo{author}{Chen, Y.}, \bibinfo{year}{2021}.
\newblock \bibinfo{title}{Soteria: Provable defense against privacy leakage in federated learning from representation perspective}, in: \bibinfo{booktitle}{Proceedings of the IEEE/CVF conference on computer vision and pattern recognition}, pp. \bibinfo{pages}{9311--9319}.
\bibitem[{de~Vente et~al.(2023)de~Vente, Vermeer, Jaccard, Wang, Sun, Khader, Truhn, Aimyshev, Zhanibekuly, Le, Galdran, Gonz\'alez~Ballester, Carneiro, G, S, Puthussery, Liu, Yang, Kondo, Kasai, Wang, Durvasula, Heras, Zapata, Ara\'ujo, Aresta, Bogunovi\'c, Arikan, Lee, Cho, Choi, Qayyum, Razzak, van Ginneken, Lemij and S\'anchez}]{devente23airogs}
\bibinfo{author}{de~Vente, C.}, \bibinfo{author}{Vermeer, K.A.}, \bibinfo{author}{Jaccard, N.}, \bibinfo{author}{Wang, H.}, \bibinfo{author}{Sun, H.}, \bibinfo{author}{Khader, F.}, \bibinfo{author}{Truhn, D.}, \bibinfo{author}{Aimyshev, T.}, \bibinfo{author}{Zhanibekuly, Y.}, \bibinfo{author}{Le, T.D.}, \bibinfo{author}{Galdran, A.}, \bibinfo{author}{Gonz\'alez~Ballester, M.A.}, \bibinfo{author}{Carneiro, G.}, \bibinfo{author}{G, D.R.}, \bibinfo{author}{S, H.P.}, \bibinfo{author}{Puthussery, D.}, \bibinfo{author}{Liu, H.}, \bibinfo{author}{Yang, Z.}, \bibinfo{author}{Kondo, S.}, \bibinfo{author}{Kasai, S.}, \bibinfo{author}{Wang, E.}, \bibinfo{author}{Durvasula, A.}, \bibinfo{author}{Heras, J.}, \bibinfo{author}{Zapata, M.A.}, \bibinfo{author}{Ara\'ujo, T.}, \bibinfo{author}{Aresta, G.}, \bibinfo{author}{Bogunovi\'c, H.}, \bibinfo{author}{Arikan, M.}, \bibinfo{author}{Lee, Y.C.}, \bibinfo{author}{Cho, H.B.}, \bibinfo{author}{Choi, Y.H.}, \bibinfo{author}{Qayyum, A.}, \bibinfo{author}{Razzak, I.}, \bibinfo{author}{van Ginneken, B.}, \bibinfo{author}{Lemij, H.G.}, \bibinfo{author}{S\'anchez, C.I.}, \bibinfo{year}{2023}.
\newblock \bibinfo{title}{Airogs: Artificial intelligence for robust glaucoma screening challenge}.
\newblock \bibinfo{journal}{arXiv preprint arXiv:2302.01738} .
\bibitem[{Wang et~al.(2024a)Wang, Hugh and Li}]{wang2024more}
\bibinfo{author}{Wang, F.}, \bibinfo{author}{Hugh, E.}, \bibinfo{author}{Li, B.}, \bibinfo{year}{2024}a.
\newblock \bibinfo{title}{More than enough is too much: Adaptive defenses against gradient leakage in production federated learning}.
\newblock \bibinfo{journal}{IEEE/ACM Transactions on Networking} .
\bibitem[{Wang et~al.(2024b)Wang, Velipasalar and Gursoy}]{wang2024maximum}
\bibinfo{author}{Wang, F.}, \bibinfo{author}{Velipasalar, S.}, \bibinfo{author}{Gursoy, M.C.}, \bibinfo{year}{2024}b.
\newblock \bibinfo{title}{Maximum knowledge orthogonality reconstruction with gradients in federated learning}, in: \bibinfo{booktitle}{Proceedings of the IEEE/CVF Winter Conference on Applications of Computer Vision}, pp. \bibinfo{pages}{3884--3893}.
\bibitem[{Wang et~al.(2023)Wang, Wang, Jin, Zhang, Hu, Wang, Sun, Yuan, Liu and Ren}]{wang2023privacy}
\bibinfo{author}{Wang, Z.}, \bibinfo{author}{Wang, H.}, \bibinfo{author}{Jin, S.}, \bibinfo{author}{Zhang, W.}, \bibinfo{author}{Hu, J.}, \bibinfo{author}{Wang, Y.}, \bibinfo{author}{Sun, P.}, \bibinfo{author}{Yuan, W.}, \bibinfo{author}{Liu, K.}, \bibinfo{author}{Ren, K.}, \bibinfo{year}{2023}.
\newblock \bibinfo{title}{Privacy-preserving adversarial facial features}, in: \bibinfo{booktitle}{Proceedings of the IEEE/CVF Conference on Computer Vision and Pattern Recognition}, pp. \bibinfo{pages}{8212--8221}.
\bibitem[{Wei et~al.(2021)Wei, Liu, Wu, Su and Iyengar}]{wei2021gradient}
\bibinfo{author}{Wei, W.}, \bibinfo{author}{Liu, L.}, \bibinfo{author}{Wu, Y.}, \bibinfo{author}{Su, G.}, \bibinfo{author}{Iyengar, A.}, \bibinfo{year}{2021}.
\newblock \bibinfo{title}{Gradient-leakage resilient federated learning}, in: \bibinfo{booktitle}{2021 IEEE 41st International Conference on Distributed Computing Systems (ICDCS)}, \bibinfo{organization}{IEEE}. pp. \bibinfo{pages}{797--807}.
\bibitem[{Wu et~al.(2023)Wu, Chen, Guo and Weinberger}]{wu2023learning}
\bibinfo{author}{Wu, R.}, \bibinfo{author}{Chen, X.}, \bibinfo{author}{Guo, C.}, \bibinfo{author}{Weinberger, K.Q.}, \bibinfo{year}{2023}.
\newblock \bibinfo{title}{Learning to invert: Simple adaptive attacks for gradient inversion in federated learning}, in: \bibinfo{booktitle}{Uncertainty in Artificial Intelligence}, \bibinfo{organization}{PMLR}. pp. \bibinfo{pages}{2293--2303}.
\bibitem[{Xu et~al.(2022)Xu, Hong, Huang, Chen and Decouchant}]{xu2022agic}
\bibinfo{author}{Xu, J.}, \bibinfo{author}{Hong, C.}, \bibinfo{author}{Huang, J.}, \bibinfo{author}{Chen, L.Y.}, \bibinfo{author}{Decouchant, J.}, \bibinfo{year}{2022}.
\newblock \bibinfo{title}{Agic: Approximate gradient inversion attack on federated learning}, in: \bibinfo{booktitle}{2022 41st International Symposium on Reliable Distributed Systems (SRDS)}, \bibinfo{organization}{IEEE}. pp. \bibinfo{pages}{12--22}.
\bibitem[{Yin et~al.(2021)Yin, Mallya, Vahdat, Alvarez, Kautz and Molchanov}]{yin2021see}
\bibinfo{author}{Yin, H.}, \bibinfo{author}{Mallya, A.}, \bibinfo{author}{Vahdat, A.}, \bibinfo{author}{Alvarez, J.M.}, \bibinfo{author}{Kautz, J.}, \bibinfo{author}{Molchanov, P.}, \bibinfo{year}{2021}.
\newblock \bibinfo{title}{See through gradients: Image batch recovery via gradinversion}, in: \bibinfo{booktitle}{Proceedings of the IEEE/CVF conference on computer vision and pattern recognition}, pp. \bibinfo{pages}{16337--16346}.
\bibitem[{Yue et~al.(2023)Yue, Yang, Zhao, An and Yang}]{yue2023ergpnet}
\bibinfo{author}{Yue, G.}, \bibinfo{author}{Yang, C.}, \bibinfo{author}{Zhao, Z.}, \bibinfo{author}{An, Z.}, \bibinfo{author}{Yang, Y.}, \bibinfo{year}{2023}.
\newblock \bibinfo{title}{Ergpnet: lesion segmentation network for covid-19 chest x-ray images based on embedded residual convolution and global perception}.
\newblock \bibinfo{journal}{Frontiers in Physiology} \bibinfo{volume}{14}, \bibinfo{pages}{1296185}.
\bibitem[{Zhang et~al.(2022)Zhang, Ekanut, Zhen and Li}]{zhang2022augmented}
\bibinfo{author}{Zhang, C.}, \bibinfo{author}{Ekanut, S.}, \bibinfo{author}{Zhen, L.}, \bibinfo{author}{Li, Z.}, \bibinfo{year}{2022}.
\newblock \bibinfo{title}{Augmented multi-party computation against gradient leakage in federated learning}.
\newblock \bibinfo{journal}{IEEE Transactions on Big Data} .
\bibitem[{Zhang et~al.(2023)Zhang, Xiaoman, Sotthiwat, Xu, Liu, Zhen and Liu}]{zhang2023generative}
\bibinfo{author}{Zhang, C.}, \bibinfo{author}{Xiaoman, Z.}, \bibinfo{author}{Sotthiwat, E.}, \bibinfo{author}{Xu, Y.}, \bibinfo{author}{Liu, P.}, \bibinfo{author}{Zhen, L.}, \bibinfo{author}{Liu, Y.}, \bibinfo{year}{2023}.
\newblock \bibinfo{title}{Generative gradient inversion via over-parameterized networks in federated learning}, in: \bibinfo{booktitle}{Proceedings of the IEEE/CVF International Conference on Computer Vision}, pp. \bibinfo{pages}{5126--5135}.
\bibitem[{Zhang et~al.(2025)Zhang, Cheng, Shen, Ribeiro, An, Chen, Zhang and Li}]{zhang2025censor}
\bibinfo{author}{Zhang, K.}, \bibinfo{author}{Cheng, S.}, \bibinfo{author}{Shen, G.}, \bibinfo{author}{Ribeiro, B.}, \bibinfo{author}{An, S.}, \bibinfo{author}{Chen, P.Y.}, \bibinfo{author}{Zhang, X.}, \bibinfo{author}{Li, N.}, \bibinfo{year}{2025}.
\newblock \bibinfo{title}{Censor: Defense against gradient inversion via orthogonal subspace bayesian sampling}.
\newblock \bibinfo{journal}{arXiv preprint arXiv:2501.15718} .
\bibitem[{Zhao et~al.(2020)Zhao, Mopuri and Bilen}]{zhao2020idlg}
\bibinfo{author}{Zhao, B.}, \bibinfo{author}{Mopuri, K.R.}, \bibinfo{author}{Bilen, H.}, \bibinfo{year}{2020}.
\newblock \bibinfo{title}{idlg: Improved deep leakage from gradients}.
\newblock \bibinfo{journal}{arXiv preprint arXiv:2001.02610} .
\bibitem[{Zhu and Blaschko()}]{citation-0}
\bibinfo{author}{Zhu, J.}, \bibinfo{author}{Blaschko, M.B.}, .
\newblock \bibinfo{title}{R-gap: Recursive gradient attack on privacy}, in: \bibinfo{booktitle}{International Conference on Learning Representations}.
\bibitem[{Zhu et~al.(2023)Zhu, Yao and Blaschko}]{zhu2023surrogate}
\bibinfo{author}{Zhu, J.}, \bibinfo{author}{Yao, R.}, \bibinfo{author}{Blaschko, M.}, \bibinfo{year}{2023}.
\newblock \bibinfo{title}{Surrogate model extension (sme): A fast and accurate weight update attack on federated learning}, in: \bibinfo{booktitle}{ICML'23: Proceedings of the 40th International Conference on Machine Learning}, \bibinfo{organization}{JMLR. org}. pp. \bibinfo{pages}{43228--43257}.
\bibitem[{Zhu et~al.(2019)Zhu, Liu and Han}]{zhu2019deep}
\bibinfo{author}{Zhu, L.}, \bibinfo{author}{Liu, Z.}, \bibinfo{author}{Han, S.}, \bibinfo{year}{2019}.
\newblock \bibinfo{title}{Deep leakage from gradients}.
\newblock \bibinfo{journal}{Advances in neural information processing systems} \bibinfo{volume}{32}.

\end{thebibliography}


\begin{thebibliography}{11}
\expandafter\ifx\csname natexlab\endcsname\relax\def\natexlab#1{#1}\fi
\providecommand{\url}[1]{\texttt{#1}}
\providecommand{\href}[2]{#2}
\providecommand{\path}[1]{#1}
\providecommand{\DOIprefix}{doi:}
\providecommand{\ArXivprefix}{arXiv:}
\providecommand{\URLprefix}{URL: }
\providecommand{\Pubmedprefix}{pmid:}
\providecommand{\doi}[1]{\href{http://dx.doi.org/#1}{\path{#1}}}
\providecommand{\Pubmed}[1]{\href{pmid:#1}{\path{#1}}}
\providecommand{\bibinfo}[2]{#2}
\ifx\xfnm\relax \def\xfnm[#1]{\unskip,\space#1}\fi
\bibitem[{Adams et~al.(2022)Adams, Makowski, Engel, Rattunde, Busch, Asbach, Niehues, Vinayahalingam, van Ginneken, Litjens et~al.}]{adams2022prostate158}
\bibinfo{author}{Adams, L.C.}, \bibinfo{author}{Makowski, M.R.}, \bibinfo{author}{Engel, G.}, \bibinfo{author}{Rattunde, M.}, \bibinfo{author}{Busch, F.}, \bibinfo{author}{Asbach, P.}, \bibinfo{author}{Niehues, S.M.}, \bibinfo{author}{Vinayahalingam, S.}, \bibinfo{author}{van Ginneken, B.}, \bibinfo{author}{Litjens, G.}, et~al., \bibinfo{year}{2022}.
\newblock \bibinfo{title}{Prostate158-an expert-annotated 3t mri dataset and algorithm for prostate cancer detection}.
\newblock \bibinfo{journal}{Computers in Biology and Medicine} \bibinfo{volume}{148}, \bibinfo{pages}{105817}.
\bibitem[{Cao et~al.(2018)Cao, Shen, Xie, Parkhi and Zisserman}]{cao2018vggface2}
\bibinfo{author}{Cao, Q.}, \bibinfo{author}{Shen, L.}, \bibinfo{author}{Xie, W.}, \bibinfo{author}{Parkhi, O.M.}, \bibinfo{author}{Zisserman, A.}, \bibinfo{year}{2018}.
\newblock \bibinfo{title}{Vggface2: A dataset for recognising faces across pose and age}, in: \bibinfo{booktitle}{2018 13th IEEE international conference on automatic face \& gesture recognition (FG 2018)}, \bibinfo{organization}{IEEE}. pp. \bibinfo{pages}{67--74}.
\bibitem[{Han et~al.(2022)Han, Wang, Chen, Chen, Guo, Liu, Tang, Xiao, Xu, Xu et~al.}]{han2022survey}
\bibinfo{author}{Han, K.}, \bibinfo{author}{Wang, Y.}, \bibinfo{author}{Chen, H.}, \bibinfo{author}{Chen, X.}, \bibinfo{author}{Guo, J.}, \bibinfo{author}{Liu, Z.}, \bibinfo{author}{Tang, Y.}, \bibinfo{author}{Xiao, A.}, \bibinfo{author}{Xu, C.}, \bibinfo{author}{Xu, Y.}, et~al., \bibinfo{year}{2022}.
\newblock \bibinfo{title}{A survey on vision transformer}.
\newblock \bibinfo{journal}{IEEE transactions on pattern analysis and machine intelligence} \bibinfo{volume}{45}, \bibinfo{pages}{87--110}.
\bibitem[{Hatamizadeh et~al.(2023)Hatamizadeh, Yin, Molchanov, Myronenko, Li, Dogra, Feng, Flores, Kautz, Xu et~al.}]{hatamizadeh2023gradient}
\bibinfo{author}{Hatamizadeh, A.}, \bibinfo{author}{Yin, H.}, \bibinfo{author}{Molchanov, P.}, \bibinfo{author}{Myronenko, A.}, \bibinfo{author}{Li, W.}, \bibinfo{author}{Dogra, P.}, \bibinfo{author}{Feng, A.}, \bibinfo{author}{Flores, M.G.}, \bibinfo{author}{Kautz, J.}, \bibinfo{author}{Xu, D.}, et~al., \bibinfo{year}{2023}.
\newblock \bibinfo{title}{Do gradient inversion attacks make federated learning unsafe?}
\newblock \bibinfo{journal}{IEEE Transactions on Medical Imaging} \bibinfo{volume}{42}, \bibinfo{pages}{2044--2056}.
\bibitem[{Jiang et~al.(2022)Jiang, Wang and Dou}]{jiang2022harmofl}
\bibinfo{author}{Jiang, M.}, \bibinfo{author}{Wang, Z.}, \bibinfo{author}{Dou, Q.}, \bibinfo{year}{2022}.
\newblock \bibinfo{title}{Harmofl: Harmonizing local and global drifts in federated learning on heterogeneous medical images}, in: \bibinfo{booktitle}{Proceedings of the AAAI Conference on Artificial Intelligence}, pp. \bibinfo{pages}{1087--1095}.
\bibitem[{Lema{\^\i}tre et~al.(2015)Lema{\^\i}tre, Mart{\'\i}, Freixenet, Vilanova, Walker and Meriaudeau}]{lemaitre2015computer}
\bibinfo{author}{Lema{\^\i}tre, G.}, \bibinfo{author}{Mart{\'\i}, R.}, \bibinfo{author}{Freixenet, J.}, \bibinfo{author}{Vilanova, J.C.}, \bibinfo{author}{Walker, P.M.}, \bibinfo{author}{Meriaudeau, F.}, \bibinfo{year}{2015}.
\newblock \bibinfo{title}{Computer-aided detection and diagnosis for prostate cancer based on mono and multi-parametric mri: a review}.
\newblock \bibinfo{journal}{Computers in biology and medicine} \bibinfo{volume}{60}, \bibinfo{pages}{8--31}.
\bibitem[{Litjens et~al.(2014)Litjens, Toth, Van De~Ven, Hoeks, Kerkstra, Van~Ginneken, Vincent, Guillard, Birbeck, Zhang et~al.}]{litjens2014evaluation}
\bibinfo{author}{Litjens, G.}, \bibinfo{author}{Toth, R.}, \bibinfo{author}{Van De~Ven, W.}, \bibinfo{author}{Hoeks, C.}, \bibinfo{author}{Kerkstra, S.}, \bibinfo{author}{Van~Ginneken, B.}, \bibinfo{author}{Vincent, G.}, \bibinfo{author}{Guillard, G.}, \bibinfo{author}{Birbeck, N.}, \bibinfo{author}{Zhang, J.}, et~al., \bibinfo{year}{2014}.
\newblock \bibinfo{title}{Evaluation of prostate segmentation algorithms for mri: the promise12 challenge}.
\newblock \bibinfo{journal}{Medical image analysis} \bibinfo{volume}{18}, \bibinfo{pages}{359--373}.
\bibitem[{Nicholas et~al.(2015)Nicholas, Anant, Henkjan, John, Justin et~al.}]{nicholas2015nci}
\bibinfo{author}{Nicholas, B.}, \bibinfo{author}{Anant, M.}, \bibinfo{author}{Henkjan, H.}, \bibinfo{author}{John, F.}, \bibinfo{author}{Justin, K.}, et~al., \bibinfo{year}{2015}.
\newblock \bibinfo{title}{Nci-proc. ieee-isbi conf. 2013 challenge: Automated segmentation of prostate structures}.
\newblock \bibinfo{journal}{The Cancer Imaging Archive} \bibinfo{volume}{5}.
\bibitem[{Wang et~al.(2024)Wang, Velipasalar and Gursoy}]{wang2024maximum}
\bibinfo{author}{Wang, F.}, \bibinfo{author}{Velipasalar, S.}, \bibinfo{author}{Gursoy, M.C.}, \bibinfo{year}{2024}.
\newblock \bibinfo{title}{Maximum knowledge orthogonality reconstruction with gradients in federated learning}, in: \bibinfo{booktitle}{Proceedings of the IEEE/CVF Winter Conference on Applications of Computer Vision}, pp. \bibinfo{pages}{3884--3893}.
\bibitem[{Zhang et~al.(2023)Zhang, Xiaoman, Sotthiwat, Xu, Liu, Zhen and Liu}]{zhang2023generative}
\bibinfo{author}{Zhang, C.}, \bibinfo{author}{Xiaoman, Z.}, \bibinfo{author}{Sotthiwat, E.}, \bibinfo{author}{Xu, Y.}, \bibinfo{author}{Liu, P.}, \bibinfo{author}{Zhen, L.}, \bibinfo{author}{Liu, Y.}, \bibinfo{year}{2023}.
\newblock \bibinfo{title}{Generative gradient inversion via over-parameterized networks in federated learning}, in: \bibinfo{booktitle}{Proceedings of the IEEE/CVF International Conference on Computer Vision}, pp. \bibinfo{pages}{5126--5135}.
\bibitem[{Zhao et~al.(2024)Zhao, Sharma, Elkordy, Ezzeldin, Avestimehr and Bagchi}]{zhao2024loki}
\bibinfo{author}{Zhao, J.C.}, \bibinfo{author}{Sharma, A.}, \bibinfo{author}{Elkordy, A.R.}, \bibinfo{author}{Ezzeldin, Y.H.}, \bibinfo{author}{Avestimehr, S.}, \bibinfo{author}{Bagchi, S.}, \bibinfo{year}{2024}.
\newblock \bibinfo{title}{Loki: Large-scale data reconstruction attack against federated learning through model manipulation}, in: \bibinfo{booktitle}{2024 IEEE Symposium on Security and Privacy (SP)}, \bibinfo{organization}{IEEE}. pp. \bibinfo{pages}{1287--1305}.

\end{thebibliography}

\end{document}



\verso{Le Jiang \textit{et~al.}}

\begin{frontmatter}

\title{Supplementary Materials for: Shadow defense against gradient inversion attack in federated learning}%

\author[a]{Le Jiang\fnref{fn1}}
\author[b]{Liyan Ma\fnref{fn1}\corref{cor1}}
\author[a]{Guang Yang} 

\address[a]{Bioengineering Department, Imperial College London, London W12 0BZ, UK}
\address[b]{School of Computer Engineering and Science, Shanghai university, Shanghai 200444, China}


\begin{abstract}
\end{abstract}

\end{frontmatter}



\section{Additional results for SOTA comparison}   \label{Additional results for SOTA comparison}
\begin{table*}[]
	\caption{Computational cost comparison of our method with SOTA defense methods on the EyePACS dataset.}
    \centering
	\resizebox{2\columnwidth}{!}{
		\begin{tabular}{llllllll}
      \hline
      Method &
        FedAvg &
        DP &
        GS &
        GC &
        Soteria &
        OUTPOST &
        Ours \\ \hline
      Time(s) &
        \multicolumn{1}{l}{566.436} &
        \multicolumn{1}{l}{698.794} &
        \multicolumn{1}{l}{699.565} &
        \multicolumn{1}{l}{703.01} &
        \multicolumn{1}{l}{1581.399} &
        \multicolumn{1}{l}{705.139} &
        \multicolumn{1}{l}{720.711} \\ \hline
      \end{tabular}
	}
  \label{table 1}
\end{table*}

Computational costs on the EyePACS dataset of SOTA methods are listed in Table \ref{table 1}. Each value is a mean of all training epochs. 

In Table \ref{table 2}, we use Image Identifiability Precision (IIP) \cite{hatamizadeh2023gradient} to quantify how much identifiable information is leaked through gradient inversion attacks in federated learning. It measures whether reconstructed images from such attacks can be matched back to the original training data of a specific client. Specifically, IIP computes the fraction of reconstructions whose closest match (based on deep feature embeddings and cosine similarity) in the training set is the exact original image used during training.The numbers in 1-IIP, 3-IIP, and 5-IIP refer to how many of the top-k closest matches are considered for determining a successful identification. For example: 1-IIP only counts a reconstruction as successful if the closest match is the original image. 3-IIP allows a match if the original image is among the top 3 closest. 5-IIP allows it among the top 5. Our method still achieves approximately best defense on these metrics.

\begin{figure*}
	\begin{center}
	\includegraphics[width=2\columnwidth]{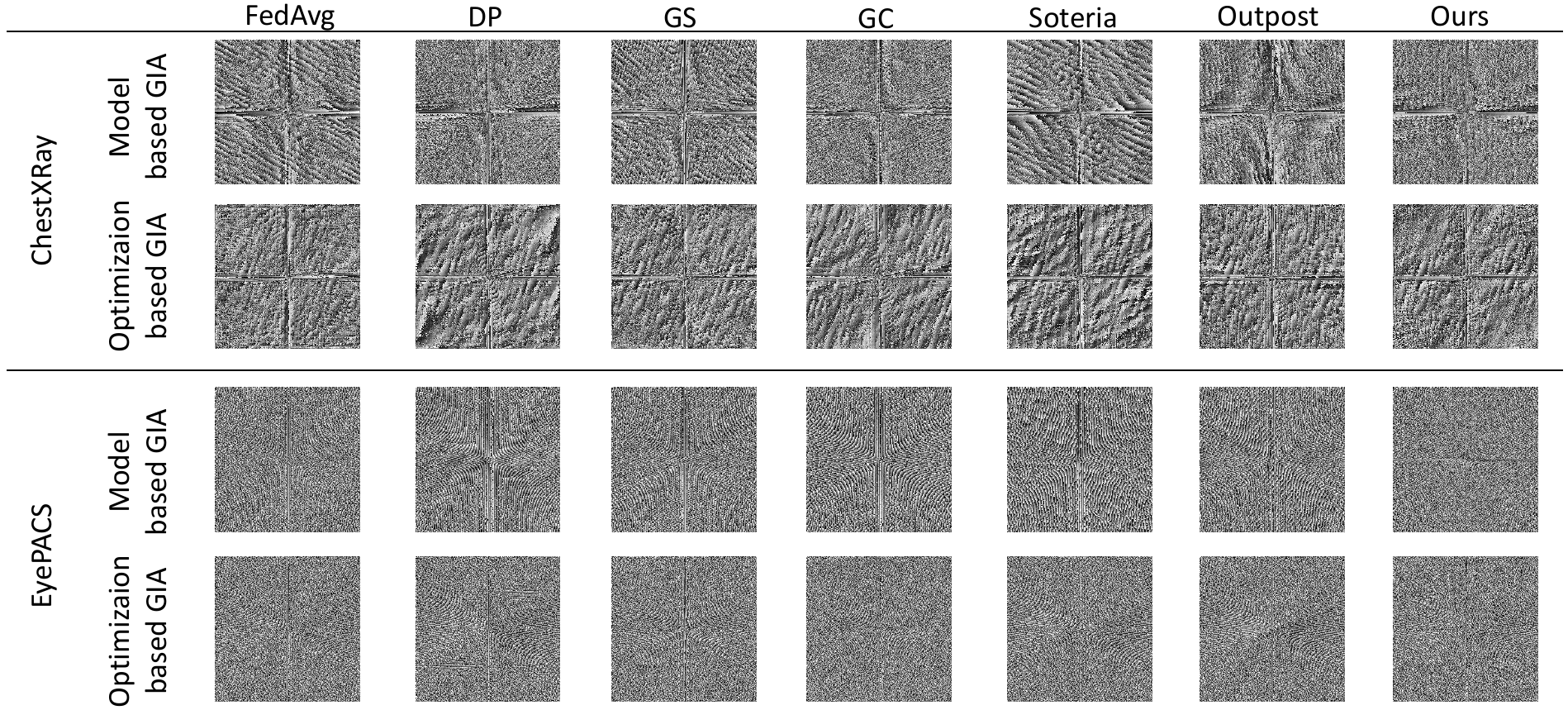}
	\end{center}
	\caption{Comparison of FFT phase for 'Rec Diff' in Fig.9 and Fig.10 of the main text. }
	\label{fig S1}
\end{figure*}

Apart from amplitude spectrum of 'Rec Diff' in Fig.9 and Fig.10 of the main text, we also show their FFT phase in \textcolor{SkyBlue}{Fig.} \ref{fig S1}. Compared with other methods, 'FFT Diff' phase of of our method can be seen as noise without clear structure or clear directional distortion. It validates that our proposed framework successfully suppresses structural information leakage.

\begin{table*}[]
	\caption{IIP comparison on the ChestXRay dataset for model-based GIA.}
    \centering
	\resizebox{2\columnwidth}{!}{
	\begin{tabular}{lllllllll}
		\hline
		& FedAvg & DP     & GS     & GC     & Soteria & OUTPOST & Censor & Ours   \\
		\hline
		IIP-1 & 0.0191 & 0.0128 & 0.0111 & 0.0093 & 0.0174  & 0.0111  & 0.0161 & 0.0059 \\
IIP-3 & 0.0402 & 0.0305 & 0.0319 & 0.0277 & 0.0333  & 0.0562  & 0.3534 & 0.0279 \\
IIP-5 & 0.0583 & 0.0465 & 0.0423 & 0.0437 & 0.0555  & 0.0771  & 0.0462 & 0.0434\\
		\hline
	\end{tabular}}
  \label{table 2}
\end{table*}

\begin{figure*}
	\begin{center}
	\includegraphics[width=2\columnwidth]{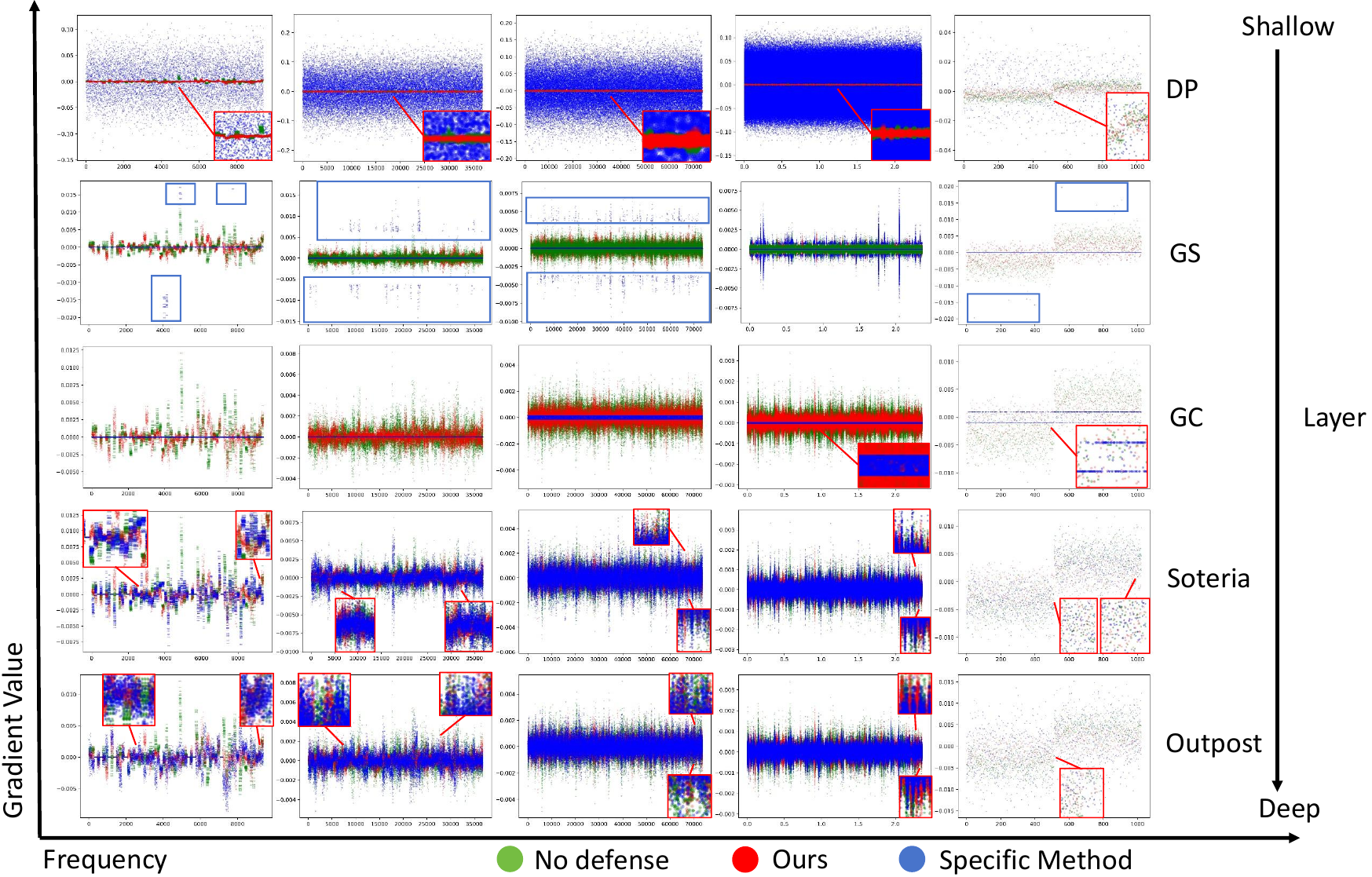}
	\end{center}
	\caption{Comparison of gradient distribution for difference methods across model layers. For each subfigure, the x-axis corresponds to gradient values while the y-axis corresponds to frequency of gradient values. The first to the last rows correspond to the shallowest to the deepest. }
	\label{fig S2}
\end{figure*}

\begin{figure*}
	\begin{center}
	\includegraphics[width=2\columnwidth]{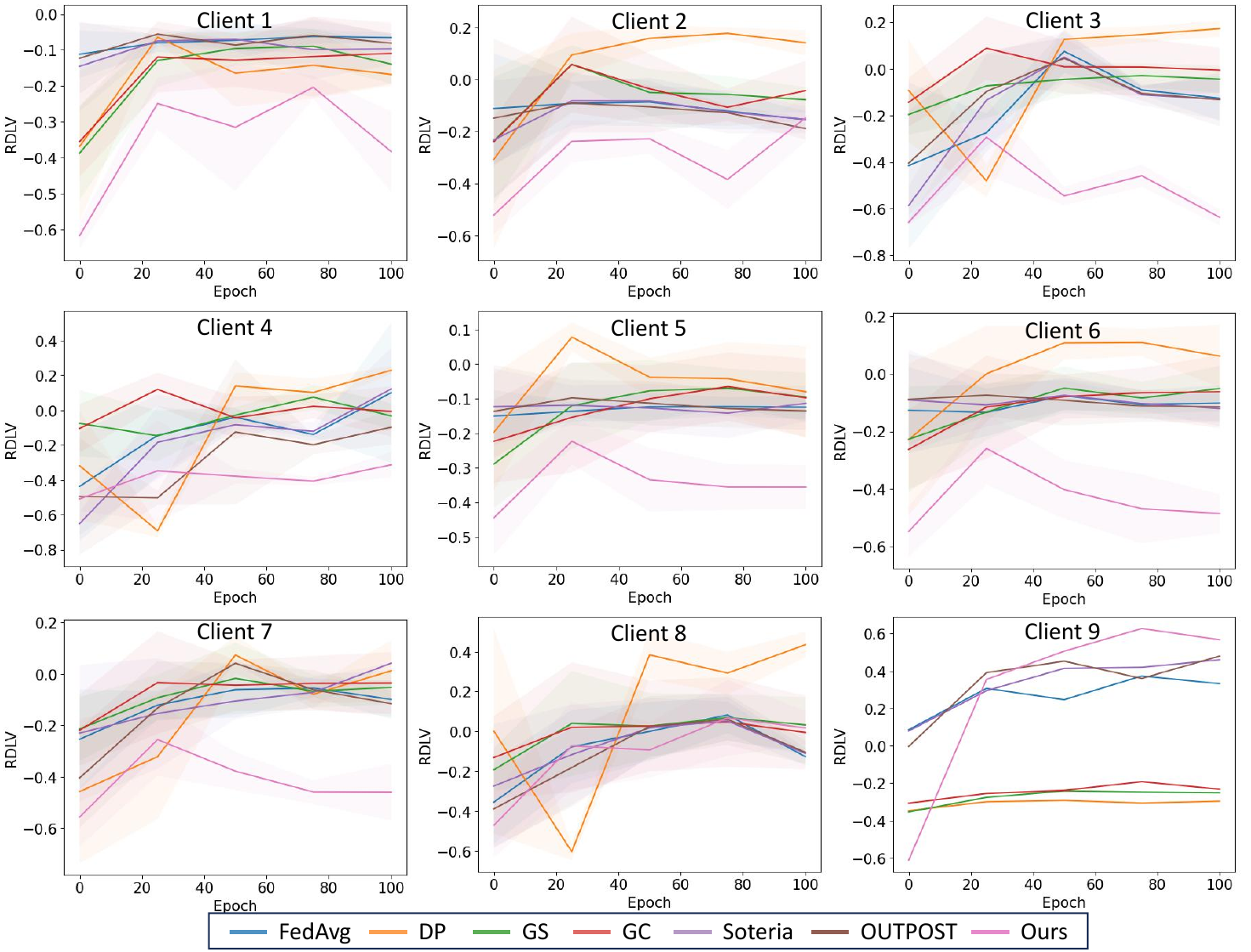}
	\end{center}
	\caption{Comparison of RDLV curve for difference methods on all clients under optimization-based GIA.}
	\label{fig S3}
\end{figure*}

In \textcolor{SkyBlue}{Fig.} \ref{fig S2}, we demonstrate gradient distributions of defensive methods. For conventional strategies, i.e., DP, GS, GC, their characteristics are obvious. For example, perturbed distribution from DP is uniformly across the entire value space. Only partial points from GS are laterally spread, while ones from GC are close to the origin of y-axis. However, for the rest methods, their distributions resemble to each other. It is unlikely to show statistics of privacy leakage directly through them. Thus, it is significant for our image-wise privacy illustration.

As illustrated in \textcolor{SkyBlue}{Fig.} \ref{fig S3}, for GIA based on optimization, the defense of gradient sparsification and gradient clipping is weaker when compared to model-based GIA. DP demonstrates inconsistent defensive performance across different clients. Soteria and OUTPOST exhibit advantages only on clients with larger datasets, specifically clients 4 and 8. With the exception of client 9, our method results in a RDLV that is nearly equal to or less than 0 for all other clients. It aligns with findings from model-based GIA that shadow models are prone to overfitting in extreme cases where only a single image is available.

\begin{figure*}
	\begin{center}
	\includegraphics[width=2\columnwidth]{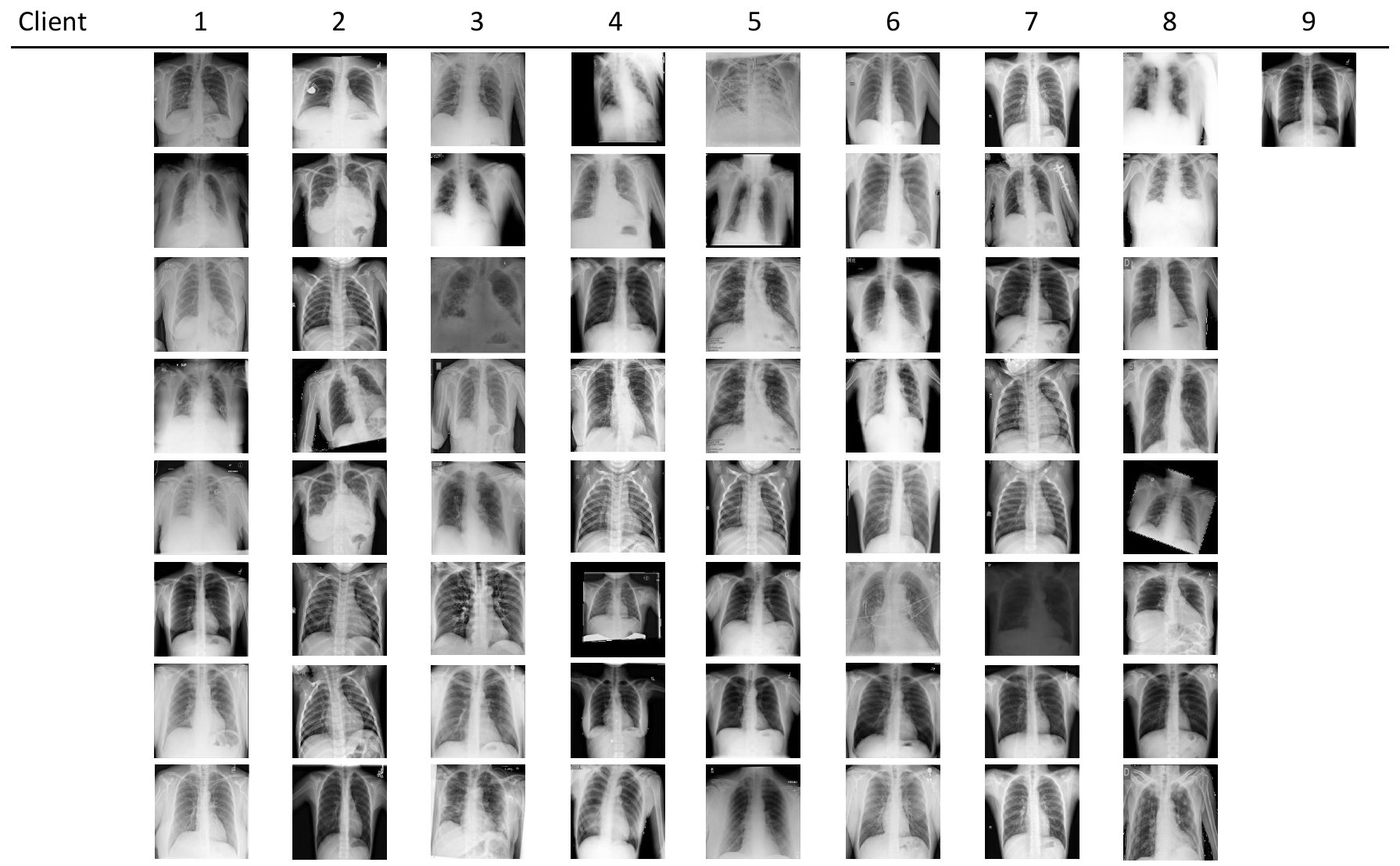}
	\end{center}
	\caption{Vulnerable images in the ChestXRay dataset.}
	\label{fig S4}
\end{figure*}

\begin{figure*}
	\begin{center}
	\includegraphics[width=2\columnwidth]{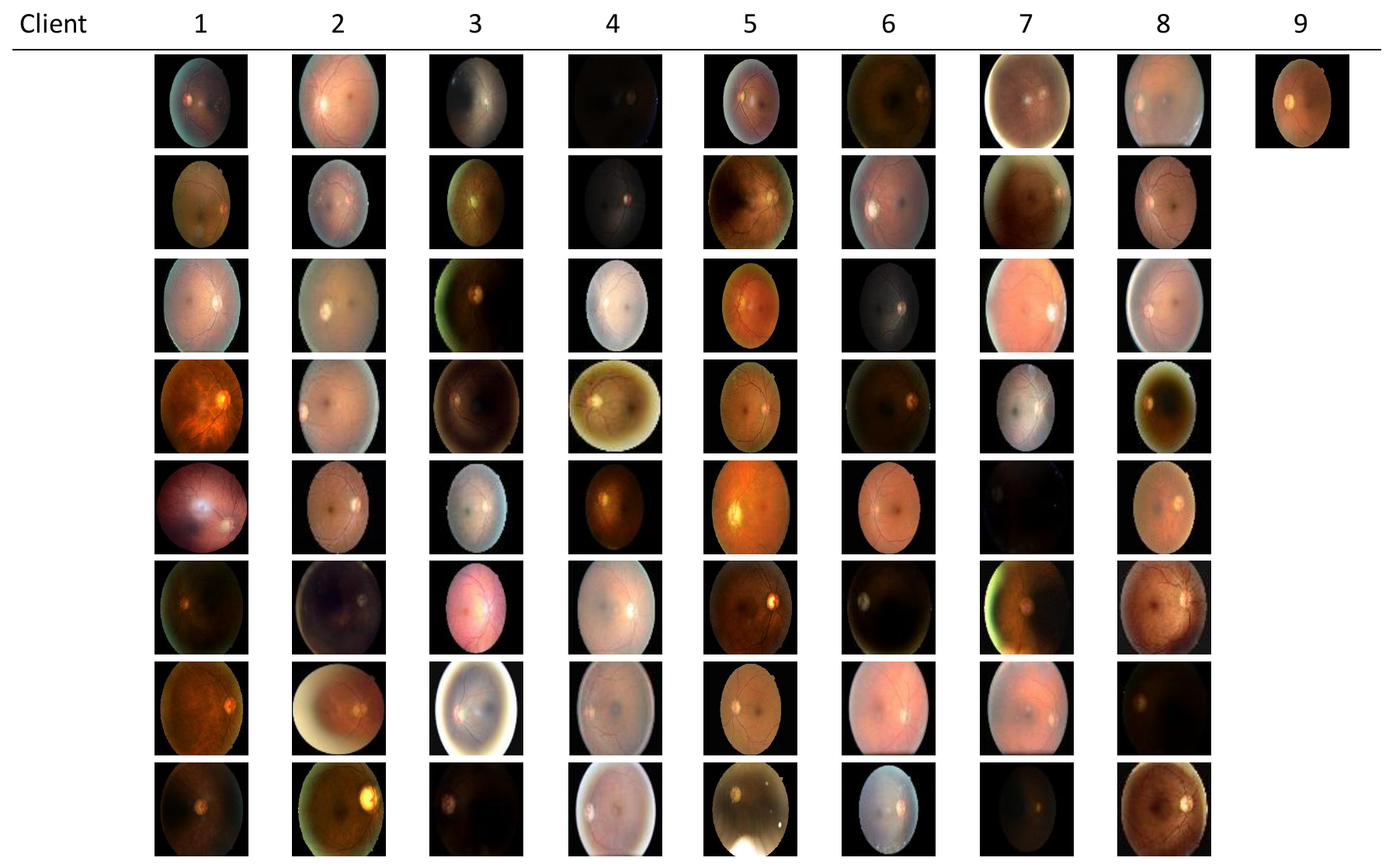}
	\end{center}
	\caption{Vulnerable images in the EyePACS dataset.}
	\label{fig S5}
\end{figure*}

\textcolor{SkyBlue}{Fig.} \ref{fig S4} and \textcolor{SkyBlue}{Fig.} \ref{fig S5} show images from the ChestXRay and EyePACS datasets that are vulnerable to GIA, respectively. Compared with other images, they are more likely to resemble reconstructed images across various training epochs.

\begin{figure*}
	\begin{center}
	\includegraphics[width=2\columnwidth]{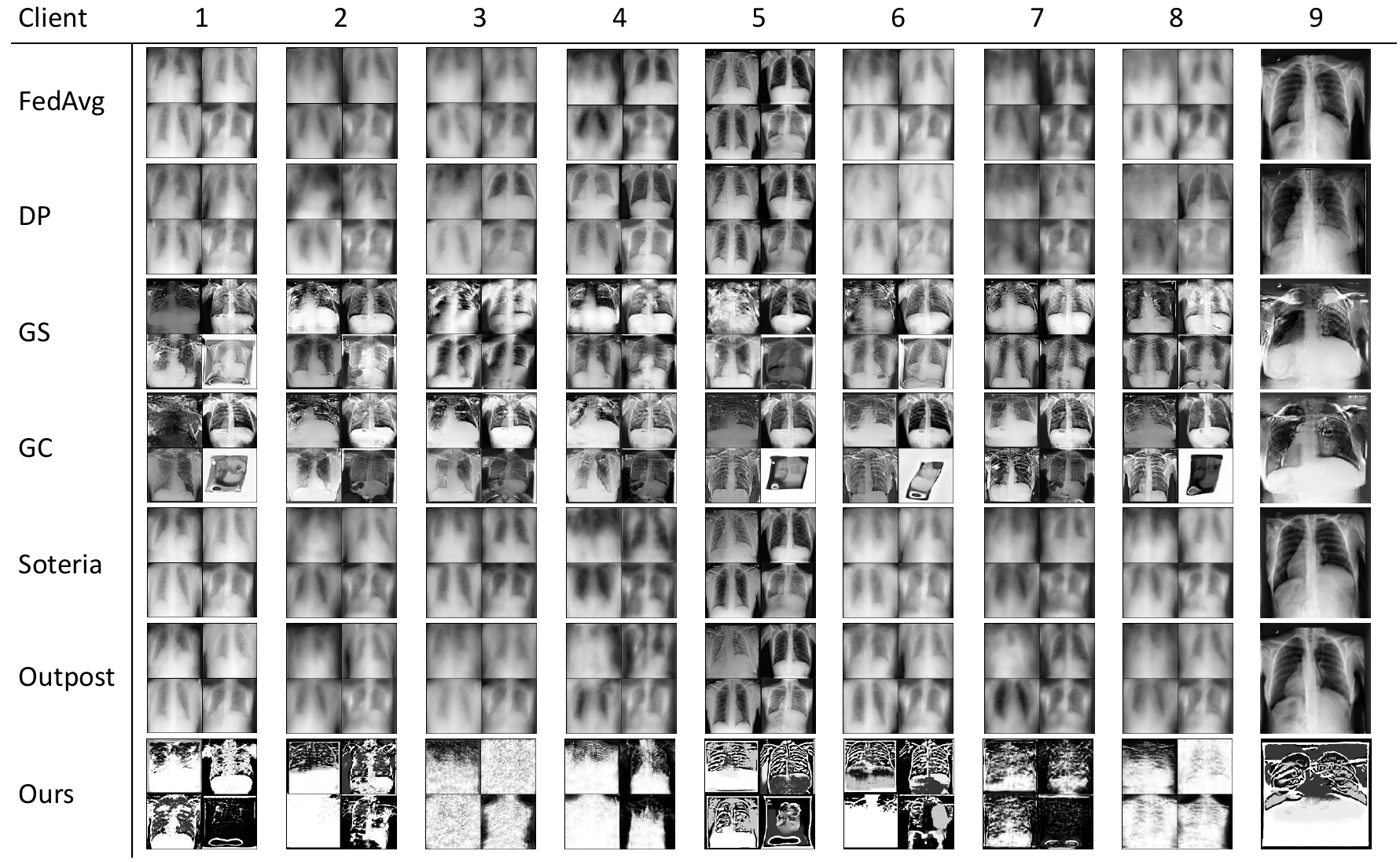}
	\end{center}
	\caption{Reconstructed images from model-based GIA at 1st global round on the ChestXRay dataset.}
	\label{fig S6}
\end{figure*}

\begin{figure*}
	\begin{center}
	\includegraphics[width=2\columnwidth]{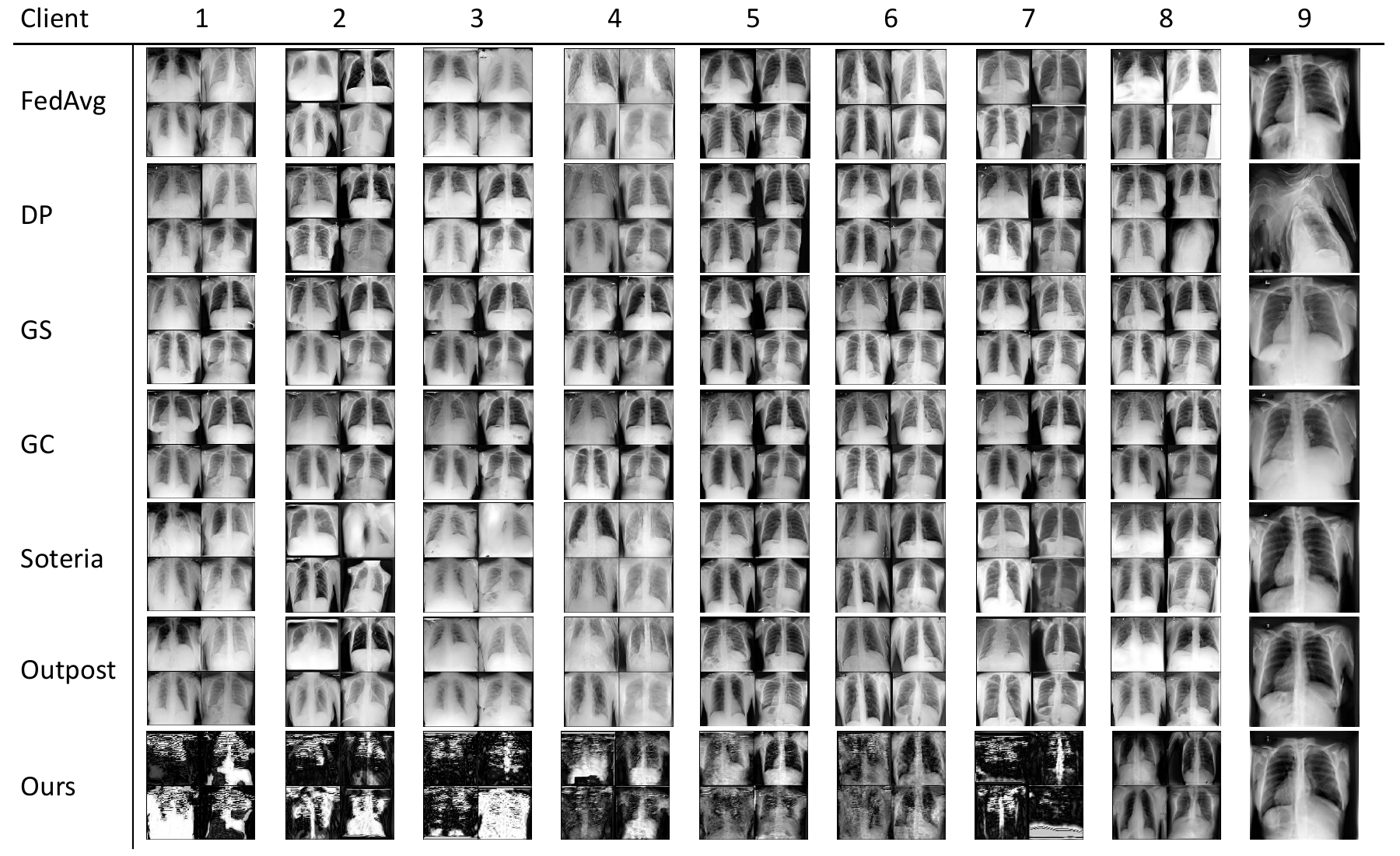}
	\end{center}
	\caption{Reconstructed images from model-based GIA at 50th global round on the ChestXRay dataset.}
	\label{fig S7}
\end{figure*}

\begin{figure*}
	\begin{center}
	\includegraphics[width=2\columnwidth]{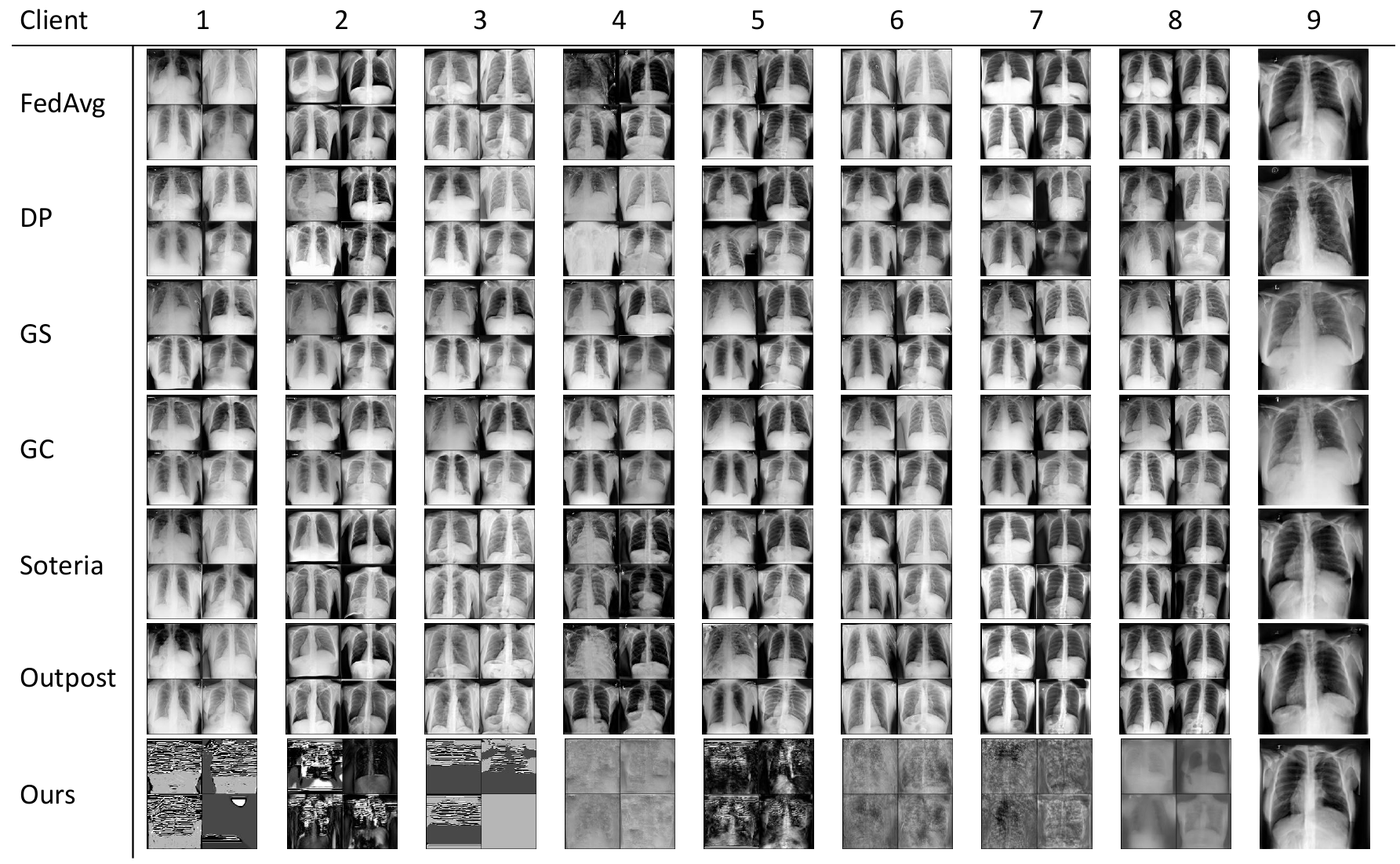}
	\end{center}
	\caption{Reconstructed images from model-based GIA at 100th global round on the ChestXRay dataset.}
	\label{fig S8}
\end{figure*}

\begin{figure*}
	\begin{center}
	\includegraphics[width=2\columnwidth]{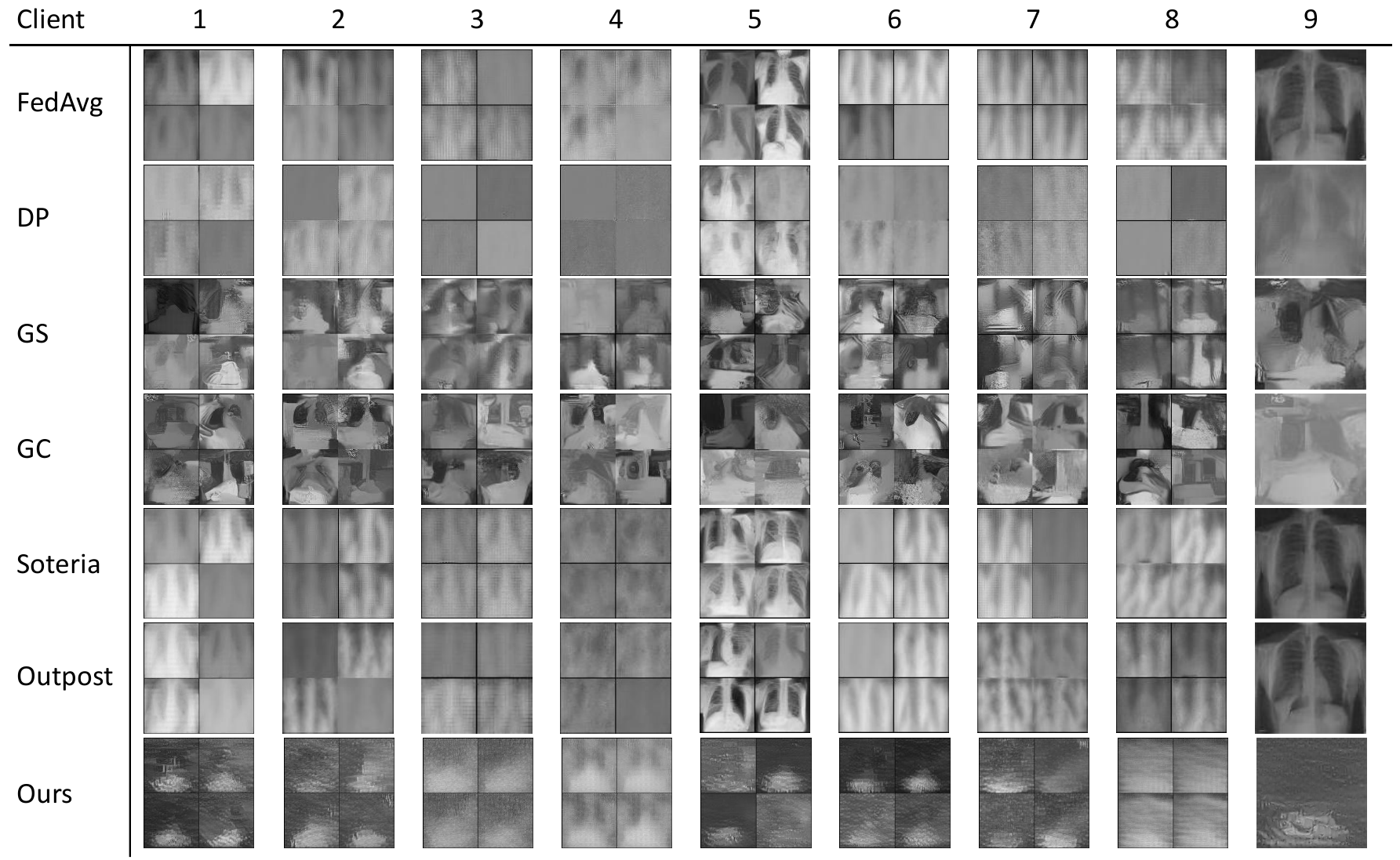}
	\end{center}
	\caption{Reconstructed images from optimization-based GIA at 1st global round on the ChestXRay dataset.}
	\label{fig S9}
\end{figure*}

\begin{figure*}
	\begin{center}
	\includegraphics[width=2\columnwidth]{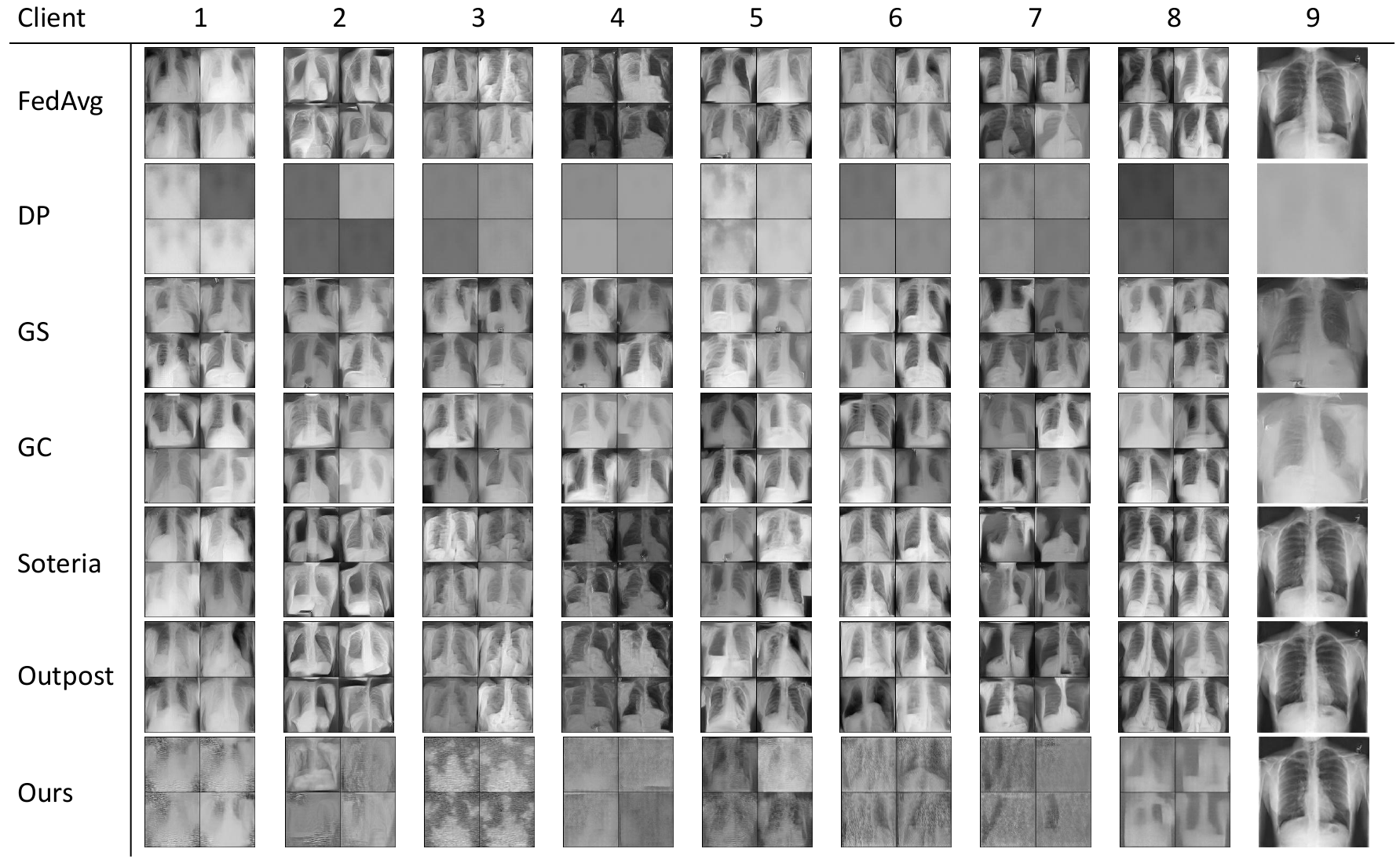}
	\end{center}
	\caption{Reconstructed images from optimization-based GIA at 100th global round on the ChestXRay dataset.}
	\label{fig S10}
\end{figure*}

\textcolor{SkyBlue}{Fig.} \ref{fig S6}, \ref{fig S7}, \ref{fig S8} depict reconstruction results of the ChestXRay dataset subjected to model-based GIA during the 1st, 50th, and 100th FL rounds. \textcolor{SkyBlue}{Fig.} \ref{fig S9}, \ref{fig S10} illustrate the ones for this dataset following optimization-based GIA at the 1st and 100th FL rounds. It can be concluded that although gradients are more substantial in early training stages, aiding the computation of the gradient matching loss, the BN statistics are not accurate, thus limiting the reconstruction efficacy for both types of GIA. For model-based GIA, due to part alignment of these prior information with the pre-training dataset for StyleGAN3, most SOTA methods suffer significant privacy leakage across the majority of client datasets. Our proposed method stands as an exception, effectively defending against such attacks for all but client 9's dataset. For optimization-based GIA, early training stages also present challenges for effective attacks. However, by later stages, except for DP and our method, other SOTA methods tend to leak structural information of the data to varying degrees.

\begin{figure*}
	\begin{center}
	\includegraphics[width=2\columnwidth]{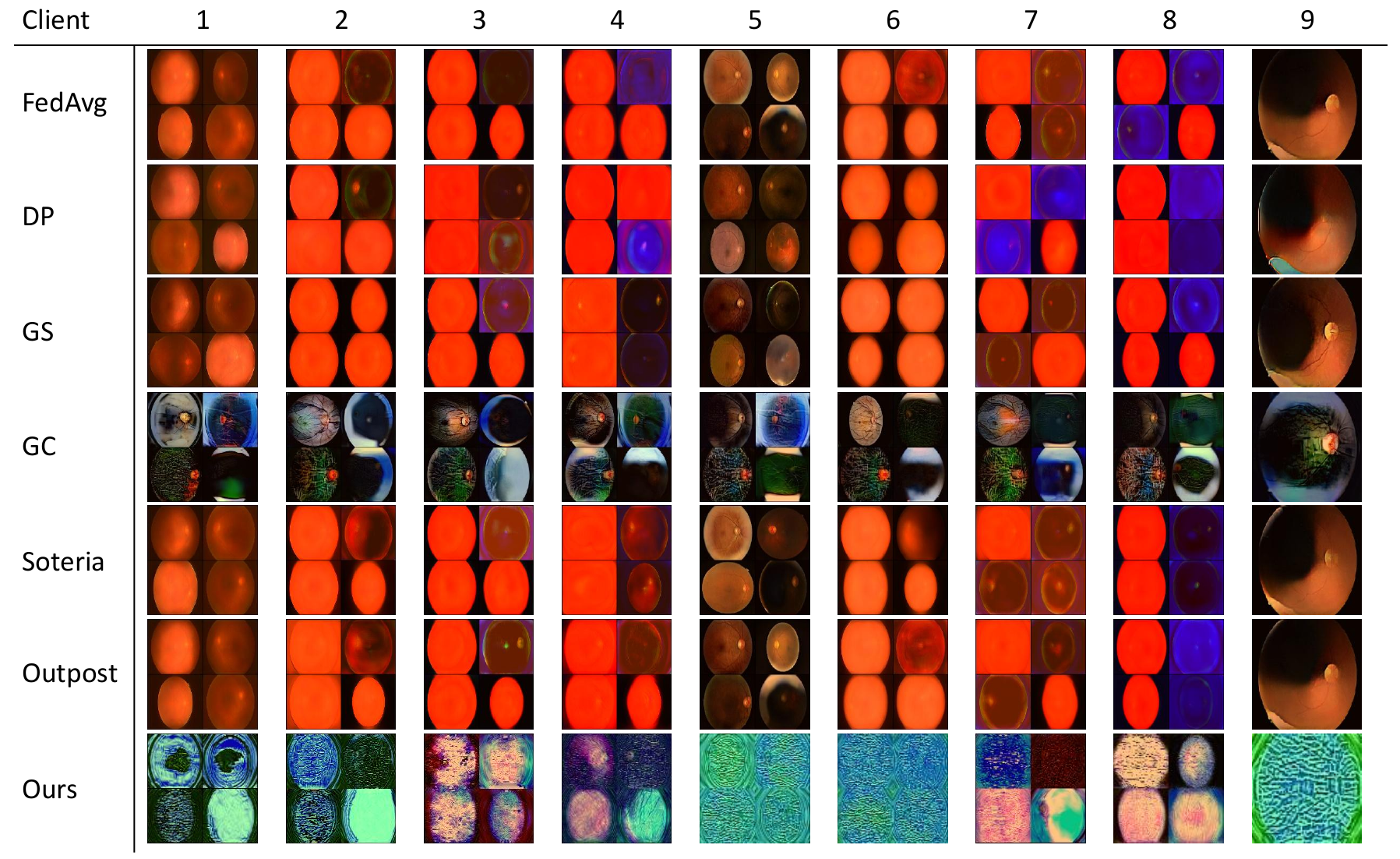}
	\end{center}
	\caption{Reconstructed images from model-based GIA at 1st global round on the EyePACS dataset.}
	\label{fig S11}
\end{figure*}

\begin{figure*}
	\begin{center}
	\includegraphics[width=2\columnwidth]{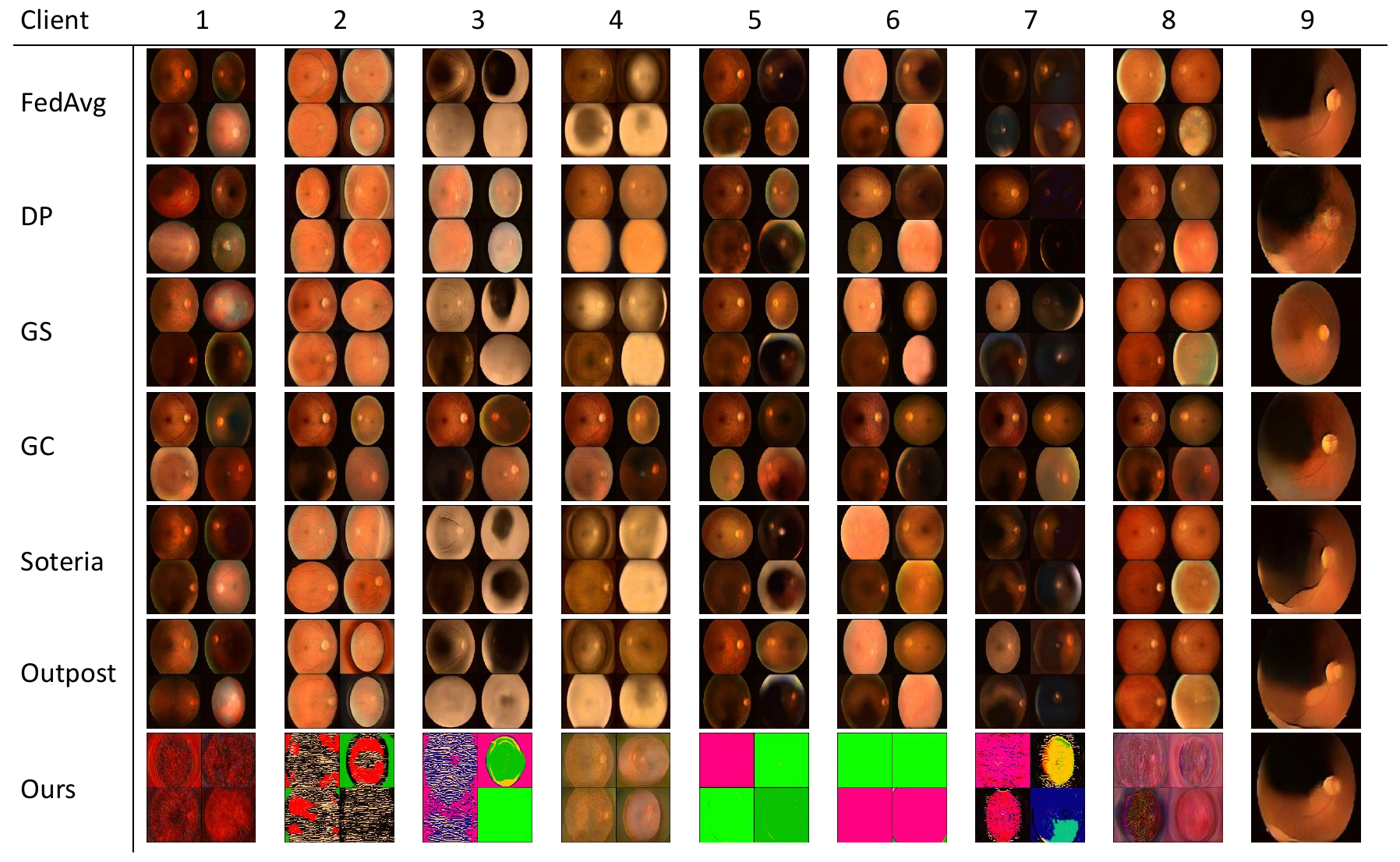}
	\end{center}
	\caption{Reconstructed images from model-based GIA at 50th global round on the EyePACS dataset.}
	\label{fig S12}
\end{figure*}

\begin{figure*}
	\begin{center}
	\includegraphics[width=2\columnwidth]{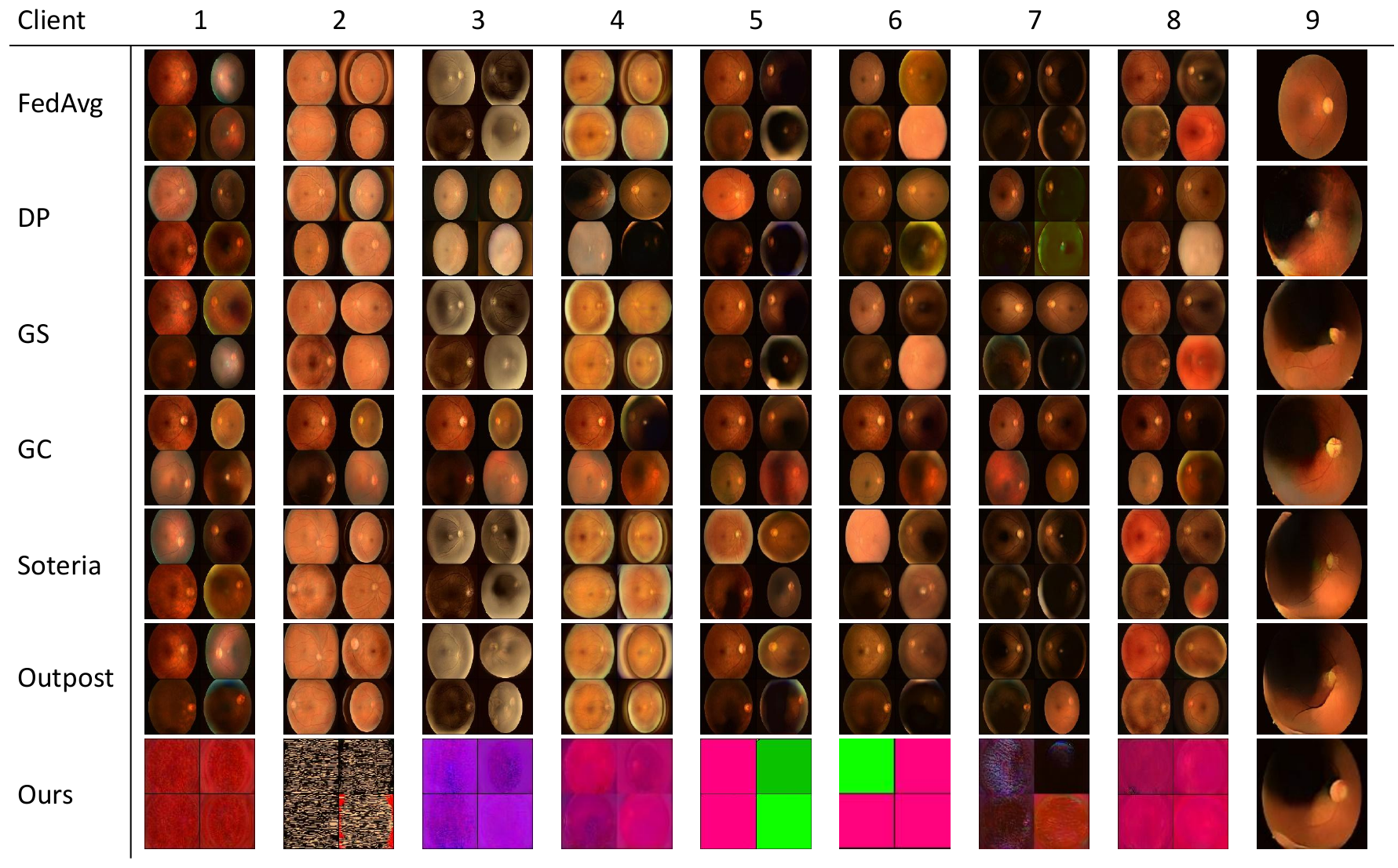}
	\end{center}
	\caption{Reconstructed images from model-based GIA at 100th global round on the EyePACS dataset.}
	\label{fig S13}
\end{figure*}

\begin{figure*}
	\begin{center}
	\includegraphics[width=2\columnwidth]{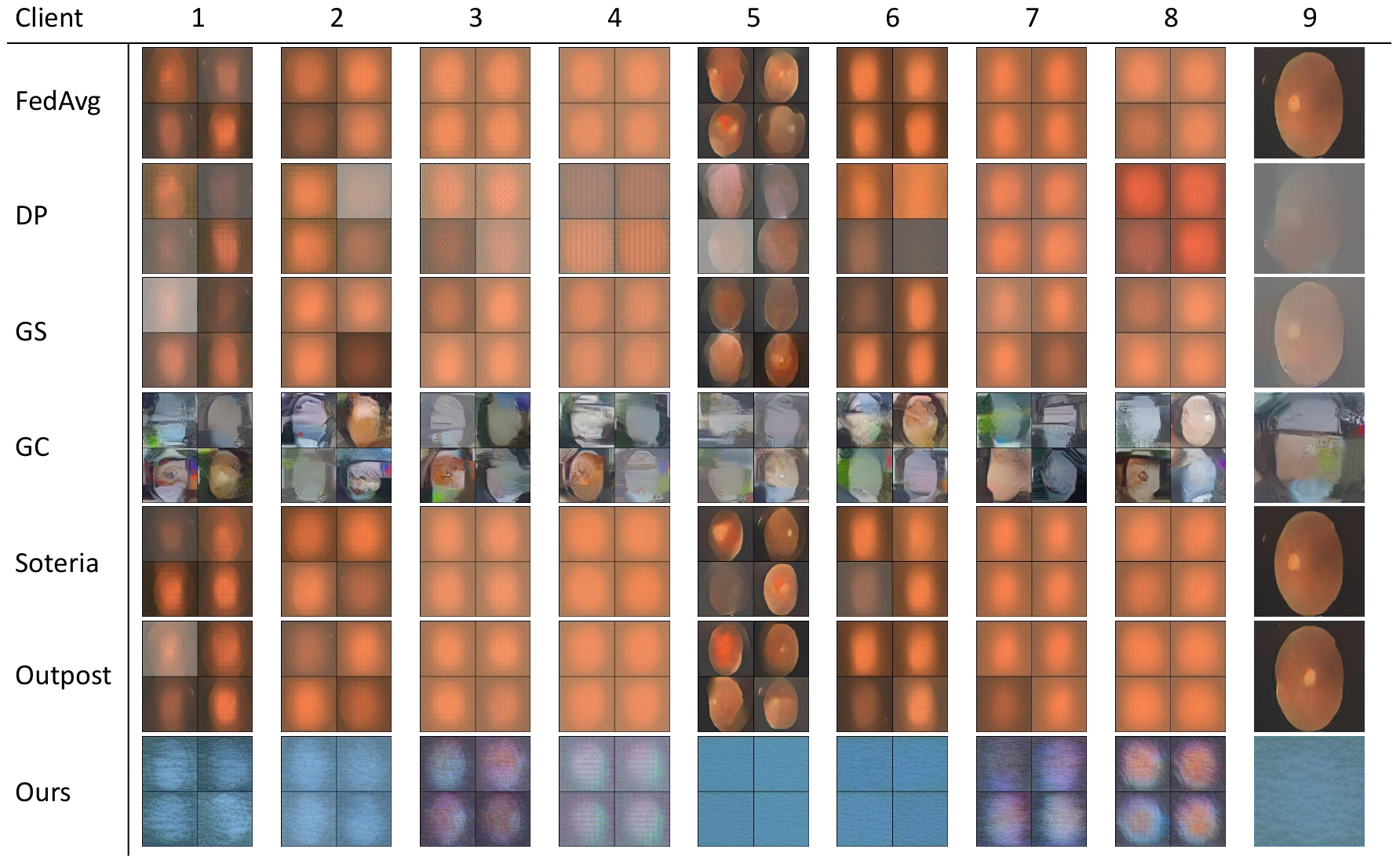}
	\end{center}
	\caption{Reconstructed images from optimization-based GIA at 1st global round on the EyePACS dataset.}
	\label{fig S14}
\end{figure*}

\begin{figure*}
	\begin{center}
	\includegraphics[width=2\columnwidth]{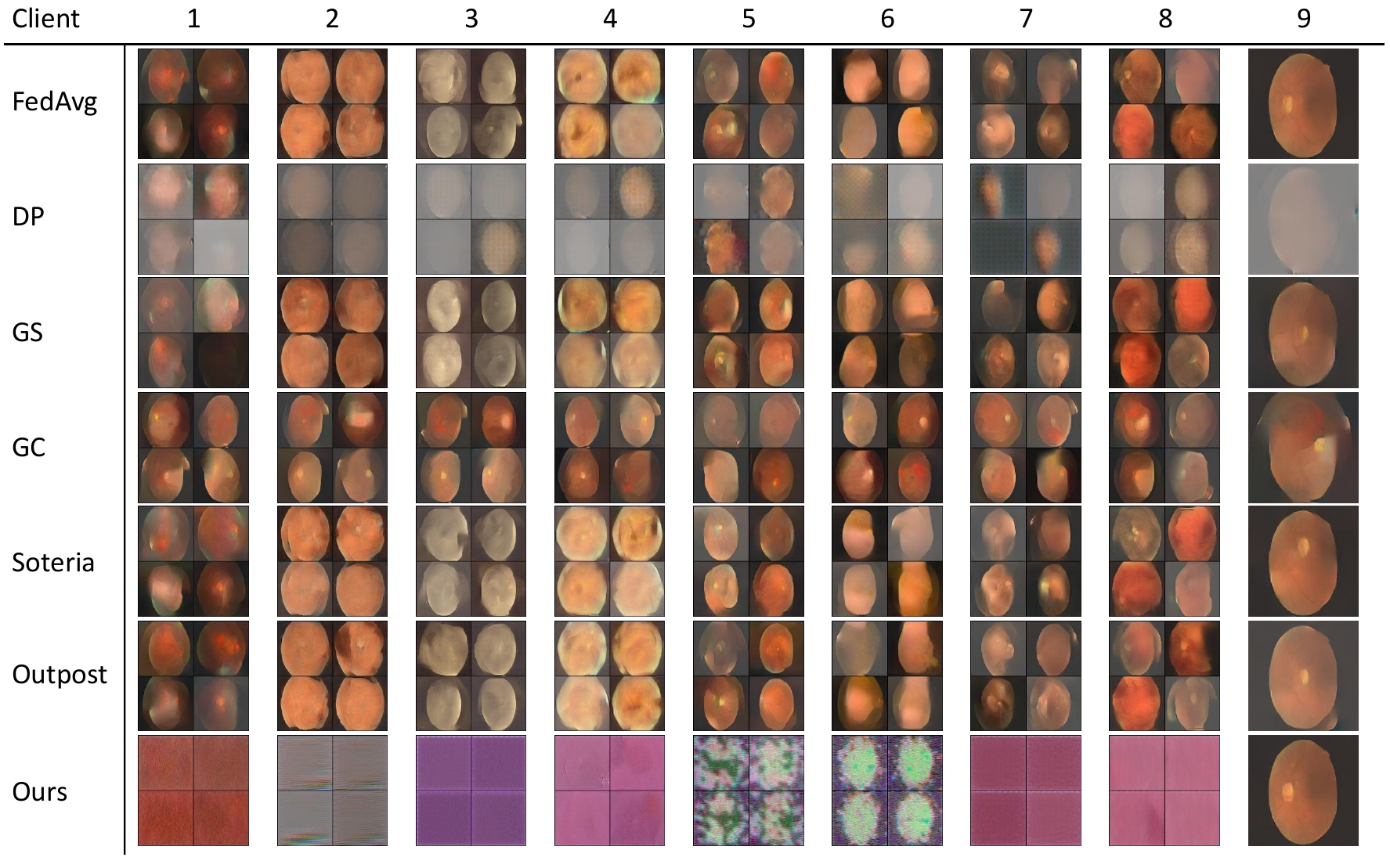}
	\end{center}
	\caption{Reconstructed images from optimization-based GIA at 100th global round on the EyePACS dataset.}
	\label{fig S15}
\end{figure*}

\textcolor{SkyBlue}{Fig.} \ref{fig S13}, \ref{fig S14}, \ref{fig S15} show reconstruction results of the EyePACS dataset from model-based GIA during the 1st, 50th, and 100th rounds of federated training. \textcolor{SkyBlue}{Fig.} \ref{fig S14}, \ref{fig S15} illustrate results of optimization-based GIA at the 1st and 100th global rounds for the same dataset. Compared to the ChestXRay dataset, it is evident that in the EyePACS dataset, clients with fewer samples, such as clients 1, 5, and 9, already exhibit significant privacy leaks for both types of GIA early in the training process. Conclusions during the middle and later stages of training are similar to those for the ChestXRay dataset. Overall, these visual results reaffirm that our shadow defensive framework provides a universally effective privacy protection capability across different medical imaging datasets and types of GIA.

\section{GIA defense on medical image segmentation}   \label{GIA defense on medical image segmentation}

To test effectiveness of our method in medical image segmentation, we apply our method in the prostate MRI dataset \cite{nicholas2015nci, lemaitre2015computer, litjens2014evaluation}. Similar to the setting in HarmoFL \cite{jiang2022harmofl}, three sequential slices are stacked into an RGB like image and the goal is to segment the middle slice. Differently, we change the batch size and do not use data augmentation for better attack. Besides, global epoch has been changed from 500 to 200. Specially, batch size is set as follows: 8 for client 1 and 5, 16 for client 2 and 4, 4 for client 3 and 6. To our best knowledge, there is no label restoration work in GIA for the image segmentation task. Therefore, we use segmentation masks as known prior, and fix them during training. We use training data of prostate158 \cite{adams2022prostate158} to pretrain our shadow model. Learning rate for obtaining latent codes and fine-tunning the shadow model are 1e-4 and 1e-6, respectively.

\begin{figure*}
	\begin{center}
	\includegraphics[width=1.5\columnwidth]{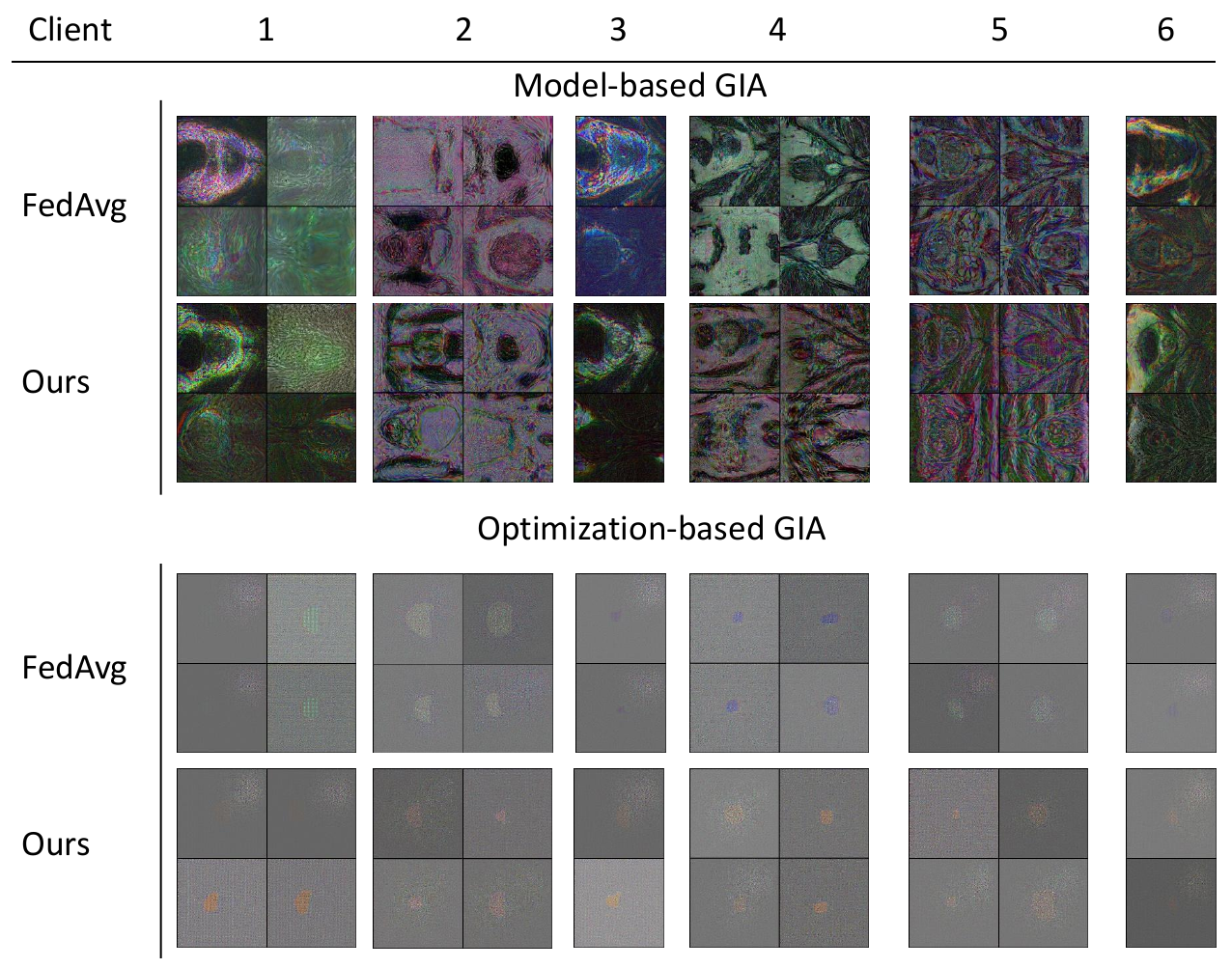}
	\end{center}
	\caption{Reconstructed images of the prostate dataset.}
	\label{fig S16}
\end{figure*}

Results are shown in Table \ref{table 3} and \textcolor{SkyBlue}{Fig.} \ref{fig S16}. For the task performance, i.e., Dice, our method only reduces it by 0.008. For all defense metrics, our method achieves improvements, proving the wide applicability of our proposed framework.

\begin{table*}[]
	\caption{Effectiveness of our method in image segmentation on the prostate dataset.}
    \centering
	\resizebox{2\columnwidth}{!}{
		\begin{tabular}{llllllllllll}
			\hline
			\multicolumn{6}{c}{Optimization-based GIA} & \multicolumn{6}{c}{Model-based GIA} \\
			\cmidrule(r){1-6}\cmidrule(r){7-12}
			\multicolumn{6}{c}{Whole Image}                 & \multicolumn{6}{c}{Whole Image}                 \\
			\hline
			Method & Dice↑ & MSE↑  & PSNR↓ & LPIPS↑ & SSIM↓ & Method & Dice↑ & MSE↑  & PSNR↓ & LPIPS↑ & SSIM↓ \\
			\hline
			FedAvg & 0.901 & 0.062 & 12.18 & 1.030  & 0.089 & FedAvg & 0.901 & 0.050 & 13.23 & 0.663  & 0.102 \\
			ours   & 0.893 & 0.066 & 11.95 & 1.043  & 0.088 & ours   & 0.893 & 0.051 & 13.09 & 0.686  & 0.094 \\
			\hline
			\multicolumn{6}{c}{Target Region} & \multicolumn{6}{c}{Target Region} \\
			\hline
			Method & Dice↑ & MSE↑  & PSNR↓ & LPIPS↑ & SSIM↓ & Method & Dice↑ & MSE↑  & PSNR↓ & LPIPS↑ & SSIM↓ \\
			\hline
			FedAvg & 0.901 & 0.038 & 14.32 & 0.416  & 0.434 & FedAvg & 0.901 & 0.027 & 15.97 & 0.452  & 0.388 \\
			ours   & 0.893 & 0.041 & 14.01 & 0.420  & 0.430 & ours   & 0.893 & 0.028 & 15.73 & 0.483  & 0.373 \\
			\hline
		\end{tabular}
	}
  \label{table 3}
\end{table*}

\section{GIA defense on Vision Transformer}   \label{GIA defense on Vision Transformer}
\begin{figure*}
	\begin{center}
	\includegraphics[width=2\columnwidth]{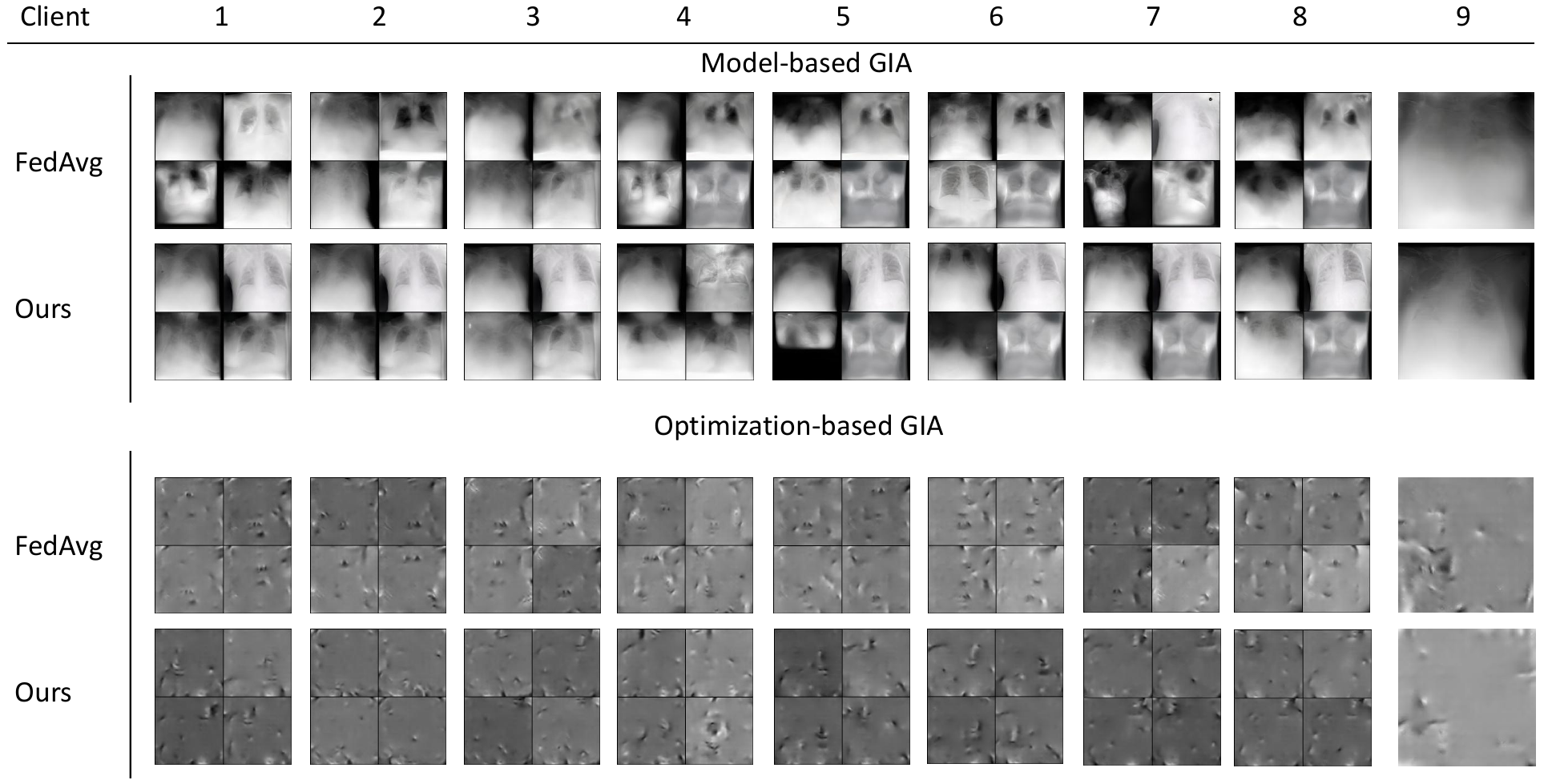}
	\end{center}
	\caption{Reconstructed images based on ViT.}
	\label{fig S17}
\end{figure*}

To test effectiveness of our method in larger models, we apply our method to Vision Transformer (ViT) \cite{han2022survey} based on the ChestXRay dataset. Results are shown in Table \ref{table 4} and \textcolor{SkyBlue}{Fig.} \ref{fig S17}. Under this circumstance, the F1 gap between `FedAvg' and `ours' is 0.043, which is larger than the one of convolutional neural network (CNN). The reason is probably that ViT tends to overfit these noise in training images, which can be also validated on the performance drop of FedAvg compared with CNN. We also find that our proposed framework is worse on defensive metrics. To find out the reason, we show reconstructed images based on ViT. Due to a lack of batch normalization layer, GIA methods do not work even in FedAvg on client 9. Thus, under this circumstance, defense is not neccessary.

\begin{table*}[]
	\caption{Effectiveness of our method on ViT.}
    \centering
	\resizebox{2\columnwidth}{!}{
		\begin{tabular}{llllllllllll}
			\hline
			\multicolumn{6}{c}{Optimization-based GIA}      & \multicolumn{6}{c}{Model-based GIA}             \\
			\cmidrule(r){1-6}\cmidrule(r){7-12}
			\multicolumn{6}{c}{Whole Image}                 & \multicolumn{6}{c}{Whole Image}                 \\
			\hline
			Method & F1↑ & MSE↑  & PSNR↓ & LPIPS↑ & SSIM↓ & Method & F1↑ & MSE↑  & PSNR↓ & LPIPS↑ & SSIM↓ \\
			\hline
			FedAvg & 0.944 & 0.031 & 16.27 & 0.721  & 0.618 & FedAvg & 0.944 & 0.046 & 14.11 & 0.486  & 0.566 \\
			ours   & 0.901 & 0.027 & 17.24 & 0.691  & 0.647 & ours   & 0.901 & 0.046 & 14.06 & 0.507  & 0.560 \\
			\hline
			\multicolumn{6}{c}{Target Region}               & \multicolumn{6}{c}{Target Region}               \\
			\hline
			Method & F1↑ & MSE↑  & PSNR↓ & LPIPS↑ & SSIM↓ & Method & F1↑ & MSE↑  & PSNR↓ & LPIPS↑ & SSIM↓ \\
			\hline
			FedAvg & 0.944 & 0.020 & 18.73 & 0.251  & 0.777 & FedAvg & 0.944 & 0.026 & 16.92 & 0.164  & 0.742 \\
			ours   & 0.901 & 0.016 & 19.94 & 0.206  & 0.804 & ours   & 0.901 & 0.028 & 16.65 & 0.199  & 0.740
			\\ \hline
			\end{tabular}
	}
  \label{table 4}
\end{table*}

\section{GIA defense against CI-Net}   \label{GIA defense against CI-Net}
\begin{figure*}
	\begin{center}
	\includegraphics[width=2\columnwidth]{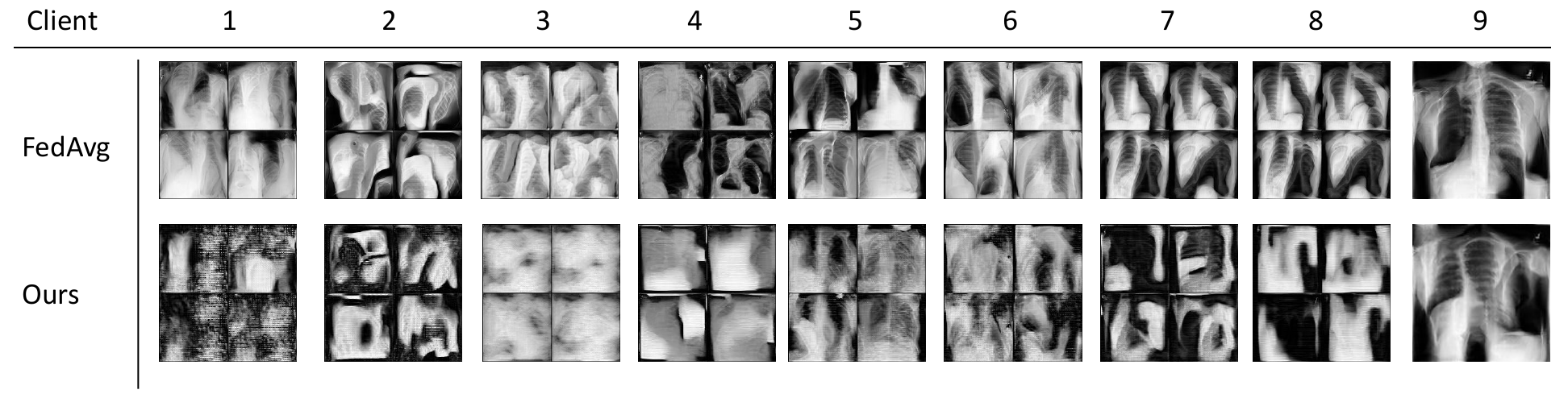}
	\end{center}
	\caption{Reconstructed images from CI-Net on the ChestXRay dataset.}
	\label{fig S18}
\end{figure*}

Unlike traditional model-based GIA, CI-Net \cite{zhang2023generative} do not use GAN as backbone. Instead, an over-parameterized CNN is utilized due to the motivation that the more the searching space is, the more likely training data will be reconstructed. Besides, CNN is prone to learn low-frequency semantics first, thus avoiding overfitting these high-frequency random-noise in images. To achieve pixel affinity of neighborhood regions, the variant of the progressive-growing network with nearest-interpolation is used and Resnet-blocks are removed. Results are shown in Table \ref{table 5} and \textcolor{SkyBlue}{Fig.} \ref{fig S18}. It can be seen that our method is workable to such attack, even if our pipeline is quite different from it, which demonstrates generalization of our framework.

\begin{table*}[]
	\caption{Effectiveness of our method against CI-Net.}
    \centering
	\resizebox{1.5\columnwidth}{!}{
		\begin{tabular}{llllllllllll}
			\hline
			\multicolumn{6}{c}{Model-based GIA}             \\
			\hline
			\multicolumn{6}{c}{Whole Image}                 \\
			\hline
			Method & F1↑   & MSE↑  & PSNR↓ & LPIPS↑ & SSIM↓ \\
			\hline
			FedAvg & 0.978 & 0.056 & 12.90 & 0.522  & 0.413 \\
			ours   & 0.967 & 0.091 & 11.54 & 0.583  & 0.307 \\
			\hline
			\multicolumn{6}{c}{Target Region}               \\
			\hline
			Method & F1↑   & MSE↑  & PSNR↓ & LPIPS↑ & SSIM↓ \\
			\hline
			FedAvg & 0.978 & 0.036 & 14.79 & 0.252  & 0.618 \\
			ours   & 0.967 & 0.062 & 13.54 & 0.366  & 0.525
			\\ \hline
			\end{tabular}
	}
  \label{table 5}
\end{table*}

\section{Analytics-based GIA}   \label{Analytics-based GIA}
Analytics-based GIA is based a malicious server, which is deviated from our assumption, i.e., an honest but curious server. To be specific, in analytics-based GIA, certain parameters are added into the task model, or some original parameters are modified to save activations of training data. We first consider the circumstance of adding parameters. We choose LOKI \cite{zhao2024loki} as an example. Although only one convolutional and two FC layers are added, the parameter size of FC layers are positively related to image resolution. When we test LOKI in ResNet18 on image resolution of 224 (only 32 in the experiment of LOKI), the memory assumption is almost 24MB, which is too suspicious for clients to notice the abnormality. Besides, due to the huge amount of attack layer parameters, training fails to converge.

\begin{figure*}
	\begin{center}
	\includegraphics[width=1.5\columnwidth]{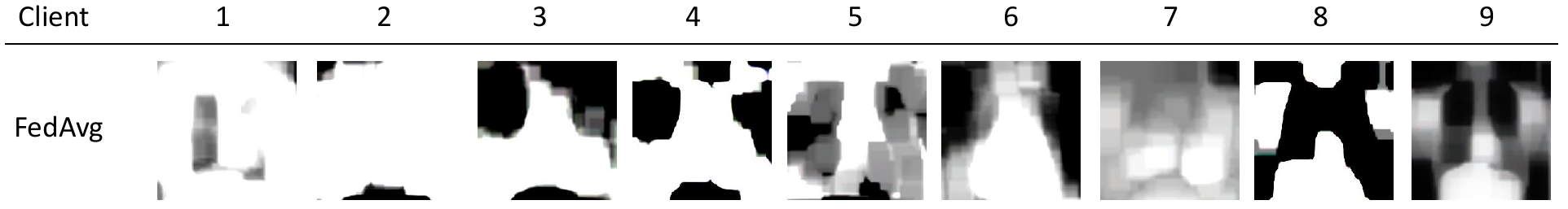}
	\end{center}
	\caption{Reconstructed images from MKOR on the ChestXRay dataset.}
	\label{fig S19}
\end{figure*}

As a better solution of analytics-based GIA, maximum knowledge orthogonality reconstruction (MKOR) \cite{wang2024maximum} carefully designs weight modifications for VGG based on mathematically proven formulations, enabling reconstruction for high-resolution images. Results are shown in Table \ref{table 6} and \textcolor{SkyBlue}{Fig.} \ref{fig S19}. Although modified parameters are hard to detect, when we use original setting of the ChestXRay dataset, training fails to converge as well (no task performance shown in experiments of MKOR). Thus, we change the optimizer from SGD to Adam and reduce the learning rate to 1e-5. It is observed that task performance is worse and the reconstruction results can hardly reveal any privacy except for client 9, which does not require additional defense strategies. 

\begin{table*}[]
	\caption{Results of MKOR.}
    \centering
	\resizebox{1.5\columnwidth}{!}{
		\begin{tabular}{llllll}
			\hline
			\multicolumn{6}{c}{Model-based GIA}             \\
			\hline
			\multicolumn{6}{c}{Whole Image}                 \\
			\hline
			Method & F1↑   & MSE↑  & PSNR↓ & LPIPS↑ & SSIM↓ \\
			\hline
			FedAvg & 0.839 & 9.527 & 0.70  & 0.696  & 0.397 \\
			\hline
			\multicolumn{6}{c}{Target Region}               \\
			\hline
			Method & F1↑   & MSE↑  & PSNR↓ & LPIPS↑ & SSIM↓ \\
			\hline
			FedAvg & 0.839 & 0.082 & 11.43 & 0.391  & 0.598
			\\ \hline
		\end{tabular}
	}
  \label{table 6}
\end{table*}

\section{GIA defense on non-medical images}   \label{GIA defense on non-medical images}
\begin{figure*}
	\begin{center}
	\includegraphics[width=2\columnwidth]{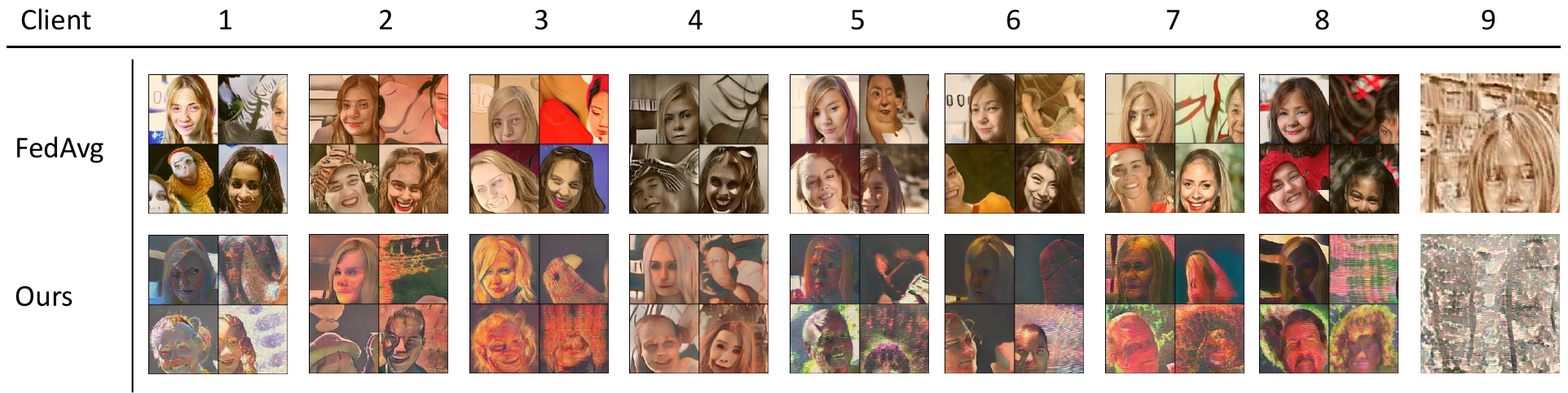}
	\end{center}
	\caption{Reconstructed images based on the VGGFace2 dataset.}
	\label{fig S20}
\end{figure*}

To test our framework in non-image datasets, we test it on the VGGFace2 dataset \cite{cao2018vggface2}. We choose 10 identities (classes) with most number of images to construct our FL dataset (except for the couple identity). In the pre-processing step, centering crop based on relative size of height and width and resizing of 224 resolution are performed. Then, we assign each of 9 clients with class-balanced division of training, validation, and testing datasets. Other experimental setting are same as the one of ChestXRay dataset, except for the usage of SGD with nesterov. Results are shown in Table \ref{table 7} and \textcolor{SkyBlue}{Fig.} \ref{fig S20}. Data privacy of these human face is effectively protected by our framework, as can be seen in defensive metrics and visualization. However, unlike medical image datasets, adding adaptive noise to face images significantly harms the task performance, since key attributes for recognizing human face are susceptible to such noise. It remains an issue to improve our method in the non-medical scenarios.

\begin{table*}[]
	\caption{Effectiveness of our method on the VGGFace2 dataset.}
    \centering
	\resizebox{1.5\columnwidth}{!}{
		\begin{tabular}{llllll}
			\hline
			\multicolumn{6}{c}{Model-based GIA}             \\
			\hline
			\multicolumn{6}{c}{Whole Image}                 \\
			\hline
			Method & F1↑   & MSE↑  & PSNR↓ & LPIPS↑ & SSIM↓ \\
			\hline
			FedAvg & 0.916 & 0.080 & 11.28 & 0.655  & 0.285 \\
			ours   & 0.791 & 0.114 & 9.73  & 0.704  & 0.267 \\
			\hline
			\multicolumn{6}{c}{Target Region}               \\
			\hline
			Method & F1↑   & MSE↑  & PSNR↓ & LPIPS↑ & SSIM↓ \\
			\hline
			FedAvg & 0.916 & 0.053 & 13.13 & 0.418  & 0.531 \\
			ours   & 0.791 & 0.075 & 11.60 & 0.443  & 0.520 \\
			\hline
			\end{tabular}
	}
  \label{table 7}
\end{table*}

\bibliographystyle{model2-names.bst}\biboptions{authoryear}
\bibliography{example}